%% file: main.tex
\documentclass[11pt]{article}
\usepackage[margin=1in]{geometry}
\usepackage[T1]{fontenc}
\usepackage{lmodern}
\usepackage[round]{natbib}
\input{macros}
\hypersetup{colorlinks=true,linkcolor=blue!50!black,citecolor=blue!50!black,urlcolor=blue!50!black}
\usepackage{rotating}
\usepackage{microtype}
\title{What Interpolators Store and What They Express:\\ An Information-Theoretic Framework for Benign Overfitting\\ and Label-Free Model Selection}
\author{Nikolaos Nikolaou\\[2pt] \normalsize University College London, UK\\ \normalsize\texttt{n.nikolaou@ucl.ac.uk}}
\date{November 2026}

\begin{document}
\maketitle

\input{sections/abstract}
\input{sections/s1_intro}
\input{sections/s2_setting}

\input{sections/s3_sufficiency}
\input{sections/s4_storage_expression}
\input{sections/s5_benign}
\input{sections/s6_selection}
\input{sections/s7_experiments}
\input{sections/s8_related}
\input{sections/s9_discussion}

\subsubsection*{Acknowledgments}
We used Claude (Anthropic; Claude Opus 5.5) to implement and run the experiments, derive certain results and edit and revise the manuscript text, tables and figures. The authors conceived and designed the study, specified the experiments, checked the code, proofs and results, and drafted the final text. They take full responsibility for the content of the submission.

\clearpage
\appendix
\renewcommand{\thesection}{\Alph{section}}
\setcounter{theorem}{0}\renewcommand{\thetheorem}{S\arabic{theorem}}
\setcounter{remark}{0}\renewcommand{\theremark}{S\arabic{remark}}
\setcounter{table}{0}\renewcommand{\thetable}{S\arabic{table}}
\setcounter{figure}{0}\renewcommand{\thefigure}{S\arabic{figure}}
\setcounter{equation}{0}\renewcommand{\theequation}{S\arabic{equation}}
\renewcommand{\theHtheorem}{S\arabic{theorem}}\renewcommand{\theHremark}{S\arabic{remark}}\renewcommand{\theHtable}{S\arabic{table}}\renewcommand{\theHfigure}{S\arabic{figure}}\renewcommand{\theHequation}{S\arabic{equation}}
\begin{center}{\Large\bfseries Appendices (supplementary material)}\end{center}

\input{appendix}

\clearpage
\bibliography{refs}
\bibliographystyle{plainnat}

\end{document}

%% file: macros.tex
\usepackage{amsmath,amssymb,amsthm,mathtools,bm}
\usepackage{graphicx,booktabs,multirow,enumitem,xcolor,array}
\usepackage{tikz}
\usetikzlibrary{arrows.meta,positioning,fit,calc}
\usepackage{hyperref}
\usepackage{url}

\theoremstyle{plain}
\newtheorem{theorem}{Theorem}
\newtheorem{proposition}[theorem]{Proposition}
\newtheorem{lemma}[theorem]{Lemma}
\newtheorem{corollary}[theorem]{Corollary}
\theoremstyle{definition}
\newtheorem{definition}{Definition}
\newtheorem{assumption}{Assumption}
\newtheorem{example}{Example}
\theoremstyle{remark}
\newtheorem{remark}{Remark}

\newcommand{\E}{\mathbb{E}}
\renewcommand{\P}{\mathbb{P}}
\newcommand{\cX}{\mathcal{X}}
\newcommand{\cY}{\mathcal{Y}}
\newcommand{\cA}{\mathcal{A}}
\newcommand{\cR}{\mathcal{R}}
\newcommand{\cE}{\mathcal{E}}
\newcommand{\ind}{\mathbf{1}}
\newcommand{\Var}{\operatorname{Var}}
\newcommand{\KL}{\mathrm{KL}}
\DeclareMathOperator*{\argmin}{arg\,min}
\DeclareMathOperator*{\argmax}{arg\,max}
\newcommand{\prop}{$^\dagger$}   
\newcommand{\best}[1]{\textbf{#1}}

%% file: sections/abstract.tex
\begin{abstract}
Overparameterized learners can fit noisy training labels perfectly and still predict well on new data. Classical complexity measures do not explain this, and neither do information-theoretic generalization bounds. We develop an information-theoretic framework that separates two things an interpolating learner does with its training sample. The first is \emph{storage}: the information that the trained model retains about the training labels. The second is \emph{expression}: the information about the training sample that reaches the model's predictions at fresh inputs.
We start from the ideal model. For a loss $\ell$, a representation of the input is \emph{$\ell$-sufficient} if some decoder of it attains the Bayes risk. The least complex sufficient representation is the Bayes act itself, so the \emph{minimal sufficient complexity} is its entropy. Any additional complexity of a model's outputs is \emph{excess complexity}. These notions generalize, and correct, an earlier ``lossless maximal compression'' (LMC) view of boosting, proposed by the author in two earlier non-archival papers.
We then show that storage cannot tell interpolators apart: every interpolator stores exactly $n\,H(Y\mid X)$ bits of label noise. Expression can. We define the \emph{expressed information} $V(x)$ as the entropy, over training samples, of the prediction at input $x$, and prove that excess risk is bounded above and below by expressed information plus a modal-bias term (exactly so for binary labels). Benign overfitting is therefore vanishing expression: a learner can store the maximum amount of label noise and still express almost none, in which case its outputs converge to the minimal sufficient representation. Nearest-neighbour rules, averaging, minimum-norm least squares and double descent all fit this picture.
Finally, we show that among overparameterized learners, expressed information is a viable criterion for model selection that needs no labels for the selection step. With disjoint data halves it costs about as much as one holdout fit. On 15 real datasets with label noise (five seeds each), it has 60\% less regret than holdout validation and 44\% less than 2-fold cross-validation, and it matches 5-fold cross-validation at a quarter of the cost. We also make explicit how expressed information relates to classical bias--variance decompositions and to the disagreement between independently trained models, and show that disagreement tracks excess risk rather than risk. We characterize when $V$ fails (underfitting candidates and asymmetric label noise) and give label-free and noise-corrected remedies. Code to reproduce every result is available at \url{https://github.com/nnikolaou/GLMC}.
\end{abstract}

%% file: sections/s1_intro.tex
\section{Introduction}\label{sec:intro}

Modern learners are routinely trained until they fit their training data perfectly, even when some of the training labels are wrong. Deep networks fit randomly relabelled data \citep{zhang2016understanding}. Random forests and boosted ensembles reach zero training error and keep improving \citep{schapire1998boosting,wyner2017explaining}. Minimum-norm least squares interpolates noisy targets and can still be consistent \citep{bartlett2020benign,hastie2022surprises}. This behaviour is called \emph{benign overfitting}. It is not universal: other interpolators, such as the 1-nearest-neighbour rule, settle at a fixed excess risk (\emph{tempered} overfitting), and some fail completely (\emph{catastrophic} overfitting) \citep{mallinar2022taxonomy}. All three kinds of learner have the same training error, zero, so whatever separates them is invisible on the training set.

In this paper we ask two questions about interpolating learners.
\begin{enumerate}[label=(Q\arabic*),leftmargin=*]
\item \emph{What separates a benign interpolator from a harmful one?} In particular, is there an information-theoretic quantity that does so, given that interpolators all memorize their training labels?
\item \emph{Can we use that quantity to choose among interpolating learners?} Training error cannot rank them, and validation labels may be scarce or noisy themselves.
\end{enumerate}

\paragraph{Why information theory, and why it has struggled here.} Information theory gives a natural language for ``how much a model has learned from its data''. Information-theoretic generalization bounds control the generalization gap by the mutual information $I(W;S)$ between the trained weights $W$ and the training sample $S$ \citep{russo2016controlling,xu2017information,steinke2020cmi}. The information bottleneck (IB) describes good representations as those that compress the input while keeping what is relevant for the label \citep{tishby2000ib,tishby2015deep}. Both views have difficulties with interpolators. Any learner that fits noisy labels must store them, so $I(W;S)$ is large for every interpolator, benign or not \citep{harutyunyan2020label,livni2023lower}. And for deterministic models the IB quantities are degenerate \citep{amjad2019learning,kolchinsky2019caveats,saxe2018ib}.

\paragraph{An earlier attempt.} Two earlier non-archival papers by the author described margin-maximizing boosting as \emph{lossless maximal compression} (LMC): a model whose output keeps all the information about the label that the input has, while being as compressed as possible \citep{nikolaou2020margin,nikolaou2021lossless}; we refer to them as the \emph{LMC papers}. That work had flaws. Its central object does not exist when labels are noisy, its empirical estimates are degenerate for any interpolator, and some of its experimental claims do not survive replication. We therefore do \emph{not} treat it as an established framework. This paper builds the framework from scratch and corrects the earlier work as it goes. Appendix~\ref{sec:erratum} gives a full erratum.

\paragraph{Our approach in one paragraph.} We separate what a learner \emph{stores} from what it \emph{expresses}. Storage is the information the trained model keeps about the training labels, $I(W;Y^n\mid X^n)$. Expression is the information about the training sample that reaches the model's prediction at a fresh input $x$. We measure it by the \emph{expressed information}
\[
V(x)\;=\;H\big(h_W(x)\big),
\]
the entropy\footnote{We write $V$ because it plays the role of a \emph{variance}. In the bias--variance picture, modal bias (Section~\ref{sec:expression}) takes the place of bias and $V$ the place of variance: under squared loss the classical variance is a limit of $V$ (Theorem~\ref{thm:sq}), and the 0-1 variances of \citet{kohavi1996bias} and \citet{domingos2000unified} are equivalent to $V$ up to explicit bounds (Proposition~\ref{prop:dictionary}).} of the prediction $h_W(x)$ at $x$ when the training sample (and any training randomness) is drawn afresh. In plain terms, $V(x)$ is zero if every training sample leads to the same prediction at $x$, and it is large if the prediction at $x$ depends on which particular sample, and so which particular noisy labels, the learner happened to see. Section~\ref{sec:plain} explains both words informally, with an analogy; in this sense, both are terms introduced in this paper (Table~\ref{tab:terms} lists which of our terms are new and which are standard). Storage and expression behave very differently for interpolators. Every interpolator stores \emph{exactly} $n\,H(Y\mid X)$ bits of label noise (Theorem~\ref{thm:storage}), so storage cannot tell them apart. Expression can: excess risk is controlled from above and below by expressed information plus a ``modal bias'' term (Theorem~\ref{thm:decomp}).

\paragraph{A running example.} We will use one simple task throughout the paper (Example~\ref{ex:running} in Section~\ref{sec:setting}). The input consists of one informative discrete feature and 16 irrelevant binary features; the label is a noisy function of the informative feature. Every learner in Table~\ref{tab:pilot} interpolates its noisy training labels, including a \emph{memorize-and-smooth} learner that we construct for this purpose,\footnote{The memorize-and-smooth learner stores the training sample as a look-up table. On an input that appears in the training sample it returns the memorized (possibly wrong) label; on any other input it predicts the majority training label among the training examples with the same value of the informative feature, that is, a smoothed vote that single noisy labels barely move. It therefore fits every training label, yet its predictions at new inputs almost never use the memorized labels. Proposition~\ref{prop:memsmooth} analyses it.} and every one has the same score under the old LMC criterion.\footnote{The \emph{legacy LMC score} of the LMC papers is $I_S(F;Y)/H_S(Y)$: the mutual information between the model's hard predictions $F$ and the labels $Y$, divided by the entropy of the labels, both computed on the training sample. A score of $1$ means that the predictions retain all the information about the labels present in the sample. Section~\ref{sec:lmc} explains why every interpolator scores (almost exactly) $1$.} Yet their excess risks differ by two orders of magnitude, and their expressed information ranks them correctly. We return to this example at each step of the development.

\begin{table}[t]\centering\small
\caption{The running example (Example~\ref{ex:running}) with $n=4096$ training points and label noise $\varepsilon=0.2$. All five learners interpolate. The legacy LMC score of \citet{nikolaou2020margin} is the same for all of them, while their excess risks differ by two orders of magnitude. Expressed information $V$ (Section~\ref{sec:expression}) ranks them correctly. Best value per column in bold (the two degenerate columns are identical for all learners). \prop: learner constructed in this paper (Proposition~\ref{prop:memsmooth}).}\label{tab:pilot}
\begin{tabular}{lcccc}\toprule
Learner & Training error & Legacy LMC score & $V$ (bits) & Excess risk\\\midrule
Memorize-and-smooth\prop & 0.001 & 0.99 & \best{0.007} & \best{0.001}\\
Random forest & 0.001 & 0.99 & 0.045 & 0.005\\
Gradient boosting & 0.001 & 0.99 & 0.406 & 0.064\\
1-nearest neighbour & 0.001 & 0.99 & 0.642 & 0.125\\
Decision tree (full depth) & 0.001 & 0.99 & 0.660 & 0.131\\\bottomrule
\end{tabular}
\end{table}

\paragraph{Contributions.} Our contributions are the following.
\begin{enumerate}[leftmargin=*]
\item \textbf{A framework for the ideal model} (Section~\ref{sec:sufficiency}). For a general loss we define $\ell$-sufficient representations, the minimal sufficient complexity $C^*_\ell=H(a^*(X))$ (the entropy of the Bayes act), maximal complexity, and excess complexity. We relate them to decision-theoretic sufficiency, to the indirect rate--distortion function, and to the information bottleneck. We show that the original LMC exists only in degenerate cases, and that its generalized form, the minimal $\ell$-sufficient representation, always exists.
\item \textbf{Storage versus expression} (Section~\ref{sec:storage-expression}). We show that every interpolator stores exactly $n\,H(Y\mid X)$ bits of label noise (sharpening a bound of \citealp{harutyunyan2020label}), and that information-theoretic generalization bounds, including those based on conditional mutual information \citep{steinke2020cmi}, therefore cannot separate benign from tempered interpolators. We introduce expressed information and prove a two-sided information decomposition of excess risk, which is exact for binary labels. We show that a learner's output entropy at fresh inputs splits into complexity and expression, and that a consistent learner's outputs converge to the minimal sufficient representation.
\item \textbf{Benign overfitting as vanishing expression} (Section~\ref{sec:benign}). We give an information-theoretic version of the benign/tempered/catastrophic taxonomy. We work it out for an explicit benign interpolator, for nearest-neighbour rules (recovering the classical result of \citealp{cover1967nearest}), for averaging, and for minimum-norm least squares, where the expressed information equals the variance term of \citet{bartlett2020benign}. Double descent appears as a peak of expressed information.
\item \textbf{Expressed information as a model-selection criterion} (Section~\ref{sec:selection}). We show that among candidates with no modal bias, excess risk is monotone in expressed information, and we argue that the overparameterized regime is where this condition is most nearly met. We explain how to estimate $V$ from a single dataset, and why disjoint splits must be used. Because it needs no held-out labels, only unlabelled inputs, it suits settings where validation labels are scarce, delayed or noisy (semi-supervised and streaming data, label noise, covariate shift; Section~\ref{sec:discussion}). We characterize where $V$ alone fails and give two remedies: a label-free consensus criterion, and a switch to noise-corrected validation when the label noise is asymmetric.
\item \textbf{Connections to bias--variance and disagreement} (Section~\ref{sec:bvd}). We show that the 0-1 variances of Kohavi--Wolpert, Domingos and James and the disagreement of two independently trained models are equivalent to $V$ up to explicit bounds, and that disagreement bounds \emph{excess} risk from above and below: the generalization disagreement equality of \citet{jiang2022disagreement} holds exactly for the 1-NN rule and fails by the Bayes risk for benign learners.
\item \textbf{Experiments} (Section~\ref{sec:experiments}) on synthetic tasks with exact excess risk and on 15 real datasets. Among overparameterized candidates on a single noisy dataset, $V$ computed from two disjoint halves has $60\%$ less regret than holdout validation and $44\%$ less than 2-fold cross-validation at about the same cost, and matches 5-fold cross-validation (15 datasets, five seeds each). At matched cost it also beats selection by conditional mutual information, which, unlike $V$, needs held-out labels.
\end{enumerate}

\paragraph{Roadmap.} Section~\ref{sec:setting} fixes notation and introduces the running example. Section~\ref{sec:sufficiency} describes the ideal model. Section~\ref{sec:storage-expression} separates storage from expression, and Section~\ref{sec:benign} uses this separation to explain benign overfitting. Section~\ref{sec:selection} turns expressed information into a selection rule, and Section~\ref{sec:experiments} tests the theory and the rule. Section~\ref{sec:related} discusses connections to existing work, and Section~\ref{sec:discussion} discusses limitations. Proofs, further theory, the erratum for the LMC papers, experimental details, additional results and the extension to regression are in the appendices, which form the supplementary material. Code to reproduce every result is available at \url{https://github.com/nnikolaou/GLMC}, and Appendix~\ref{sec:code} describes it.

%% file: sections/s2_setting.tex
\section{Setting, notation and a running example}\label{sec:setting}

Let us first fix the notation that we will use throughout, and introduce the simple example that we will return to in every section.

\paragraph{Data and Bayes classifier.} A pair $(X,Y)$ is drawn from a distribution $P$ on $\cX\times\cY$. In most of the paper $\cY=[K]=\{1,\dots,K\}$ is a set of $K$ classes; Section~\ref{sec:sq} treats real-valued $Y$. For the information-theoretic statements we take $\cX$ to be finite or countable with $H(X)<\infty$ (continuous inputs are handled by the learner-level quantities of Section~\ref{sec:storage-expression}, which do not need this). We write
\[
\eta(x)=\big(\P(Y=1\mid X=x),\dots,\P(Y=K\mid X=x)\big)
\]
for the vector of class probabilities at $x$. The \emph{Bayes classifier} predicts the most likely class, $b^*(x)=\argmax_k\eta_k(x)$, and the \emph{Bayes risk} $R^*=\P(b^*(X)\ne Y)$ is the smallest achievable error. Its \emph{excess risk}, $R(h)-R^*$ with $R(h)=\P(h(X)\ne Y)$, measures how far a classifier $h$ is from that ideal. When the labels are noisy, $R^*>0$: even the best classifier makes mistakes, because the label is not a function of the input.

\paragraph{Learners.} A training sample $S=((X_1,Y_1),\dots,(X_n,Y_n))$ consists of $n$ independent draws from $P$. A \emph{learner} $A$ maps the sample, and possibly some internal randomness $\xi$ (a random seed, independent of $S$), to a trained model $W=A(S,\xi)$, which induces a classifier $h_W:\cX\to[K]$. We say that the learner \emph{interpolates} if $h_W(X_i)=Y_i$ for every training point, almost surely. The important point to keep in mind is that $W$ is random: a different sample, or a different seed, gives a different trained model. Much of this paper is about how much the predictions of $h_W$ vary because of this randomness.

\paragraph{Information measures.} $H(\cdot)$ denotes Shannon entropy, $H(\cdot\mid\cdot)$ conditional entropy and $I(\cdot\,;\cdot)$ mutual information \citep{cover2006elements}. Unless stated otherwise we measure them in bits. For a probability vector $q$ we write $H(q)=-\sum_kq_k\log_2q_k$, and $h(t)=H((t,1-t))$ is the binary entropy. Two quantities of the data distribution will appear repeatedly: $H(Y\mid X)$, the \emph{label noise} (the uncertainty about $Y$ that remains when $X$ is known), and $I(X;Y)$, the information the input carries about the label.

\paragraph{The running example.} We will illustrate every definition and result on the following task. It is chosen so that all quantities can be computed exactly, and so that it contains the two ingredients that matter: label noise, and input features that carry no information about the label.

\begin{example}[Running example]\label{ex:running}
The input is $X=(Z,U)$, where $Z$ is uniform on $\{0,1,\dots,7\}$ and $U$ is uniform on $\{0,1\}^{16}$, independent of $Z$. The label is $Y\in\{0,1\}$, and it depends on $Z$ only: the clean label is $\ind\{Z\ge4\}$, and it is flipped with probability $\varepsilon$ when $Z$ is even and with probability $\varepsilon/2$ when $Z$ is odd. So $Z$ is the informative part of the input, and the 16 bits of $U$ are pure distractors.
\end{example}

For $\varepsilon=0.2$, which we use throughout unless stated otherwise, the quantities above are as follows.
\begin{itemize}[leftmargin=*]
\item The Bayes classifier is $b^*(X)=\ind\{Z\ge4\}$, and the Bayes risk is $R^*=\frac12(0.2)+\frac12(0.1)=0.15$.
\item The label noise is $H(Y\mid X)=\frac12h(0.2)+\frac12h(0.1)=0.595$ bits per example.
\item $H(X)=3+16=19$ bits, $H(b^*(X))=1$ bit, and $\eta(X)$ takes four equally likely values, so $H(\eta(X))=2$ bits.
\end{itemize}
In words: a good model of this task needs to know only one bit about each input (which half $Z$ falls in), out of the 19 bits the input contains. We will see in Section~\ref{sec:sufficiency} that these three numbers, 1, 2 and 19 bits, are respectively the minimal sufficient complexity under 0-1 loss, the minimal sufficient complexity under log-loss, and the maximal complexity. In Section~\ref{sec:storage-expression} we will see that the label noise of $0.595$ bits per example is exactly what every interpolator must store.

\paragraph{Losses.} Sections~\ref{sec:storage-expression}--\ref{sec:selection} focus on classification error (0-1 loss). The definitions of Section~\ref{sec:sufficiency} hold for a general loss $\ell:\cA\times\cY\to[0,\infty]$, where $\cA$ is a set of actions. For a probability vector $q$, let $L(a,q)=\sum_kq_k\,\ell(a,k)$ be the expected loss of action $a$ when the label has distribution $q$. The \emph{Bayes act} at $x$, written $a^*(x)$, is the action that minimizes $L(a,\eta(x))$. Two cases will be used repeatedly.
\begin{itemize}[leftmargin=*]
\item Under \emph{0-1 loss}, $\cA=[K]$ and the Bayes act is the Bayes classifier, $a^*=b^*$.
\item Under \emph{log-loss}, the action is a probability vector $q$, $\ell(q,k)=-\log q_k$, and the Bayes act is the vector of true class probabilities, $a^*=\eta$.
\end{itemize}
We assume, as is standard, that minimizers exist and that the Bayes act is unique for almost every $x$ (for 0-1 loss: there are no ties). Appendix~\ref{sec:assumptions} states these regularity conditions precisely.

\paragraph{Terminology.} Because this paper connects several literatures, and introduces some terms of its own, Table~\ref{tab:terms} lists the main terms, says which are standard, which adapt an existing notion, and which are introduced here, and points to where each is defined.

\begin{table}[t]\centering\small
\caption{Terminology. ``New'': term (and, where stated, quantity) introduced in this paper. ``Adapted'': an existing notion specialized or renamed here. ``Standard'': used as in the cited literature.}\label{tab:terms}
\footnotesize
\begin{tabular}{>{\raggedright\arraybackslash}p{0.33\linewidth}>{\raggedright\arraybackslash}p{0.11\linewidth}>{\raggedright\arraybackslash}p{0.48\linewidth}}\toprule
Term & Status & Definition and origin\\\midrule
$\ell$-sufficient, $\ell$-deficiency & Adapted & Decision-theoretic sufficiency and deficiency \citep{blackwell1953equivalent,lecam1964sufficiency} for one loss; Def.~\ref{def:suff}\\
Recalibrated risk $R_\ell(T)$ & New name & Risk of the best decoder of $T$; standard idea; \S\ref{sec:suff}\\
Minimal sufficient complexity $C^*_\ell=H(a^*(X))$ & New & Under log-loss, the end point of the IB curve \citep{tishby2000ib}; Def.~\ref{def:complexities}\\
Maximal complexity $C_{\max}$ & Adapted & From the LMC papers \citep{nikolaou2020margin}; Def.~\ref{def:complexities}\\
Excess complexity $\cE_\ell$ & New & Redundancy above the indirect rate--distortion function; Def.~\ref{def:complexities}\\
G-LMC & New & Minimal $\ell$-sufficient representation; generalizes the LMC of \citet{nikolaou2020margin}; Def.~\ref{def:complexities}\\
Indirect rate--distortion function & Standard & \citet{dobrushin1962information,witsenhausen1980indirect}; eq.~\eqref{eq:irdf}\\
Storage $I(W;Y^n\mid X^n)$ & New term & The quantity appears in \citet{harutyunyan2020label}; the name ``storage'' is ours; \S\ref{sec:plain}, Thm.~\ref{thm:storage}\\
Expression, expressed information $V(x)$ & New & Def.~\ref{def:V}; relation to existing variances in \S\ref{sec:bvd}\\
Modal prediction $m(x)$ & New name & Domingos' ``main prediction'' \citep{domingos2000unified}; Def.~\ref{def:V}\\
Modal bias & New name & Domingos' bias, here weighted by the margin; Thm.~\ref{thm:decomp}\\
Benign / tempered / catastrophic & Standard & \citet{mallinar2022taxonomy}; information reading in \S\ref{sec:taxonomy}\\
Memorize-and-smooth & New & An explicit benign interpolator; Prop.~\ref{prop:memsmooth}\\
Min-$V$, $V_2$, $V_5$ & New & Selection by least expressed information, estimated from 2 disjoint halves or 5 folds; \S\ref{sec:selection}\\
Consensus criterion, switch rule & New & Label-free modal-bias estimate; asymmetry-triggered noise correction; \S\ref{sec:sel:beyond}\\
Disagreement, GDE & Standard & \citet{jiang2022disagreement}; \S\ref{sec:bvd}\\
CMI, supersample & Standard & \citet{steinke2020cmi}; \S\ref{sec:storage}\\\bottomrule
\end{tabular}
\end{table}

%% file: sections/s3_sufficiency.tex
\section{The ideal model: sufficiency, minimal and excess complexity}\label{sec:sufficiency}

Before studying what learners do, it is useful to describe what an ideal model would look like in information-theoretic terms. In this section we ask: what is the least information about the input that a model's output must carry to be optimal, and how much more than that does a given model carry? This gives us three reference points, the minimal sufficient, maximal and excess complexity, which we will use in Sections~\ref{sec:storage-expression} and~\ref{sec:benign} to describe what benign interpolators converge to. Everything in this section is a property of the data distribution and of a fixed model; learning from samples enters only in Section~\ref{sec:storage-expression}.

\subsection{Representations and sufficiency}\label{sec:suff}

A \emph{representation} of the input is a random variable $T$ computed from $X$, possibly with extra randomness that does not depend on $Y$. Formally, $Y - X - T$ is a Markov chain. The output $F=f(X)$ of a fixed model is a representation, but so is an intermediate layer, a clustering of the inputs, or the input itself. To judge how useful a representation is for a loss $\ell$, we allow ourselves to post-process it in the best possible way. The \emph{recalibrated risk} of $T$ is
\[
R_\ell(T)=\inf_g\ \E\,\ell\big(g(T),Y\big),
\]
where the infimum is over all functions $g$ from the values of $T$ to actions. In plain terms, $R_\ell(T)$ is the risk of the best decision rule that only gets to see $T$. Recalibration makes the theory invariant to relabelling the values of $T$, as mutual information is. For example, a classifier that always outputs the wrong one of two classes is, after recalibration, a perfect classifier. Under log-loss the recalibrated risk is simply the conditional entropy, $R_{\log}(T)=H(Y\mid T)$.

\begin{definition}[$\ell$-sufficiency and $\ell$-deficiency]\label{def:suff}
A representation $T$ is \emph{$\ell$-sufficient} if $R_\ell(T)=R^*_\ell$, where $R^*_\ell=\E\,\ell(a^*(X),Y)$ is the Bayes risk for $\ell$. Its \emph{$\ell$-deficiency} is $\Delta_\ell(T)=R_\ell(T)-R^*_\ell\ge0$.
\end{definition}

In words, $T$ is $\ell$-sufficient if nothing is lost, as far as the loss $\ell$ is concerned, by looking at $T$ instead of at the whole input $X$. The deficiency measures how much is lost. Sufficiency in this sense is the decision-theoretic sufficiency of \citet{blackwell1953equivalent} and \citet{lecam1964sufficiency}, specialized to a single decision problem; deficiency is the single-loss analogue of Le Cam's deficiency. Under log-loss, $\ell$-sufficiency is ordinary statistical sufficiency of $T$ for $Y$, and the deficiency is the information lost, $\Delta_{\log}(T)=I(X;Y\mid T)$. The LMC papers called $\ell$-sufficient representations ``lossless'' \citep{nikolaou2020margin}; we use the standard term here.

The next result says that sufficiency has a simple characterization: a representation is sufficient exactly when the Bayes act can be read off from it.

\begin{proposition}[Sufficiency means the Bayes act is recoverable]\label{prop:suff}
$T$ is $\ell$-sufficient if and only if $a^*(X)=g(T)$ almost surely for some function $g$. Under log-loss, $T$ is sufficient if and only if $\eta(X)$ is a function of $T$, which holds if and only if $I(T;Y)=I(X;Y)$.
\end{proposition}

All proofs for this section are in Appendix~\ref{sec:proofs-suff}. For the running example (Example~\ref{ex:running}) under 0-1 loss, $T$ is sufficient if and only if it reveals whether $Z\ge4$. Under log-loss it must reveal more: it must also reveal whether $Z$ is even or odd, since the noise level, and hence $\eta$, differs between the two.

\subsection{Minimal sufficient complexity and the G-LMC}\label{sec:minimal}

Many representations are sufficient; the input $X$ itself always is. We are interested in the \emph{least complex} one. We measure the complexity of a representation by the information it retains about the input, $I(T;X)$. For a deterministic model output $F=f(X)$, this is just the entropy of the output, $I(F;X)=H(F)$.

\begin{theorem}[Minimal sufficient complexity]\label{thm:minimal}
Every $\ell$-sufficient representation $T$ satisfies
\[
I(T;X)\ \ge\ C^*_\ell:=H\big(a^*(X)\big),
\]
with equality if and only if $T$ depends on $X$ only through $a^*(X)$ (the Markov chain $T - a^*(X) - X$). A sufficient deterministic output $F=f(X)$ attains equality if and only if it is a relabelling of the Bayes act, $F=\varphi(a^*(X))$ for a bijection $\varphi$.
\end{theorem}

Let us restate the theorem in plain terms. To be optimal for a loss, a model must at least be able to tell apart inputs that call for different optimal actions. The entropy of the optimal action is therefore a floor on the complexity of any optimal model, and the Bayes act itself sits exactly on that floor. Anything that distinguishes inputs \emph{within} a set that shares the same optimal action is, for this loss, unnecessary.

\begin{definition}[Complexities and the G-LMC]\label{def:complexities}
\begin{enumerate}[leftmargin=*]
\item The \emph{minimal sufficient complexity} for $\ell$ is $C^*_\ell=H(a^*(X))$. A representation that is $\ell$-sufficient and attains $I(T;X)=C^*_\ell$ is a \emph{minimal $\ell$-sufficient representation}; we also call it a \emph{generalized lossless maximal compression} (G-LMC) of $X$ for $\ell$.
\item The \emph{maximal complexity} is $C_{\max}=H(X)$, the complexity of a model that distinguishes every input. When the outputs range over a model family $\mathcal F$, we use $\sup_{f\in\mathcal F}H(f(X))$ instead.
\item The \emph{excess complexity} of a model output $F$ is
\[
\cE_\ell(F)\;=\;I(F;X)-\cR_\ell\big(R_\ell(F)\big)\ \ge\ 0,
\]
where $\cR_\ell$ is the indirect rate--distortion function defined in \eqref{eq:irdf} below.
\end{enumerate}
\end{definition}

The excess complexity measures how much more complex $F$ is than it needs to be \emph{for the risk it achieves}. For a sufficient model the definition simplifies.

\begin{proposition}[Excess complexity of sufficient models]\label{prop:excess}
If $F$ is $\ell$-sufficient, then $\cE_\ell(F)=I(F;X)-C^*_\ell=H\big(F\mid a^*(X)\big)$. Under log-loss, moreover, $\cE_{\log}(F)=H(F\mid Y)-H(\eta(X)\mid Y)$.
\end{proposition}

So, for a sufficient model, the excess complexity is the information that its output carries beyond the Bayes act: the uncertainty that remains about the output when the optimal action is known. The log-loss identity is useful in practice, because it involves the label $Y$ rather than the unknown $\eta(X)$; it gives a held-out estimator of excess complexity (Appendix~\ref{sec:heldout}).

\begin{example}[Running example, continued]\label{ex:complexities}
Recall from Section~\ref{sec:setting} that in Example~\ref{ex:running}, $H(X)=19$ bits, $H(b^*(X))=1$ bit and $H(\eta(X))=2$ bits. So under 0-1 loss the minimal sufficient complexity is $C^*_{01}=1$ bit and the G-LMC is $b^*(X)=\ind\{Z\ge4\}$; under log-loss, $C^*_{\log}=2$ bits and the G-LMC is $\eta(X)$, which records the half and the parity of $Z$. A model whose output is the whole input, $F=X$, is sufficient for both losses, and its excess complexity is $19-1=18$ bits under 0-1 loss: everything it records about $U$ and about $Z$ beyond its half is unnecessary. A model that outputs $Z$ has excess complexity $3-1=2$ bits under 0-1 loss and $3-2=1$ bit under log-loss.
\end{example}

\subsection{Connection to rate--distortion and the information bottleneck}\label{sec:rd}

The minimal sufficient complexity is one point on a trade-off curve. If we allow some deficiency, we can compress further. The \emph{indirect rate--distortion function} \citep{dobrushin1962information,witsenhausen1980indirect} gives the least complexity compatible with a given risk:
\begin{equation}\label{eq:irdf}
\cR_\ell(D)\;=\;\inf\big\{\,I(T;X)\ :\ Y - X - T,\ \ R_\ell(T)\le D\,\big\}.
\end{equation}
It is ``indirect'' because the representation is computed from $X$ but judged on $Y$, which is only observed through $X$. Theorem~\ref{thm:minimal} says that at zero deficiency, $\cR_\ell(R^*_\ell)=C^*_\ell$, attained by $T=a^*(X)$. The excess complexity of Definition~\ref{def:complexities} is the distance of a model above this curve, which in source-coding terms is its \emph{redundancy}.

Under log-loss, $R_{\log}(T)=H(Y\mid T)=H(Y)-I(T;Y)$, so the constraint $R_{\log}(T)\le D$ is a constraint $I(T;Y)\ge H(Y)-D$, and \eqref{eq:irdf} becomes the information bottleneck curve \citep{tishby2000ib}. The minimal sufficient complexity $C^*_{\log}=H(\eta(X))$ is therefore the full-relevance end point of the IB curve, and the G-LMC under log-loss is the minimal sufficient statistic $\eta(X)$. (With past and future of a process in place of $X$ and $Y$, $C^*_{\log}$ is also the statistical complexity of computational mechanics; \citealp{crutchfield1989inferring}.) For other losses, \eqref{eq:irdf} is a loss-specific bottleneck. The 0-1 version is the relevant one for classification: in Example~\ref{ex:running} it ends at 1 bit, not 2, because a classifier does not need to know the noise level.

\subsection{The original LMC and why we work at the population level}\label{sec:lmc}

We can now say precisely what was wrong with the original definition, and why its generalization above avoids the problem. The LMC papers called a deterministic output $F=f(X)$ a \emph{lossless maximal compression} if $I(F;X)=I(F;Y)=I(X;Y)$ \citep[Def.~4 and Prop.~1]{nikolaou2020margin}. In words: the output keeps all the information that $X$ has about $Y$, and retains nothing else about $X$. When the labels are noiseless this is the same as being a G-LMC under log-loss. When they are noisy it is not, and the requirement can almost never be met.

\begin{proposition}[The original LMC rarely exists]\label{prop:gk}
A deterministic $F$ with $I(F;X)=I(F;Y)=I(X;Y)$ exists if and only if the G\'acs--K\"orner common information of $X$ and $Y$ \citep{gacs1973common} equals $I(X;Y)$. For binary labels, this holds if and only if $H(Y\mid X)=0$ or $I(X;Y)=0$.
\end{proposition}

So for binary classification with informative features, the original LMC exists only if there is no label noise at all. In Example~\ref{ex:running} with $\varepsilon>0$, no LMC exists, while the G-LMC always exists (it is $b^*(X)$ under 0-1 loss). The reason is that requiring $I(F;X)=I(F;Y)$ forces $H(F\mid Y)=0$, so $F$ would have to be a function of the noisy label as well as of the input.

There is a second, empirical, problem. On a training sample with distinct inputs, the empirical label noise is zero: each input appears once, with one label. Any classifier that interpolates the sample then satisfies the empirical version of the LMC equations exactly (Appendix~\ref{sec:degeneracy}). This is why, in Table~\ref{tab:pilot}, all five interpolators have the same legacy score of $0.99$, despite their very different test behaviour. \emph{The lesson is that information quantities computed on the training sample cannot distinguish interpolators.} For this reason every quantity in the rest of this paper is defined either at the population level, as in this section, or through the learner's behaviour over independent training samples, as in the next. Appendix~\ref{sec:erratum} lists the claims of the earlier papers that this affects and those that survive.

%% file: sections/s4_storage_expression.tex
\section{What a learner stores and what it expresses}\label{sec:storage-expression}

In the last section we described the ideal model: an output that carries the minimal sufficient complexity $C^*_\ell=H(a^*(X))$ and no excess. We will now bring learning into the picture. A learner sees a finite noisy sample, and an interpolating learner reproduces every noisy label in it. We ask two questions. How much information about the training labels must such a learner keep (Section~\ref{sec:storage})? And how much of it reaches its predictions on new inputs (Section~\ref{sec:expression})? We will see that the first quantity is the same for every interpolator, while the second controls excess risk.

Recall from Section~\ref{sec:setting} that $W=A(S,\xi)$ is the trained model, which is random through both the sample $S$ and the seed $\xi$, and that $h_W$ is the classifier it induces. Let $\hat\varepsilon=\frac1n\sum_i\ind\{h_W(X_i)\ne Y_i\}$ be its training error and $\bar\varepsilon=\E\hat\varepsilon$ the expected training error. Let $\phi_K(e)=h(e)+e\log_2(K-1)$, the function that appears in Fano's inequality.

\subsection{Storing and expressing, in plain terms}\label{sec:plain}

Before the formal definitions, let us explain the two words informally. Both describe what an interpolating learner does with the noisy labels it was trained on. \emph{Storing} a label means that the trained model keeps it. \emph{Expressing} it means that the label still affects the model's predictions at \emph{new} inputs. Both terms, in this sense, are introduced in this paper (Table~\ref{tab:terms}).

\paragraph{Storage: what the model keeps from its training labels.} Suppose we could look inside the trained model and reconstruct, from it and the training inputs, which training points were labelled $1$ and which $0$, including the mislabelled ones. Then the model has stored those labels. An interpolator reproduces every training label, wrong ones included, so all of them can be read back off. Theorem~\ref{thm:storage} makes this exact: every interpolator stores $n\,H(Y\mid X)$ bits, all of the label noise in its sample. In the running example (Example~\ref{ex:running}) with $n=4096$, this is about $2{,}437$ bits, the same for every learner in Table~\ref{tab:pilot}. So storage cannot tell a good interpolator from a bad one, and neither can generalization bounds built on the information in the weights.

\paragraph{Expression: what reaches predictions at new inputs.} Now fix one new input $x$ and imagine retraining the learner many times, each time on a fresh sample, recording its prediction at $x$. If the prediction is always the same, the particular sample, noisy labels included, made no difference at $x$: nothing was expressed there. If the prediction changes with the sample, the prediction at $x$ carries information about which noisy labels the learner happened to see. Expressed information $V(x)$ (Definition~\ref{def:V}) measures exactly this.

\paragraph{Why the distinction matters.} A learner can store everything and express almost nothing. The memorize-and-smooth learner of Proposition~\ref{prop:memsmooth} returns the memorized label on a training input, but on a new input it predicts the majority label among training points with the same informative feature, which individual noisy labels barely move. Since new inputs almost never coincide with training inputs, the memorized labels are almost never used: in Table~\ref{tab:pilot} it expresses $0.007$ bits and has excess risk $0.001$. The 1-NN rule stores exactly as much, but predicts with the label of the nearest training point, so each prediction is effectively a fresh noisy label; it expresses all the noise it stores ($V\to H(Y\mid X)=0.595$ bits; Proposition~\ref{prop:knn}) and its excess risk stays near $0.12$. Excess risk is controlled by expression, not by storage (Theorem~\ref{thm:decomp}). Benign overfitting means that the noise is stored but not expressed.

\paragraph{An analogy.} Consider students who memorize an answer key that contains some typos. \emph{Storage} is how much of the key, typos included, is in a student's memory; every student who can recite the key perfectly stores the same amount. \emph{Expression} is whether those typos appear when the student answers \emph{new} questions. One student has memorized the key but answers new questions from an understanding of the topic, so the typos never surface (benign). Another answers each new question by recalling the closest memorized entry and copying it, typo and all (tempered). Grading the students on the memorized questions cannot separate them, since both answer all of them ``correctly''. The difference appears only on new questions, and that is what $V$ measures, without needing the correct answers to those questions.

\subsection{Storage: interpolators memorize all of the label noise}\label{sec:storage}

\begin{theorem}[Storage of label noise; cf.\ \citealp{harutyunyan2020label}, Thm.~2.1]\label{thm:storage}
For every learner,
\[
I(W;Y^n\mid X^n)\ \ge\ n\big[H(Y\mid X)-\phi_K(\bar\varepsilon)\big].
\]
If the learner interpolates, then $I(W;Y^n\mid X^n)=n\,H(Y\mid X)$ exactly, and so $I(W;S)\ge n\,H(Y\mid X)$.
\end{theorem}

The quantity $I(W;Y^n\mid X^n)$ is the information that the trained model holds about the training labels beyond what the training inputs already say about them. It is the natural measure of how much the model has \emph{memorized}. The theorem says that a model that fits its labels well must have memorized almost all of their noise, and that an interpolator has memorized exactly all of it: $H(Y\mid X)$ bits per example. In the running example (Example~\ref{ex:running}) with $n=4096$, every interpolator in Table~\ref{tab:pilot} stores exactly $4096\times0.595\approx2437$ bits of label noise, regardless of how well it predicts. The inequality is a rearrangement of Theorem~2.1 of \citet{harutyunyan2020label}, who use it to bound the expected number of training errors. What we add is the exact equality for interpolators and the consequence below; Appendix~\ref{sec:proofs-storage} also splits the stored information into a task part and a noise part.

An immediate consequence concerns information-theoretic generalization bounds. These bound the expected generalization gap by $I(W;S)$; for example, for a $\sigma$-sub-Gaussian loss, $|\E\,\mathrm{gap}|\le\sqrt{2\sigma^2I(W;S)/n}$ \citep{xu2017information}.

\begin{corollary}[Mutual-information bounds cannot separate interpolators]\label{cor:xr}
For every interpolator and every $n$, the bound of \citet{xu2017information} is at least $\sqrt{2\sigma^2H(Y\mid X)}$ (with $H$ in nats). It therefore does not vanish, and it cannot distinguish a benign interpolator from a tempered one.
\end{corollary}

In plain terms: both a benign and a tempered interpolator have a generalization gap that tends to the Bayes risk (their training error is zero and their test error tends to $R^*$ or above), and the bound is equally loose for both. What separates them is their \emph{excess} risk, which the bound does not see. This is not a defect of one particular bound; it is a consequence of storage being pinned at $nH(Y\mid X)$ (see also \citealp{livni2023lower,haghifam2023limitations}). To separate interpolators, we need a quantity that looks at the predictions, not at the weights.

\paragraph{Conditional mutual information.} \citet{steinke2020cmi} proposed a refinement that avoids some of the problems of $I(W;S)$. Draw a \emph{supersample} $\tilde Z\in(\cX\times\cY)^{n\times2}$ of $2n$ i.i.d.\ examples arranged in $n$ pairs, and a uniformly random selector $U\in\{0,1\}^n$ that picks one example from each pair to form the training sample $S=\tilde Z_U$. The \emph{conditional mutual information} of the learner is $\mathrm{CMI}=I(W;U\mid\tilde Z)$: how much the trained model reveals about which example of each pair it was trained on. Unlike $I(W;S)$, it is always finite, $\mathrm{CMI}\le n\ln2$, and for losses in $[0,1]$ it gives $|\E\,\mathrm{gap}|\le\sqrt{2\,\mathrm{CMI}/n}$ \citep[Thm.~2]{steinke2020cmi}. Its behaviour on interpolators is nevertheless the same in kind.

\begin{corollary}[CMI cannot separate interpolators either]\label{cor:cmi}
For every interpolator, $\mathrm{CMI}\ \ge\ \frac n2\big(\E R(h_W)\big)^2\ \ge\ \frac n2(R^*)^2$ (in nats). So the CMI of a benign interpolator grows linearly in $n$, like that of a tempered one, and the CMI bound on the gap is at least $R^*$ for both.
\end{corollary}

In plain terms, the CMI framework conditions away the randomness of the data and keeps only the randomness of the selection, which makes it finite and often much tighter. But it still bounds the generalization \emph{gap}, and for an interpolator the gap is its whole test error, which is at least the Bayes risk. In the running example with $\varepsilon=0.2$, every interpolator has $\mathrm{CMI}\ge0.011\,n$ nats. We will see in Section~\ref{sec:expression} that the CMI construction is still useful: applied to predictions at fresh inputs rather than to the weights, it gives one of the components of expressed information (Proposition~\ref{prop:supersample}).

\subsection{Expression: the information that reaches predictions}\label{sec:expression}

Let us now introduce the central quantity of the paper. Fix an input $x$, and imagine retraining the learner many times on fresh samples (and with fresh seeds). Each time we record the prediction $h_W(x)$. Let $q(\cdot\mid x)$ be the distribution of these predictions over $[K]$:
\[
q(k\mid x)=\P_{S,\xi}\big(h_W(x)=k\big).
\]

\begin{definition}[Expressed information]\label{def:V}
The \emph{expressed information} of a learner at $x$ is the entropy of its prediction distribution,
\[
V(x)\;=\;H\big(q(\cdot\mid x)\big)\;=\;I\big(h_W(x)\,;\,S,\xi\big).
\]
Its average over a fresh test input $X'\sim P_X$, independent of the sample, is $\E\,V(X')=I\big(h_W(X')\,;\,S,\xi\mid X'\big)$. The \emph{modal prediction} $m(x)=\argmax_kq(k\mid x)$ is the most frequent prediction at $x$.
\end{definition}

The two expressions for $V(x)$ are equal because $h_W(x)$ is a function of $(S,\xi)$. Let us explain both readings in plain terms.
\begin{itemize}[leftmargin=*]
\item As an \emph{entropy}, $V(x)$ measures how unpredictable the prediction at $x$ is before we know which sample the learner will be trained on. It is $0$ if every sample leads to the same prediction, and it is at most $\log_2K$ bits.
\item As a \emph{mutual information}, $V(x)$ measures how much the prediction at $x$ reveals about the particular training sample (and seed). If the learner has memorized a noisy label near $x$, its prediction at $x$ will reveal that label, and $V(x)$ will be large. If it has smoothed the label away, the prediction will not depend on it, and $V(x)$ will be small.
\end{itemize}
So $V$ measures the part of what was stored that is \emph{expressed} at test time.

Expressed information can also be read in the supersample construction of \citet{steinke2020cmi} described in Section~\ref{sec:storage}.

\begin{proposition}[Expressed information in the supersample construction]\label{prop:supersample}
Let $S=\tilde Z_U$ as in Section~\ref{sec:storage}, with $\xi$ independent of $(\tilde Z,U)$. For every $x$,
\[
V(x)\;=\;\underbrace{I\big(h_W(x);\tilde Z\big)}_{\text{which pool}}\;+\;\underbrace{I\big(h_W(x);U\mid\tilde Z\big)}_{\text{which half of the pool}}\;+\;\underbrace{I\big(h_W(x);\xi\mid\tilde Z,U\big)}_{\text{which seed}}.
\]
\end{proposition}

So the CMI-style quantity at a fresh input, $I(h_W(x);U\mid\tilde Z)$, is the part of the expressed information that reflects which half of a fixed pool of data the learner saw, and it is at most $V(x)$. It is the prediction-level, fresh-input counterpart of $\mathrm{CMI}=I(W;U\mid\tilde Z)$: where CMI asks how much the \emph{weights} reveal about the selection, this term asks how much a \emph{prediction at a new input} reveals about it. Our estimator from two disjoint halves (Section~\ref{sec:sel:est}) is exactly this construction for one draw of $U$: one copy is trained on $\tilde Z_U$ and the other on $\tilde Z_{1-U}$, and their disagreement on fresh inputs is observed. The proof is in Appendix~\ref{sec:proofs-expr}.

The modal prediction $m(x)$ is what the learner predicts ``on average''; it is the prediction of an infinitely large bagged ensemble of independently trained copies of the learner. Expressed information is the hard-prediction, Shannon-entropy counterpart of the 0-1 ``variance'' of \citet{kohavi1996bias} and \citet{domingos2000unified}; we discuss the relation in Section~\ref{sec:related}.

The next theorem is the main result of this section. It says that excess risk is governed by two things: how much a learner expresses ($V$), and whether its modal prediction is right. Let $\Delta(x)=\eta_{b^*(x)}(x)-\min_k\eta_k(x)$ and $\gamma(x)=\eta_{b^*(x)}(x)-\max_{k\ne b^*(x)}\eta_k(x)$ be the largest and smallest gaps between the probability of the Bayes class and that of another class. For binary labels both are $|2\eta_1(x)-1|$, the \emph{margin} of the problem at $x$.\footnote{This margin is a property of the data distribution, not of a trained model: it measures how far the probability of the more likely class at $x$ is from $\frac12$, i.e.\ how predictable the label at $x$ is even in principle. It should not be confused with the margin $y\,f(x)$ of a classifier's real-valued output, as used for boosting and support vector machines and in the LMC papers.}

\begin{theorem}[Information decomposition of excess risk]\label{thm:decomp}
For every learner,
\begin{align*}
\E\,R(h_W)-R^*\ &\le\ \tfrac12\,\E\big[\Delta(X)\,V(X)\big]\;+\;\E\big[\Delta(X)\,\ind\{m(X)\ne b^*(X)\}\big],\\
\E\,R(h_W)-R^*\ &\ge\ \E\big[\gamma(X)\,\phi_K^{-1}\big(V(X)\big)\,\ind\{m(X)=b^*(X)\}\big]\;+\;\tfrac12\,\E\big[\gamma(X)\,\ind\{m(X)\ne b^*(X)\}\big],
\end{align*}
where $\phi_K^{-1}$ is the inverse of $\phi_K$ on $[0,1-1/K]$. For binary labels the decomposition is exact: with $\pi(x)=h^{-1}(V(x))\in[0,\frac12]$,
\begin{equation}\label{eq:exact}
\E\,R(h_W)-R^*\;=\;\E\Big[\,|2\eta_1(X)-1|\,\Big(\pi(X)\,\ind\{m(X)=b^*(X)\}+\big(1-\pi(X)\big)\,\ind\{m(X)\ne b^*(X)\}\Big)\Big].
\end{equation}
\end{theorem}

The proof (Appendix~\ref{sec:proofs-expr}) combines the exact identity $\E R(h_W)-R^*=\E[\sum_kq(k\mid X)(\eta_{b^*}(X)-\eta_k(X))]$ with a two-sided bound on the probability mass off the mode in terms of entropy (Lemma~\ref{lem:mode}; cf.\ \citealp{feder1994relations}). In plain terms, the theorem says that a learner can lose accuracy in two ways.
\begin{itemize}[leftmargin=*]
\item \emph{Through expression.} At inputs where its modal prediction is correct, a learner errs only when it departs from its mode, which it does with a probability controlled by $V(x)$ from above and below. In the binary case, $\pi(x)=h^{-1}(V(x))$ is exactly the probability of departing from the mode.
\item \emph{Through modal bias.} At inputs where its modal prediction is wrong, the learner errs at least half of the time, whatever its expressed information.
\end{itemize}
Each loss is weighted by the margin of the problem at $x$: errors cost little where the classes are nearly balanced.

\begin{corollary}[Consistency means vanishing expression and vanishing modal bias]\label{cor:consistency}
Let $(A_n)$ be a sequence of learners and assume $\gamma(X)>0$ almost surely. Then $\E R(h_{W_n})\to R^*$ if and only if $\E V_n(X')\to0$ and $\P(m_n(X)\ne b^*(X))\to0$.
\end{corollary}

\subsection{Output entropy splits into complexity and expression}\label{sec:split}

We now connect expression back to the complexities of Section~\ref{sec:sufficiency}. There, a model was a fixed function. A learner's output at a fresh input, $F=h_W(X')$, is a \emph{random} function of $X'$: it depends on $X'$, and also on the sample and the seed. Since $(S,\xi)$ is independent of the test pair, $F$ is a stochastic representation of $X'$ in the sense of Section~\ref{sec:suff}.

\begin{proposition}[Complexity plus expression]\label{prop:split}
For $F=h_W(X')$,
\[
H(F)\;=\;I(F;X')\;+\;\E\,V(X').
\]
\end{proposition}
\begin{proof}
$H(F)=I(F;X')+H(F\mid X')$, and $H(F\mid X'=x)=H(q(\cdot\mid x))=V(x)$.
\end{proof}

So the total entropy of a learner's predictions on new inputs has two parts. The first, $I(F;X')$, is \emph{complexity}: variation of the prediction that is explained by the input, as in Section~\ref{sec:sufficiency}. The second, $\E V$, is \emph{expression}: variation that is explained only by the training sample. The ideal output, the G-LMC $b^*(X')$, has complexity $C^*_{01}=H(b^*(X'))$ and no expression. The next result says that a consistent learner approaches exactly this.

\begin{corollary}[Consistent learners converge to the G-LMC]\label{cor:toglmc}
Under the assumption of Corollary~\ref{cor:consistency}, if $\E R(h_{W_n})\to R^*$ then $\P(h_{W_n}(X')\ne b^*(X'))\to0$, and
\[
H\big(h_{W_n}(X')\big)\to C^*_{01},\qquad I\big(h_{W_n}(X');X'\big)\to C^*_{01},\qquad \E V_n(X')\to0.
\]
In particular, the excess complexity of the learner's output over the minimal sufficient complexity vanishes, even if the learner interpolates and stores $n\,H(Y\mid X)$ bits of label noise.
\end{corollary}

This corollary gives the picture of benign overfitting that we develop in the next section. Storage is pinned at its maximum: an interpolator memorizes all the label noise in its sample (Theorem~\ref{thm:storage}). But the learner's \emph{predictions} on new inputs can still converge to the minimal sufficient representation. Benign overfitting means that what is stored is not expressed.

\begin{example}[Running example, continued]\label{ex:split}
In Example~\ref{ex:running}, $H(Y)=1$ bit. Section~\ref{sec:knn} shows that, as $n\to\infty$, the prediction of the 1-nearest-neighbour rule at a fresh input behaves like a fresh noisy label: $\E V\to H(Y\mid X)=0.595$ bits and $I(F;X')\to I(X;Y)=0.405$ bits, which add up to $H(F)=1$ bit. A benign interpolator, such as the memorize-and-smooth rule of Proposition~\ref{prop:memsmooth}, instead has $\E V\to0$ and $I(F;X')\to1$ bit, the minimal sufficient complexity. Both store the same $0.595$ bits per training example. In Table~\ref{tab:pilot}, at $n=4096$, the 1-NN rule expresses $0.642$ bits and the memorize-and-smooth rule $0.007$ bits.
\end{example}

\subsection{Beyond classification error}\label{sec:sq}

The decomposition of Theorem~\ref{thm:decomp} is for 0-1 loss, but the idea carries over to other losses. We briefly state the squared-loss version, which links expressed information to the classical variance. Let $Y\in\mathbb R$, let $f_W(x)$ be the learner's real-valued prediction, and let $v(x)$ and $\bar f(x)$ be its variance and mean over $(S,\xi)$. A continuous prediction carries infinite information, so we add a small amount of independent Gaussian noise before measuring it: let $V_\sigma(x)=I(f_W(x)+\sigma Z\,;\,S,\xi)$ with $Z\sim N(0,1)$, in nats.

\begin{theorem}[Variance is low-resolution expressed information]\label{thm:sq}
For every $x$, $V_\sigma(x)\le\frac12\log(1+v(x)/\sigma^2)$ and $\lim_{\sigma\to\infty}2\sigma^2V_\sigma(x)=v(x)$. Consequently
\[
\E\,R(f_W)-R^*\;=\;\E\big(\bar f(X)-\E[Y\mid X]\big)^2\;+\;\lim_{\sigma\to\infty}2\sigma^2\,\E\,V_\sigma(X).
\]
\end{theorem}

So under squared loss, the familiar variance term \emph{is} expressed information, measured at low resolution; the proof uses the I-MMSE relation of \citet{guo2005immse}. For log-loss with probabilistic outputs, the analogous result is the bias--variance decomposition of \citet{heskes1998bias}, whose variance term is again an expressed-information quantity (Appendix~\ref{sec:logloss}). Appendix~\ref{sec:regression} works out the regression case further: every exact interpolator of real-valued labels stores infinitely much information about them (and $n[h(Y\mid X)+\log(1/\Delta)]$ bits at resolution $\Delta$), and the label-free rule ``least variance between two half-sample fits'' selects well when the candidates have similar bias, but not otherwise, because interpolation does not keep bias small in regression.

\subsection{Relation to bias--variance decompositions and to disagreement}\label{sec:bvd}

Expressed information is not the first quantity to measure how much a learner's predictions vary across training samples. Bias--variance decompositions for 0-1 loss \citep{kohavi1996bias,domingos2000unified,james2003variance} and, more recently, the disagreement between independently trained models \citep{jiang2022disagreement,nakkiran2020distributional,baek2022agreement} do the same. In this subsection we make the relations exact. We will see that these quantities are all functions of the same prediction distribution $q(\cdot\mid x)$ as $V$, that they are equivalent to $V$ up to explicit monotone bounds, and that Theorem~\ref{thm:decomp} tells us what each of them measures.

Let us first recall the classical quantities, all computed at a fixed input $x$ from the prediction distribution $q=q(\cdot\mid x)$ and the modal prediction $m=m(x)$ (Domingos' ``main prediction'').
\begin{itemize}[leftmargin=*]
\item \emph{Domingos' variance} is the probability of departing from the main prediction, $\pi_D(x)=1-q(m\mid x)$. His \emph{bias} is $B(x)=\ind\{m(x)\ne b^*(x)\}$, which we called modal bias.
\item \emph{Kohavi and Wolpert's variance} is the Gini impurity of the predictions, $\sigma^2_{KW}(x)=\frac12\big(1-\sum_kq(k\mid x)^2\big)$.
\item The \emph{disagreement} of two independently trained copies is $\mathrm{Dis}(x)=\P\big(h_{W_1}(x)\ne h_{W_2}(x)\big)=1-\sum_kq(k\mid x)^2=2\sigma^2_{KW}(x)$.
\item \emph{James' variance effect} is the increase in expected loss caused by departing from the main prediction, $\mathrm{VE}(x)=\E\big[\ind\{h_W(x)\ne Y\}\big]-\P(m(x)\ne Y)$, and his \emph{systematic effect} is $\mathrm{SE}(x)=\P(m(x)\ne Y)-\P(b^*(x)\ne Y)$.
\end{itemize}
Note that the disagreement of two copies is exactly twice the Kohavi--Wolpert variance; this simple identity links the disagreement literature to the bias--variance literature.

\begin{proposition}[A dictionary of variances]\label{prop:dictionary}
At every $x$,
\[
\phi_K^{-1}\big(V(x)\big)\ \le\ \pi_D(x)\ \le\ \mathrm{Dis}(x)=2\sigma^2_{KW}(x)\ \le\ 2\pi_D(x)\ \le\ V(x).
\]
For $K=2$ the four quantities are increasing functions of one another: $\pi_D=h^{-1}(V)$ and $\mathrm{Dis}=2\pi_D(1-\pi_D)$. Moreover, $\E R(h_W)-R^*=\E[\mathrm{SE}(X)]+\E[\mathrm{VE}(X)]$, and for $K=2$
\[
\mathrm{SE}(x)+\mathrm{VE}(x)=|2\eta_1(x)-1|\Big(B(x)+\big(1-2B(x)\big)\,\pi_D(x)\Big),
\]
which is the exact decomposition \eqref{eq:exact}.
\end{proposition}

In plain terms: all these ``variances'' measure the same thing, the spread of the prediction distribution, on different scales. What distinguishes $V$ is that it is also a mutual information, $I(h_W(x);S,\xi)$. This is what links variance to storage (Theorem~\ref{thm:storage}) and to the complexity of the output (Proposition~\ref{prop:split}), and what gives the matching upper and lower bounds of Theorem~\ref{thm:decomp} for any number of classes. The last identity is Domingos' observation that ``unbiased'' variance ($B=0$) increases error, while ``biased'' variance ($B=1$) decreases it, now weighted by the margin. Proposition~\ref{prop:ensemble} is its consequence for averaging.

\paragraph{What does disagreement measure?} \citet{jiang2022disagreement} observed that the expected disagreement of two copies of a network, trained with different seeds, is close to its test error, and proved that this \emph{generalization disagreement equality} (GDE) holds when the ensemble is class-wise calibrated. Theorem~\ref{thm:decomp} explains when this happens and when it cannot.

\begin{proposition}[Disagreement bounds excess risk from both sides, not risk]\label{prop:disagreement-main}
Let $\mathrm{Dis}=\E\,\mathrm{Dis}(X)$ for copies trained on independent samples (and seeds).
\begin{enumerate}[label=(\alph*),leftmargin=*]
\item If $q(\cdot\mid x)=\eta(x)$ for every $x$, i.e.\ the learner's predictions are distributed like a fresh label, then $\E R(h_W)=\mathrm{Dis}$ exactly.
\item For $K=2$, if the modal predictions are correct ($m=b^*$ a.s.), then
\[
\E R(h_W)-R^*\ \le\ \E\,\pi_D(X)\ \le\ \mathrm{Dis}\ \le\ 2\,\E\,\pi_D(X)\ \le\ \E V(X),
\]
and if moreover $|2\eta_1-1|\ge\gamma_0>0$ a.s., then $\E R(h_W)-R^*\ge\frac{\gamma_0}{2}\mathrm{Dis}$.
\item For the 1-NN rule, $\mathrm{Dis}\to2\E[\eta_1(X)(1-\eta_1(X))]=\lim\E R(h_W)$: the GDE holds exactly in the limit. For any consistent sequence of learners, $\mathrm{Dis}\to0$ while $\E R(h_W)\to R^*$, so the GDE fails by exactly the Bayes risk.
\end{enumerate}
\end{proposition}

Let us restate the proposition in plain terms. Disagreement, like $V$, measures \emph{excess} risk (part b). It equals the total risk only when the learner's predictions behave like samples of the label distribution (part a), which is the ``distributional generalization'' of interpolating classifiers described by \citet{nakkiran2020distributional}. Tempered interpolators such as 1-NN behave this way (part c; Example~\ref{ex:split}), so for them the GDE holds. A benign interpolator under label noise does not: its predictions concentrate on $b^*$, its disagreement vanishes, and its test error stays at $R^*>0$. In the running example at $n=4096$ (Section~\ref{sec:exp:synth}, Table~\ref{tab:dis}), the disagreement of 1-NN is $0.30$ against a test error of $0.27$, while the random forest has disagreement $0.014$ against a test error of $0.154$.

Two further remarks connect to recent work. First, the randomness matters. $V=I(h_W(x);S,\xi)=I(h_W(x);S)+I(h_W(x);\xi\mid S)$ splits into a sample part and a seed part, much as the fine-grained variance decomposition of \citet{adlam2020understanding} splits variance into data, initialization and label-noise parts. The GDE of \citet{jiang2022disagreement} uses copies trained on the \emph{same} data with different seeds, so it sees only the seed part. For a deterministic learner such as 1-NN that part is zero, even though the 1-NN rule expresses $H(Y\mid X)$ bits through the sample. Second, agreement between classifiers is also used to estimate accuracy without labels \citep{platanios2016bayesian} and to predict performance under distribution shift \citep{baek2022agreement}. These methods typically rely on independence or calibration assumptions. The consensus criterion of Section~\ref{sec:sel:beyond} uses agreement differently: it estimates each candidate's \emph{modal bias} by its disagreement with the consensus, with the explicit guarantee $|D_j-\beta_j|\le\delta$ and no independence assumption (Appendix~\ref{sec:consensus}).

%% file: sections/s5_benign.tex
\section{Benign overfitting as vanishing expression}\label{sec:benign}

In the last section we saw that every interpolator stores all the label noise in its sample (Theorem~\ref{thm:storage}), and that excess risk is controlled by expressed information and modal bias (Theorem~\ref{thm:decomp}). We will now use these two results to explain when interpolation is harmless. We first state the resulting taxonomy, and then work it out for the main mechanisms studied in the benign-overfitting literature: memorizing while smoothing, local averaging, ensembling, and minimum-norm interpolation. Proofs are in Appendix~\ref{sec:proofs-benign}.

\subsection{An information-theoretic taxonomy}\label{sec:taxonomy}

\citet{mallinar2022taxonomy} classify interpolating learners as benign, tempered or catastrophic according to their limiting excess risk. Combining Theorem~\ref{thm:storage} with Corollary~\ref{cor:consistency} gives an information-theoretic reading of the same taxonomy. As $n\to\infty$, an interpolator always stores $n\,H(Y\mid X)$ bits. It overfits
\begin{itemize}[leftmargin=*]
\item \emph{benignly} if $\E V_n\to0$ and its modal bias vanishes: it stores all of the noise and expresses none of it;
\item \emph{temperedly} if $\E V_n\to c$ for some $c\in(0,\log_2K]$ while its modal bias vanishes: it expresses a fixed fraction of what it stores, and its excess risk tends to a constant set by $c$ through Theorem~\ref{thm:decomp};
\item \emph{catastrophically} if its modal bias does not vanish: even its average prediction is wrong on a set of positive probability.
\end{itemize}
The important point is that the dividing line between benign and tempered overfitting is a property of predictions at fresh inputs, $V$, and not of the trained model's memory, which is the same for both.

\subsection{Memorizing without expressing}\label{sec:memsmooth}

Our first example shows that the two extremes, maximal storage and minimal expression, can be attained at the same time. It is deliberately simple, and it is built for the running example.

\begin{proposition}[A benign interpolator]\label{prop:memsmooth}
Consider the setting of Example~\ref{ex:running}, with $d$ irrelevant bits instead of 16 and $\eta$ depending on $Z$ only, with $\eta_1(z)\ne\frac12$. The \emph{memorize-and-smooth} learner predicts the training label at $x$ if $x$ appears in the training sample, and otherwise predicts the majority training label among examples with the same $Z$. Let $n\to\infty$ and $d=d_n$ with $n^22^{-d_n}\to0$. Then the learner interpolates with probability tending to one, it stores $n\big(H(Y\mid X)-o(1)\big)$ bits of label noise, and $\E V_n\to0$ and $\E R\to R^*$.
\end{proposition}

In words, the learner memorizes every training label exactly, but because fresh inputs almost never coincide with training inputs, the memorized labels are almost never used for prediction. All predictions on new inputs come from a smoothed majority vote, which does not depend on individual noisy labels. In Table~\ref{tab:pilot} this learner expresses $0.007$ bits and has excess risk $0.001$. The random forest behaves in a similar way: each tree memorizes, but the average over trees smooths the memorized labels away from the training points ($V=0.045$).

\subsection{Nearest neighbours: from tempered to benign}\label{sec:knn}

The $k$-nearest-neighbour ($k$-NN) rule is the classical example of a learner whose overfitting can be tuned. With $k=1$ it interpolates and is tempered; with $k$ growing it is consistent. Expressed information describes the whole transition.

\begin{proposition}[Nearest neighbours]\label{prop:knn}
Let $K=2$, $\cX\subseteq\mathbb R^d$, let $\eta_1$ be continuous, and let $k$ be fixed and odd. Let $\beta_k(t)=\P(\mathrm{Bin}(k,t)>k/2)$. As $n\to\infty$, for almost every $x$,
\[
V_n(x)\ \to\ h\Big(\min\big\{\beta_k(\eta_1(x)),\,1-\beta_k(\eta_1(x))\big\}\Big)\ \le\ h\Big(\min\big\{\tfrac12,\ e^{-2k(\eta_1(x)-1/2)^2}\big\}\Big),
\]
and the modal prediction converges to $b^*$. In particular, for the 1-NN rule, $\E V_n\to H(Y\mid X)$, and Theorem~\ref{thm:decomp} gives $\E R-R^*\to2\E[\eta_1(1-\eta_1)]-R^*$, the classical limit of \citet{cover1967nearest}.
\end{proposition}

Let us restate the result for $k=1$ in plain terms. Asymptotically, the 1-NN rule expresses \emph{exactly} the label noise: its prediction at a new input is the label of a nearby training point, which is a fresh noisy draw from $\eta(x)$. So the 1-NN rule passes on all the noise it stores, at the rate of $H(Y\mid X)$ bits per prediction. In the running example this is $0.595$ bits, and at $n=4096$ we measure $0.642$ bits (Table~\ref{tab:pilot}; our checks in Appendix~\ref{sec:mech-checks} match the limits of Proposition~\ref{prop:knn} to within $0.003$ bits). The Cover--Hart excess risk of 1-NN thus follows from an information-theoretic statement. For larger fixed $k$, the expressed information decays exponentially in $k$ wherever the margin is bounded away from zero; the rule remains tempered, with a smaller constant.

When $k_n\to\infty$ and $k_n/n\to0$, the rule is consistent, so $\E V_n\to0$ by Corollary~\ref{cor:consistency}. Under the standard smoothness and margin conditions \citep{audibert2007fast} we can also give a rate.

\begin{proposition}[Rate of vanishing expression]\label{prop:knn-rate}
Let $X$ have a density on $[0,1]^d$ bounded above and below, let $\eta_1$ be Lipschitz, and assume the margin condition $\P(0<|\eta_1(X)-\frac12|\le t)\le C_0t^\alpha$. With $k_n\asymp n^{2/(d+2)}$, the $k_n$-NN rule has $\E V_n=O\big(n^{-\alpha/(d+2)}\big)$.
\end{proposition}

The rate is half the exponent of the familiar excess-risk rate $n^{-(1+\alpha)/(d+2)}$; the difference is the margin weighting in Theorem~\ref{thm:decomp}. In simulation with $d=\alpha=1$ we measure slopes of $-0.32$ for $\E V_n$ and $-0.68$ for excess risk, against the predicted $-\frac13$ and $-\frac23$ (Appendix~\ref{sec:mech-checks}).

\subsection{Averaging removes expression but keeps modal bias}\label{sec:averaging}

Ensembling is the most common way to make interpolators benign. Theorem~\ref{thm:decomp} says exactly what it does.

\begin{proposition}[Averaging]\label{prop:ensemble}
Let $g_B$ be the majority vote of $B$ independently trained copies of a binary learner ($B$ odd). Then $g_B$ has the same modal prediction as one copy, and its expressed information satisfies $V_{g_B}(x)\le h\big(\min\{\frac12,e^{-2B(1/2-\pi(x))^2}\}\big)$, where $\pi(x)=h^{-1}(V(x))$ is the single learner's probability of departing from its mode.
\end{proposition}

So averaging removes expressed information exponentially fast in $B$, but it leaves the modal bias untouched. It reduces risk where the mode is right and locks in the error where the mode is wrong. This is an information-theoretic form of the observation that bagging helps unstable learners and does not help biased ones \citep{breiman1996bagging}. Appendix~\ref{sec:proofs-benign} gives the exact limiting change in risk, and a related result for stable learners: replace-one stability $\beta_n$ at $x$ implies $V_n(x)\le h(\min\{\frac12,n\beta_n(x)\})$, so uniformly stable learners \citep{bousquet2002stability} express little.

\subsection{Minimum-norm interpolation and double descent}\label{sec:ridgeless}

The best-understood benign interpolator is minimum-norm least squares in high dimensions \citep{bartlett2020benign,hastie2022surprises}. Using the squared-loss version of expressed information (Theorem~\ref{thm:sq}), we can compute $V$ exactly for it.

\begin{proposition}[Minimum-norm least squares]\label{prop:ridgeless}
Let $y=X\beta+\varepsilon$ with a fixed design $X\in\mathbb R^{n\times p}$, $p\ge n$, noise $\varepsilon\sim N(0,\sigma_\varepsilon^2I)$, and estimator $\hat\beta=X^+y$. Let $x'$ be a test input with covariance $\Sigma$. With the label noise as the only source of randomness,
\[
V_\sigma(x')=\tfrac12\log\Big(1+\tfrac{\sigma_\varepsilon^2}{\sigma^2}\big\|(X^+)^\top x'\big\|^2\Big),\qquad \lim_{\sigma\to\infty}2\sigma^2\,\E V_\sigma(x')=\sigma_\varepsilon^2\operatorname{tr}(C),
\]
where $C=(XX^\top)^{-1}X\Sigma X^\top(XX^\top)^{-1}$.
\end{proposition}

The limit $\sigma_\varepsilon^2\operatorname{tr}(C)$ is exactly the variance term in the analysis of \citet{bartlett2020benign}. So, in their setting, minimum-norm interpolation is benign precisely when its label-noise expressed information vanishes, which by their theorem happens exactly when the covariance $\Sigma$ satisfies their effective-rank conditions. Their conditions can be read as conditions under which noise that is stored in many unimportant directions is not expressed in the important ones.

The same view explains \emph{double descent} \citep{belkin2019reconciling}. As the number of random features $p$ grows past the number of examples $n$, the test error of the minimum-norm fit first peaks at $p\approx n$ and then falls again. At the peak the fit is forced through every noisy label with the fewest possible degrees of freedom, so its predictions are extremely sensitive to the noise: expressed information is near its maximum. As $p$ grows further, the noise is spread over many directions and expressed less. In Section~\ref{sec:exp:dd} we show that test error and $V$ peak together, on synthetic and real data, and that a small ridge penalty removes both peaks. This is consistent with variance-based accounts of double descent \citep{adlam2020understanding,yang2020rethinking,mei2022random}; the information view adds that storage is constant along the whole interpolating branch, so the entire curve is driven by expression.

%% file: sections/s6_selection.tex
\section{Expressed information as a model-selection criterion}\label{sec:selection}

In the last two sections we used expressed information to \emph{explain} benign overfitting. We will now use it to \emph{choose} among learners. We first explain why choosing among interpolators is hard for standard tools (Section~\ref{sec:sel:why}). We then show that $V$ ranks candidates correctly when they have no modal bias, and argue that the overparameterized regime is where this is most nearly the case (Section~\ref{sec:sel:rank}). Section~\ref{sec:sel:est} explains how to estimate $V$ from a single dataset, and why this must be done with disjoint splits. Finally, Section~\ref{sec:sel:beyond} describes when $V$ alone is not enough, and what to do then.

\subsection{Why selecting among interpolators is hard}\label{sec:sel:why}

Suppose we have several candidate learners, for example the five interpolators of Table~\ref{tab:pilot}, and one noisy training set. Let us recall what the previous sections tell us about the usual tools.
\begin{itemize}[leftmargin=*]
\item \emph{Training error} is zero for every candidate, so it cannot rank them.
\item \emph{Information stored} in the weights is $n\,H(Y\mid X)$ for every candidate (Theorem~\ref{thm:storage}), so neither storage nor the mutual-information bounds built on it can rank them (Corollary~\ref{cor:xr}).
\item \emph{Information measured on the training sample}, such as the legacy LMC score, is degenerate for interpolators (Section~\ref{sec:lmc}).
\item \emph{Validation} works, but it needs labels that are not used for training, and in a noisy problem these labels are noisy too. With $n_{\rm val}$ validation points, the standard error of a validation error is about $\sqrt{R/n_{\rm val}}$, which, for small validation sets, is often larger than the differences between good candidates.
\end{itemize}
Expressed information avoids all four problems. It looks at predictions at fresh inputs, not at the training set; it does not need validation labels, only unlabelled inputs; and, by Theorem~\ref{thm:decomp}, it is directly tied to excess risk. Being free of ground truth at the selection step, it can also be used where held-out labels are not available at all, for instance in semi-supervised or streaming settings; Section~\ref{sec:discussion} discusses these uses and the unsupervised analogue.

\subsection{Expressed information ranks candidates without modal bias}\label{sec:sel:rank}

Recall from Theorem~\ref{thm:decomp} that excess risk has two sources: expression, measured by $V$, and modal bias, the mass of inputs where the learner's most frequent prediction is wrong. If the candidates have no modal bias, only the first source is left, and $V$ should rank them. The next proposition makes this precise.

\begin{proposition}[Ranking candidates without modal bias]\label{prop:rank}
Consider candidate learners $j=1,\dots,J$ whose modal predictions are all correct, $m_j(X)=b^*(X)$ almost surely.
\begin{enumerate}[label=(\alph*),leftmargin=*]
\item For binary labels, with $\pi_j(x)=h^{-1}(V_j(x))$ the probability that candidate $j$ departs from its mode at $x$,
\[
\E\,R(h_{W_j})-R^*\;=\;\E\big[\,|2\eta_1(X)-1|\;\pi_j(X)\,\big].
\]
Hence, if $V_i(x)\le V_j(x)$ for almost every $x$, then candidate $i$ has excess risk no larger than candidate $j$. If the margin $|2\eta_1-1|$ is constant, as it is for symmetric label noise on a noiseless task, then ranking by the mean off-mode probability $\bar\pi_j=\E\,\pi_j(X)$ is exactly ranking by excess risk.
\item For $K$ classes, $\E[\gamma(X)\,\phi_K^{-1}(V_j(X))]\le\E\,R(h_{W_j})-R^*\le\frac12\E[\Delta(X)\,V_j(X)]$, so the excess risk of every candidate is sandwiched between two increasing functions of its pointwise expressed information.
\end{enumerate}
\end{proposition}
\begin{proof}
Both parts follow from Theorem~\ref{thm:decomp} with $\ind\{m_j\ne b^*\}=0$. Monotonicity in (a) holds because $h^{-1}$ is increasing.
\end{proof}

In plain terms, among learners whose ``average prediction'' is right, the one whose predictions depend least on the particular sample is the best. Note that the proposition needs no labels at all: $V_j$ and $\pi_j$ are properties of the candidates' predictions only.

\paragraph{Why the overparameterized regime.} The condition of Proposition~\ref{prop:rank}, no modal bias, is not something we can check directly. But there are good reasons to expect it to hold approximately among flexible, overparameterized learners, and to fail mainly for rigid ones.
\begin{itemize}[leftmargin=*]
\item The modal prediction of a learner is the prediction of an infinite bagged ensemble of independent copies of it (Section~\ref{sec:expression}). For flexible learners this ensemble is typically close to consistent: for $k$-NN, the mode converges to $b^*$ for every fixed $k$ (Proposition~\ref{prop:knn}), and similar statements hold for trees and forests. What separates flexible interpolators is then how much they express, not where their average prediction lies.
\item In the running example (Example~\ref{ex:running}) at $n=4096$, the modal bias of every interpolator is at most $0.0004$, while their excess risks range from $0.001$ to $0.13$. Over all 72 noisy interpolating configurations of our synthetic experiment (Section~\ref{sec:exp:synth}), modal bias accounts on average for about $4\%$ of excess risk.
\item Learners with substantial modal bias are those that cannot represent the Bayes classifier at all, such as a linear model on a curved boundary. These are typically the \emph{underparameterized} candidates, which do not interpolate. It is tempting to screen them out by their training fit before applying min-$V$, but under label noise this also removes the well-regularized candidates, which do not fit noisy labels either and are often the best (Section~\ref{sec:exp:beyond}). They are better handled by the extension of Section~\ref{sec:sel:beyond}, which estimates modal bias directly.
\end{itemize}
So, among overparameterized candidates, Proposition~\ref{prop:rank} suggests a simple rule: \emph{choose the candidate with the smallest expressed information.} We call this rule \emph{min-$V$}. We test it in Section~\ref{sec:exp:select}.

\subsection{Estimating expressed information}\label{sec:sel:est}

Expressed information is defined over independent training samples. Let us now discuss how to estimate it in two situations: when several independent samples are available, and when, as is usual, there is only one.

\paragraph{Plug-in estimate.} Let $h_1,\dots,h_R$ be $R$ trained copies of a candidate, and let $x_1,\dots,x_M$ be unlabelled inputs (for example the inputs of a test set, or any unlabelled data from the same distribution). For each $x_m$, let $\hat q(\cdot\mid x_m)$ be the empirical distribution of the $R$ predictions. We estimate $V$ by
\[
\widehat V=\frac1M\sum_{m=1}^M\Big[H\big(\hat q(\cdot\mid x_m)\big)+\frac{\hat k_m-1}{2R\ln2}\Big],
\]
where $\hat k_m$ is the number of distinct predictions at $x_m$; the second term is the Miller--Madow bias correction \citep{miller1955note}. Similarly, $\hat{\bar\pi}=\frac1M\sum_m(1-\max_k\hat q(k\mid x_m))$ estimates the mean off-mode probability. The plug-in entropy is biased low for small $R$, but the bias affects all candidates in a similar way. In our experiments (Appendix~\ref{sec:estimator}) the corrected estimate from $R=5$ runs recovers $62$--$91\%$ of the reference value and ranks four learners correctly for every $R\ge5$.

\paragraph{Several independent samples.} If the $R$ copies are trained on independent samples, with independent label noise, $\widehat V$ is a consistent estimate of $\E V$ as $R,M\to\infty$. This situation arises when several labelled batches, or several independent noisy labellings, are available.

\paragraph{One dataset: use disjoint splits.} Usually there is a single labelled dataset. The natural idea is to reuse the fold models of $R$-fold cross-validation. There is a catch, and it matters. Fold models of $R$-fold cross-validation with $R>2$ share most of their training points and, crucially, the \emph{same noisy labels}. A mislabelled training point is seen by $R-1$ of the $R$ fold models, and they will tend to agree near it. So the expression of label noise, which is precisely what $V$ is meant to measure, is largely hidden.

\begin{remark}[Overlapping folds hide expression]\label{rem:overlap}
Consider the 1-NN rule and a test input $x$ whose nearest neighbour in the full training set is the point $(X_i,Y_i)$. That point is in the training set of $R-1$ of the $R$ fold models of $R$-fold cross-validation, and each of them predicts $Y_i$ at $x$. So the estimated probability of departing from the mode at $x$ is at most $1/R$, whatever the noise level. In contrast, Proposition~\ref{prop:knn} says that the true off-mode probability of 1-NN tends to $\min(\eta_1(x),1-\eta_1(x))$.
\end{remark}

The remedy is to train the copies on \emph{disjoint} parts of the data. With two disjoint halves, the two copies see independent labels, and the only cost is that each sees half the data. Their predictions also give a 2-fold cross-validation estimate for free. We write $V_2$ for $V$ estimated from the two fold models of 2-fold cross-validation, and $V_5$ for $V$ estimated from the five fold models of 5-fold cross-validation. Remark~\ref{rem:overlap} predicts that $V_5$ is less reliable than $V_2$, despite using more models; Section~\ref{sec:exp:select} confirms this.

\paragraph{Cost.} Computing $V_2$ costs two fits on half of the data, plus predictions on unlabelled inputs. In our measurements (Section~\ref{sec:exp:select}) this is $1.25$ times the cost of one holdout fit, of which the unlabelled predictions are about $1\%$.

\subsection{When expressed information alone is not enough}\label{sec:sel:beyond}

Proposition~\ref{prop:rank} assumes that the candidates have no modal bias. We now discuss what happens when this fails, and give two remedies. The theory behind this subsection is in Appendix~\ref{sec:consensus}.

\paragraph{Failure 1: underfitting candidates.} If the pool contains rigid learners, such as strongly regularized linear models, these have low $V$ (their predictions hardly depend on the sample) and may have large modal bias. Min-$V$ will then prefer them wrongly. Some estimate of modal bias is needed.

\paragraph{Remedy 1: a label-free consensus criterion.} The modal bias of candidate $j$ is $\beta_j=\P(m_j(X)\ne b^*(X))$. We cannot compute it, since $b^*$ is unknown, but we can compare each candidate's modal prediction with the \emph{consensus} $c(x)$, the plurality vote of all candidates' modal predictions. Let $D_j=\P(m_j(X)\ne c(X))$ be candidate $j$'s disagreement with the consensus. We score candidate $j$ by
\begin{equation}\label{eq:lf}
\mathrm{LF}_j\;=\;D_j\;+\;\bar\pi_j\Big(1-\tfrac{K}{K-1}\,D_j\Big),
\end{equation}
and choose the candidate with the lowest score. The form of \eqref{eq:lf} mirrors the exact decomposition \eqref{eq:exact}: if the consensus were the Bayes classifier, $D_j$ would be the modal-bias mass, and a candidate would err where its mode is wrong, plus where it departs from a correct mode. The criterion uses no labels at all, only predictions on unlabelled inputs. Appendix~\ref{sec:consensus} shows that disagreement estimates modal bias up to the consensus error $\delta=\P(c(X)\ne b^*(X))$, in the sense that $|D_j-\beta_j|\le\delta$, and that $\delta\le2\bar\beta$, twice the average modal bias of the candidates. So the consensus can be trusted unless a large part of the pool shares the same bias. When many candidates are near-copies of one model family, giving one vote per cluster of near-identical candidates restores this (Appendix~\ref{sec:consensus}).

\paragraph{Failure 2: asymmetric label noise.} If the training labels are flipped more often in one direction than the other, every candidate learns the same shifted decision boundary. Their modal predictions all agree, and all are wrong on the same region. This shared bias is invisible to $V$, to the consensus, and also to validation on equally noisy labels. For binary labels flipped $0\to1$ with probability $\rho_0$ and $1\to0$ with probability $\rho_1$, the shift moves the threshold on $\eta_1$ from $\frac12$ to $(\frac12-\rho_0)/(1-\rho_0-\rho_1)$; it vanishes exactly when $\rho_0=\rho_1$ (Proposition~\ref{prop:noise-shift}).

\paragraph{Remedy 2: detect asymmetry and switch to noise-corrected validation.} Label noise can be corrected if its transition matrix $T_{jk}=\P(\tilde Y=k\mid Y=j)$ is known, by the backward correction of \citet{natarajan2013noisy} and \citet{patrini2017making}: the loss of predicting $\hat y$ for a noisy label $\tilde y$ is replaced by $1-(T^{-1})_{\tilde y\hat y}$, which is unbiased for the clean 0-1 loss. We estimate $T$ without extra labels, by cross-tabulating the consensus of the candidates' out-of-fold predictions against the noisy training labels. Under symmetric noise, correction cannot change the ranking of candidates at all, because the corrected loss is an increasing affine function of the noisy one (Remark~\ref{rem:symcorr}); it can only help when the noise is asymmetric. This suggests a simple \emph{switch rule}: measure the asymmetry of $\hat T$ (its largest deviation from a symmetric noise matrix with the same mean flip rate); if it exceeds $0.1$, select by noise-corrected cross-validation; otherwise, use the label-free criterion.

\paragraph{Summary of the recommendations.} Putting this section together, we suggest the following.
\begin{enumerate}[leftmargin=*]
\item Among overparameterized, flexible candidates, select by min-$V$, estimated on unlabelled inputs from copies trained on \emph{disjoint} parts of the data ($V_2$ from a single dataset).
\item When the pool also contains rigid or regularized candidates, use the label-free criterion \eqref{eq:lf}, which adds a consensus estimate of modal bias.
\item When the label noise may be asymmetric, check the asymmetry of $\hat T$, and switch to noise-corrected cross-validation if it is large.
\end{enumerate}
Section~\ref{sec:exp:select} tests the first rule and Section~\ref{sec:exp:beyond} the other two.

%% file: sections/s7_experiments.tex
\section{Experiments}\label{sec:experiments}

We now test the theory of Sections~\ref{sec:storage-expression}--\ref{sec:benign} and the selection rules of Section~\ref{sec:selection}. Let us first say what each experiment is for.
\begin{itemize}[leftmargin=*]
\item Section~\ref{sec:exp:synth} checks the decomposition of Theorem~\ref{thm:decomp} on synthetic data, where excess risk and $V$ can be computed exactly, and illustrates the benign/tempered distinction of Section~\ref{sec:taxonomy}.
\item Section~\ref{sec:exp:dd} shows that double descent is a peak of expressed information (Section~\ref{sec:ridgeless}), and that min-$V$ picks the right width among interpolating ones.
\item Section~\ref{sec:exp:boost} revisits the information-plane picture of boosting from the LMC papers, which motivated this paper.
\item Section~\ref{sec:exp:real} checks that $V$ ranks learners on real data with label noise.
\item Section~\ref{sec:exp:select} is the main model-selection experiment: choosing among overparameterized learners, under two protocols.
\item Section~\ref{sec:exp:cmi} compares min-$V$ with selection by conditional mutual information at matched cost.
\item Section~\ref{sec:exp:beyond} tests the extensions of Section~\ref{sec:sel:beyond} when the pool also contains underparameterized learners.
\end{itemize}
Details of all experiments (datasets, learner configurations, noise models, protocols and compute) are in Appendix~\ref{sec:expdetails}, and further results in Appendix~\ref{sec:addresults}. Appendix~\ref{sec:code} maps every figure and table to the script that produces it. In all tables, \best{bold} marks the best value in each column of each comparison, and \prop{} marks methods proposed in this paper.

\subsection{Exact validation on synthetic data}\label{sec:exp:synth}

\paragraph{Setup.} We use the running example (Example~\ref{ex:running}), for which $\eta$ and $b^*$ are known, so excess risk and $V$ can be computed exactly over all $2^{19}$ inputs (in practice, on a fixed Monte Carlo set of 20{,}000 test inputs). We train 8 learners: memorize-and-smooth (Proposition~\ref{prop:memsmooth}), 1-NN, $k$-NN with $k\approx\sqrt n$, a full-depth tree, a random forest, gradient boosting with exponential loss, a 512-unit MLP trained to interpolate, and weakly regularized logistic regression. We use $n\in\{256,1024,4096\}$ and label noise $\varepsilon\in\{0,0.1,0.2,0.3,0.4\}$, which gives 120 configurations, each with $R=30$ independent training samples.

\begin{figure}[t]\centering
\includegraphics[width=0.42\linewidth]{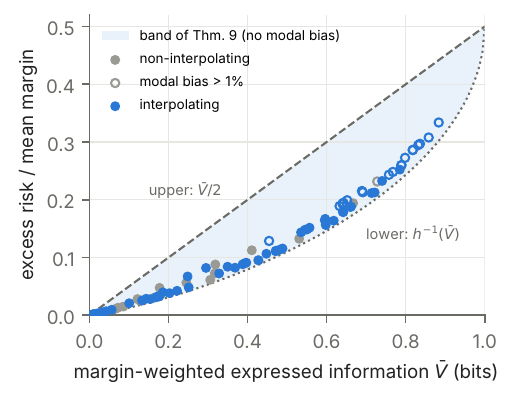}\hfill
\includegraphics[width=0.42\linewidth]{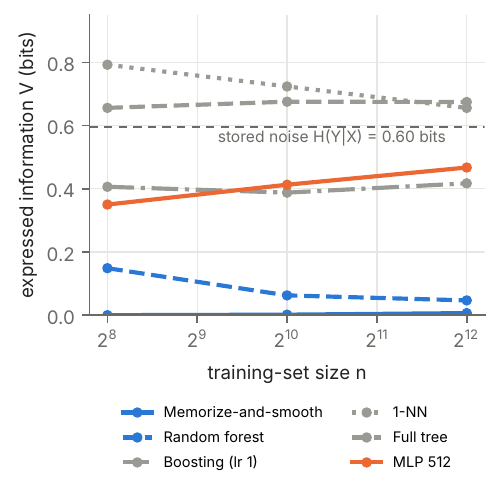}
\caption{Synthetic validation on the running example. \emph{Left:} the two-sided bound of Theorem~\ref{thm:decomp} for 120 configurations (blue: interpolating). Both axes are normalised by the mean margin $\bar m=\E|2\eta_1(X)-1|$, which differs between noise levels: the vertical axis is exact excess risk divided by $\bar m$, the horizontal axis is the margin-weighted expressed information $\bar V=\E[|2\eta_1(X)-1|\,V(X)]/\bar m$. For binary labels the theorem then gives $h^{-1}(\bar V)\le(\E R-R^*)/\bar m\le\bar V/2+\beta/\bar m$, with $\beta$ the margin-weighted modal bias; the shaded band is this range for $\beta=0$. The lower edge holds for every learner, the upper edge when $\beta=0$; hollow points have $\beta/\bar m>1\%$. \emph{Right:} $V$ against training-set size at $\varepsilon=0.2$. Memorize-and-smooth and the random forest are benign ($V\to0$); 1-NN and the full tree are tempered, with $V$ approaching the label noise $H(Y\mid X)=0.595$ bits (dashed); the MLP's $V$ grows with $n$.}\label{fig:synth}
\end{figure}

\paragraph{Results.} Figure~\ref{fig:synth} (left) shows that $V$ and excess risk are tightly linked. The Spearman correlation between them is $0.96$ over all 120 configurations, and $0.92$ over the 72 noisy interpolating ones. For comparison, training error has a correlation of $0.14$ with excess risk. Every configuration lies within the bounds of Theorem~\ref{thm:decomp}. To show this with a single band for all noise levels, the figure normalises by the mean margin and weights $V$ by the margin, as in the theorem; the band from $h^{-1}(\bar V)$ to $\bar V/2$ is then the tightest range that can be stated in terms of the mean of $V$ alone (the lower edge follows from the exact binary identity by convexity of $h^{-1}$, and the upper edge is attained when $V(x)$ takes only the values $0$ and $1$). The points lie close to the lower edge (for configurations with $\bar V>0.05$ and no modal bias, a median of $11\%$ of the way from the lower to the upper edge), which means that the off-mode probability is spread fairly evenly over inputs, so that $h^{-1}(\bar V)$ is itself a good estimate of normalised excess risk. The modal bias of every interpolator is small at moderate noise (at most $0.0004$ at $n=4096$ and $\varepsilon=0.2$), as Section~\ref{sec:sel:rank} anticipated, so almost all of the excess risk of interpolators comes from expression. It grows at high noise (up to $0.019$ at $\varepsilon=0.4$), partly because the mode is then estimated from only 30 copies.

Figure~\ref{fig:synth} (right) illustrates the taxonomy of Section~\ref{sec:taxonomy}. Memorize-and-smooth and the random forest are benign: their $V$ decreases towards $0$ as $n$ grows. The 1-NN rule and the full tree are tempered: their $V$ stays close to $H(Y\mid X)=0.595$ bits, as Proposition~\ref{prop:knn} predicts for 1-NN. The interpolating MLP is tempered too, but becomes worse with more data: from $n=256$ to $n=4096$ its excess risk grows from $0.051$ to $0.075$ and its $V$ from $0.35$ to $0.47$ bits. All of these learners store the same $n\,H(Y\mid X)$ bits (Theorem~\ref{thm:storage}).

\paragraph{Excess complexity of the trained outputs.} The ideal-model quantities of Section~\ref{sec:sufficiency} can also be measured. The 0-1 excess complexity of a learner's output, $H(F\mid b^*(X))$, has a Spearman correlation of $0.95$ with excess risk. The held-out log-loss estimator of Proposition~\ref{prop:excess} has correlation $0.85$, but it underestimates the true value by $0.35$ bits on average (Appendix~\ref{sec:heldout} explains why: it is a lower bound for non-sufficient outputs).

\paragraph{Disagreement measures excess risk, not risk.} Table~\ref{tab:dis} checks Proposition~\ref{prop:disagreement-main} on the running example at $n=4096$, with 10 copies per learner. For every learner, disagreement lies between excess risk and $V$, as part (b) predicts. For the tempered learners (1-NN, the full tree, boosting), disagreement over independent samples is close to the test error, so the generalization disagreement equality of \citet{jiang2022disagreement} approximately holds. For the benign or consistent ones (memorize-and-smooth, the random forest, and the non-interpolating $k$-NN with $k\approx\sqrt n$), disagreement is close to zero while the test error stays at $R^*=0.15$, as part (c) predicts. Disagreement between copies trained on the same sample with different seeds, the setting of the GDE, is zero for the deterministic learners, including 1-NN, whose test error is $0.27$.

\begin{table}[t]\centering\small
\caption{Disagreement, excess risk and expressed information on the running example ($n=4096$, $\varepsilon=0.2$, $R^*=0.150$; 10 copies per learner; exact risks). ``Dis.\ (samples)'': disagreement of copies trained on independent samples; ``Dis.\ (seeds)'': copies trained on one sample with different seeds, as in the GDE. \prop: learner constructed in this paper. Descriptive table; nothing is bolded.}\label{tab:dis}
\resizebox{\linewidth}{!}{%
\begin{tabular}{lcccccc}\toprule
Learner & Test error & Excess risk & Dis.\ (samples) & Dis.\ (seeds) & Test error $-$ Dis. & $V$ (bits)\\\midrule
Memorize-and-smooth\prop & 0.150 & 0.001 & 0.002 & 0.000 & 0.148 & 0.005\\
$k$-NN, $k\approx\sqrt n$ & 0.150 & 0.001 & 0.002 & 0.000 & 0.148 & 0.006\\
Random forest & 0.154 & 0.005 & 0.014 & 0.006 & 0.140 & 0.033\\
Gradient boosting & 0.213 & 0.063 & 0.170 & 0.080 & 0.043 & 0.360\\
1-nearest neighbour & 0.273 & 0.124 & 0.295 & 0.000 & $-0.021$ & 0.593\\
Decision tree (full depth) & 0.280 & 0.130 & 0.306 & 0.103 & $-0.026$ & 0.613\\
\bottomrule
\end{tabular}}
\end{table}

\subsection{Double descent and width selection}\label{sec:exp:dd}

\paragraph{Setup.} We fit random ReLU features by minimum-norm least squares on $\pm1$ labels and predict by sign, sweeping the number of features $p$ from $16$ to $4096$ at $n=512$, with 40 independent training sets per width. We repeat this on two real datasets (\texttt{digits}, 10 classes, $n=629$, with one-hot targets; and \texttt{wdbc}, $n=199$), with 20\% injected label noise.

\begin{figure}[t]\centering
\includegraphics[width=0.9\linewidth]{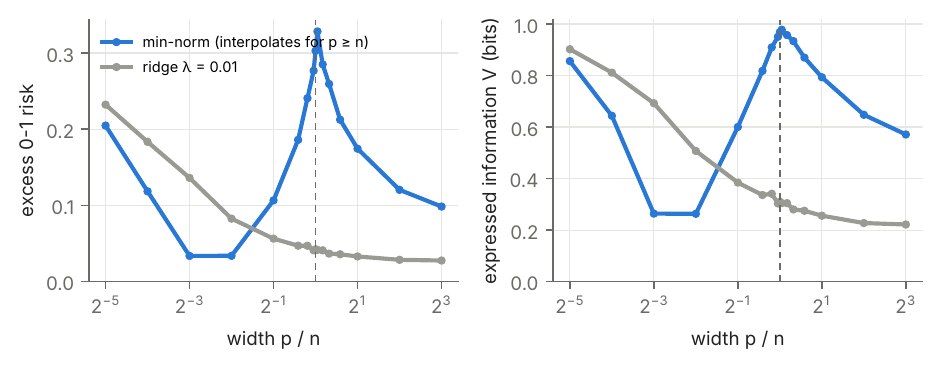}
\caption{Random-features least squares at $n=512$ with label noise $0.2$. \emph{Left:} exact excess risk; \emph{right:} expressed information; both against $p/n$ (log scale). Minimum-norm interpolation (blue) shows a double-descent peak at $p\approx n$ in both quantities; a small ridge penalty (grey) removes both.}\label{fig:dd}
\end{figure}

\paragraph{Double descent is a peak of expression.} Figure~\ref{fig:dd} shows the familiar double-descent curve. At $\varepsilon=0.2$, excess risk is $0.034$ at $p=64$, peaks at $0.33$ at $p=528\approx n$, and falls to $0.099$ at $p=4096$. Expressed information follows the same curve: from $0.26$ bits, to a peak of $0.98$ bits (close to its maximum of 1 bit), and down to $0.57$ bits. A small ridge penalty ($\lambda=10^{-2}$) removes both peaks. Across all 64 configurations, the Spearman correlation of $V$ with excess risk is $0.99$, and the bound of Theorem~\ref{thm:decomp} holds in every case. The real datasets show the same picture (Appendix~\ref{sec:add-dd}): on \texttt{digits} the test error peaks at $0.88$ together with $V=3.07$ bits, against a ceiling of $\log_210=3.32$ bits.

\paragraph{Choosing the width among interpolating ones.} Restricting to interpolating widths ($p\ge n$, minimum-norm), we ask whether min-$V$ chooses the best width. Over the six cases (synthetic with $\varepsilon\in\{0,0.2\}$, and the two real datasets with and without injected noise), the Spearman correlation between $V$ and test error ranges from $0.98$ to $1.00$. Min-$V$ chooses the best width in five of the six cases, and in the sixth (noiseless \texttt{wdbc}) its regret is $0.0029$. A random interpolating width would have regret between $0.0018$ and $0.23$.

\subsection{The information plane of boosting, revisited}\label{sec:exp:boost}

The LMC papers argued, from information-plane trajectories, that boosting compresses its output as it maximizes margins, and that this compression explains why it keeps improving after reaching zero training error \citep{nikolaou2020margin,nikolaou2021lossless}. Let us now revisit that picture with the tools of this paper.

\begin{figure}[t]\centering
\includegraphics[width=0.5\linewidth]{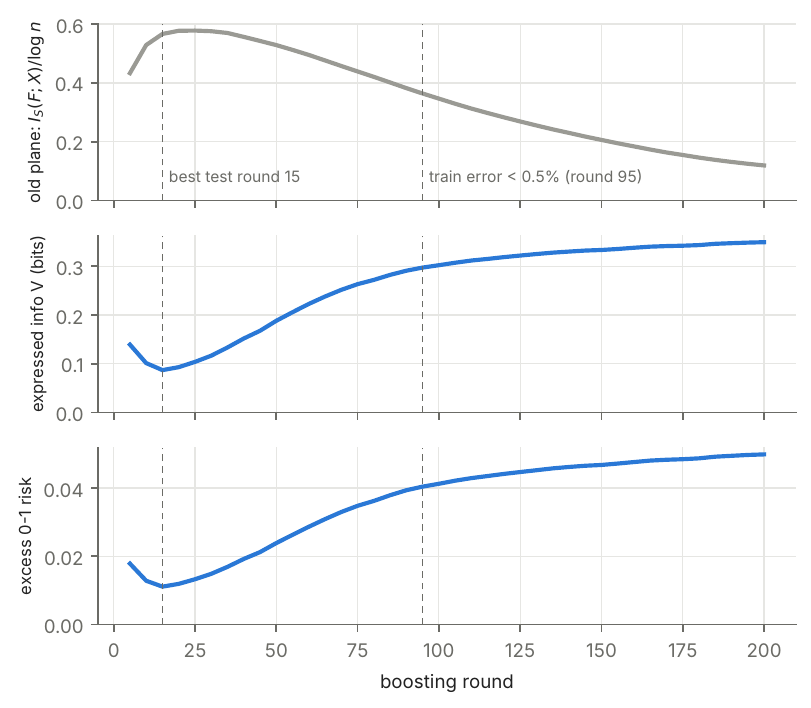}
\caption{Gradient boosting on the running example ($\varepsilon=0.2$, shrinkage $0.1$). \emph{Top:} the information-plane coordinate used in the LMC papers, which rises and then falls (``compression''). \emph{Middle:} expressed information. \emph{Bottom:} exact excess risk. The dotted lines mark the best round (15) and the round where training error reaches $0.5\%$.}\label{fig:boost}
\end{figure}

On the running example with $\varepsilon=0.2$ (Figure~\ref{fig:boost}), the old information-plane coordinate rises and then falls, from $0.58$ to $0.12$, exactly the ``compression'' that the earlier papers reported. But over the same rounds, the excess risk \emph{increases} from its minimum of $0.011$ at round 15 to $0.050$ at round 200, a factor of $4.5$. Expressed information rises with it, from $0.09$ to $0.35$ bits. So, under label noise, the apparent compression goes together with \emph{worse} generalization, and it is expression, not compression, that tracks test performance. The same happens on four real datasets with 20\% injected noise (Appendix~\ref{sec:add-boost}): test error rises from its minimum to its final value, for example from $0.101$ to $0.144$ on \texttt{wdbc} and from $0.093$ to $0.165$ on \texttt{biopsy}, while the old coordinate shows ``compression''. Without injected noise, on the nearly noiseless \texttt{wdbc} and \texttt{biopsy}, compression and falling test error coincide, as reported earlier. The earlier observations are thus reproduced where the labels are nearly noiseless, and reversed where they are not. Appendix~\ref{sec:erratum} discusses this in detail.

\subsection{Expressed information ranks learners on real data}\label{sec:exp:real}

\paragraph{Setup.} We use five datasets (\texttt{wdbc}, \texttt{biopsy}, \texttt{OJ}, a class-balanced \texttt{Caravan}, and \texttt{digits}). A fixed 30\% of each is held out as a clean test set, and we inject symmetric label noise at rate $\rho\in\{0,0.1,0.2,0.3\}$ into the training labels only. We train 19 learner configurations ($k$-NN, trees, random forests, gradient boosting, logistic regression, RBF SVMs and MLPs, each at several capacity levels; Appendix~\ref{sec:expdetails}), each $R=12$ times on random halves of the training pool with fresh noise draws. The Bayes risk is unknown, so we report clean test error, which ranks learners exactly as excess risk does within a dataset.

\begin{figure}[t]\centering
\includegraphics[width=\linewidth]{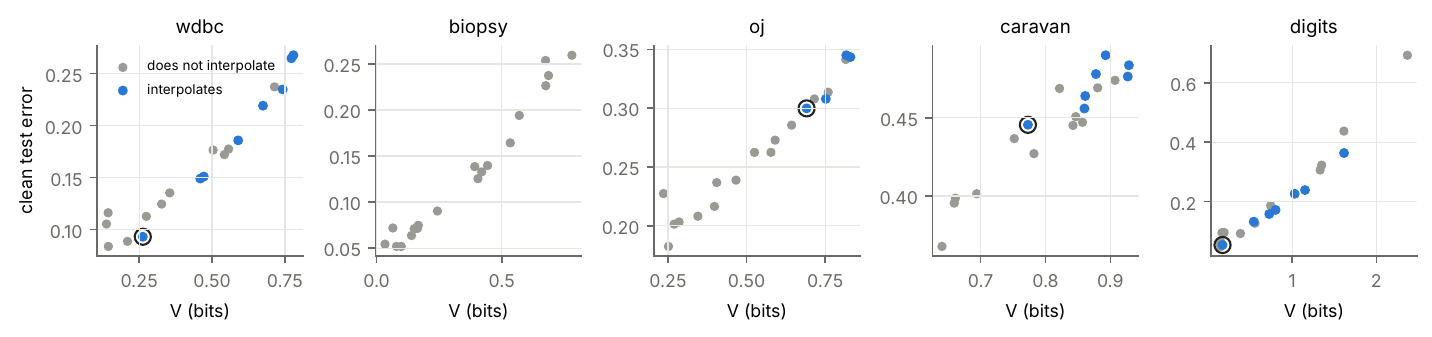}
\caption{Real data with 20\% injected label noise: clean test error against expressed information for 19 learner configurations per dataset (blue: interpolating). The circled point is the interpolator with the smallest $V$, chosen without validation labels.}\label{fig:real}
\end{figure}

\paragraph{Results.} Across the 20 dataset--noise combinations, the median Spearman correlation between $V$ and clean test error is $0.94$, and $0.96$ when restricted to interpolators (Figure~\ref{fig:real}). The legacy LMC score is between $0.91$ and $1.00$ for every interpolator under injected noise, while their test errors range from $0.04$ to $0.50$: it does not rank them. Choosing the interpolator with the smallest $V$ selects the one with the lowest test error in 16 of 17 cases (the remaining case costs $0.0015$); a random interpolator would cost $0.059$ on average. Finally, as Proposition~\ref{prop:ensemble} predicts, the majority vote of the $R$ copies beats a single copy in every combination, and the gain grows with the noise rate (for example from $0.027$ to $0.150$ on \texttt{digits} as $\rho$ goes from $0$ to $0.3$).

\subsection{Selecting among overparameterized learners}\label{sec:exp:select}

This is the main experiment on model selection. We ask whether min-$V$ (Section~\ref{sec:sel:rank}) is a viable alternative to holdout validation and cross-validation for choosing among overparameterized learners. We use two protocols, which differ in what is available to the practitioner.

\paragraph{Protocol A: independent training samples.} This is the setting in which $V$ is defined. On the five datasets of Section~\ref{sec:exp:real}, each candidate is trained 12 times on independent random halves of a training pool, each with its own label-noise draw. Holdout uses a separate validation set with labels corrupted by the same noise process (113--359 points), or a random subset of 40 of its points. We consider symmetric noise ($0.2$ and $0.3$; 60 cases) and class-dependent noise (30 cases). Regret is the mean clean test error of a single trained model of the chosen configuration, minus that of the best configuration. Here we restrict the choice to configurations that interpolate.

\paragraph{Protocol B: one noisy dataset, matched compute.} This is the realistic setting. On 15 datasets (the five above and ten PMLB benchmarks, each of at most 2{,}000 points), 30\% is held out as a clean test set and the rest receives \emph{one} draw of label noise: symmetric ($0.2$), class-dependent, or instance-dependent (labels near the decision boundary are flipped more often; mean flip rate $0.16$). Each dataset--noise combination is repeated with five seeds, which change the train/test split, the subsample (for datasets above 2{,}000 points), the noise draw and the folds. There are 225 cases. The candidates are the six configurations designed to interpolate: 1-NN, a full-depth tree, a random forest with leaf size 1, gradient boosting with 300 depth-6 trees, an RBF SVM with $C=10^4$, and a weakly regularized 512-unit MLP. Each is fitted once on 80\% of the training set (for holdout), in 2-fold and 5-fold cross-validation, and on the full training set. Regret is the clean test error of the full refit of the chosen candidate, minus that of the best. We measure CPU time for every fit and every prediction. Means are over all 225 cases; 95\% intervals for paired differences use a cluster bootstrap that resamples the 45 dataset--noise combinations, so the five seeds of one dataset are not treated as independent.

\begin{table}[t]\centering\small
\caption{Selecting among overparameterized learners: regret in clean test error (lower is better; mean over cases). Best per column in bold, within each panel. \prop: proposed in this paper. \emph{Panel A} (independent training samples, interpolating configurations of 19): 12 independent fits per candidate for every criterion. \emph{Panel B} (one noisy dataset; six overparameterized candidates; 15 datasets $\times$ 5 seeds per noise model): cost is the median CPU time relative to one holdout fit, including predictions on unlabelled inputs; ``exact'' counts the cases (of 225) in which the best candidate was chosen. $^\ast$The consensus criterion and the noise-corrected CV use the consensus of a reference pool of all 19 configurations; their costs include fitting that pool. A random choice has regret $0.085$ in Panel B.}\label{tab:select}
\resizebox{\linewidth}{!}{%
\begin{tabular}{lcccccc}\toprule
\multicolumn{7}{l}{\emph{Panel A: independent training samples}}\\
& \multicolumn{2}{c}{Symmetric (60 cases)} & \multicolumn{2}{c}{Class-dependent (30 cases)} & & \\
Criterion & full validation & 40 points & full validation & 40 points & & \\\midrule
Holdout (noisy labels) & 0.0031 & 0.0094 & 0.0051 & 0.0136 & & \\
Min-$V$\prop{} (no validation labels) & \best{0.0010} & \best{0.0010} & \best{0.0037} & \best{0.0037} & & \\
Consensus criterion \eqref{eq:lf}\prop{} (no validation labels) & 0.0014 & 0.0014 & \best{0.0037} & \best{0.0037} & & \\\midrule
\multicolumn{7}{l}{\emph{Panel B: one noisy dataset, measured compute}}\\
Criterion & Symmetric & Class-dep. & Instance-dep. & All (225) & Cost & Exact\\\midrule
Holdout (80/20) & 0.0083 & 0.0218 & 0.0107 & 0.0136 & \best{1.00} & 156\\
2-fold CV & 0.0068 & 0.0173 & 0.0053 & 0.0098 & 1.24 & 164\\
5-fold CV & 0.0036 & 0.0100 & 0.0033 & 0.0056 & 4.99 & 177\\
5-fold CV, noise-corrected$^\ast$ & 0.0036 & 0.0105 & 0.0030 & 0.0057 & 9.30 & 178\\
Min-$V_2$\prop{} (2 disjoint halves) & 0.0022 & 0.0121 & 0.0021 & 0.0055 & 1.25 & 181\\
Min-$V_5$\prop{} (5 overlapping folds) & 0.0025 & 0.0229 & 0.0048 & 0.0101 & 5.05 & 173\\
Consensus criterion$^\ast$\prop{}, 2 disjoint halves & 0.0023 & \best{0.0057} & 0.0019 & \best{0.0033} & 2.39 & \best{187}\\
Consensus criterion$^\ast$\prop{}, 5 folds & \best{0.0015} & 0.0071 & \best{0.0015} & 0.0034 & 9.40 & 186\\\bottomrule
\end{tabular}}
\end{table}

\begin{figure}[t]\centering
\includegraphics[width=\linewidth]{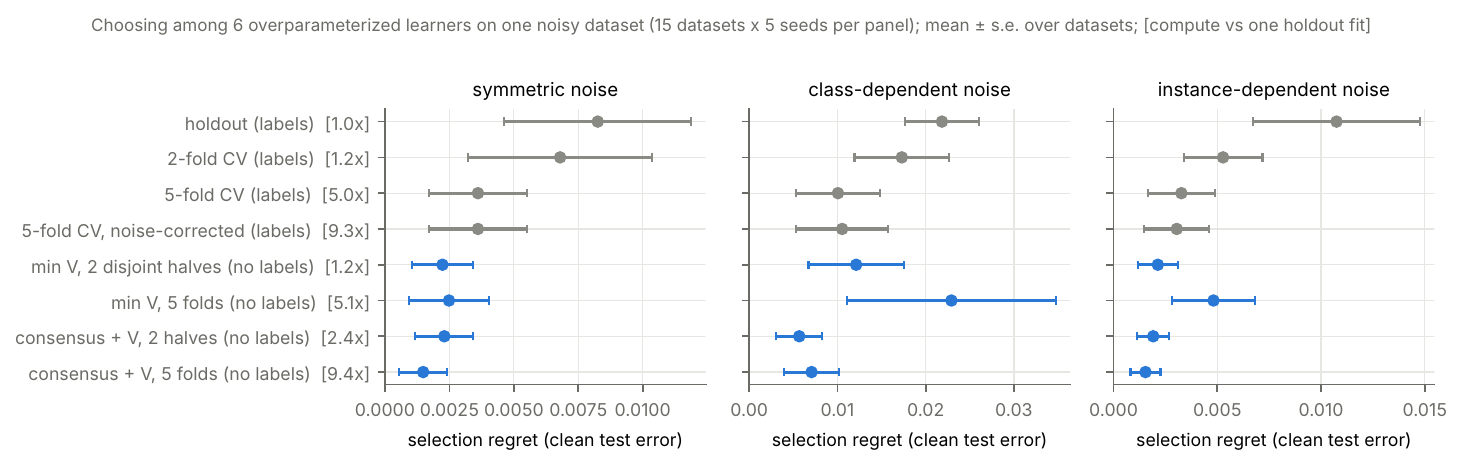}
\caption{Protocol B in detail: selection regret for each noise model (mean over 15 datasets $\times$ 5 seeds; error bars: standard error over datasets, after averaging the seeds of each dataset). Grey: criteria that use held-out labels; blue: criteria based on expressed information, which use no held-out labels. The consensus criteria and noise-corrected CV use a reference pool of all 19 configurations, whose cost is included. Bracketed: median compute relative to one holdout fit.}\label{fig:overparam}
\end{figure}

\paragraph{Results with independent samples (Table~\ref{tab:select}, Panel A).} When independent training samples are available, min-$V$ is the best or tied-best criterion in every setting. Under symmetric noise it has a third of the regret of holdout with the full validation set ($0.0010$ against $0.0031$), and a ninth of it with 40 validation points ($0.0010$ against $0.0094$). The 95\% bootstrap intervals for the difference from holdout are $-0.002$ [$-0.005$, $0.001$] and $-0.008$ [$-0.012$, $-0.005$] under symmetric noise (full validation set and 40 points), and $-0.001$ [$-0.004$, $0.000$] and $-0.010$ [$-0.015$, $-0.006$] under class-dependent noise. So the advantage is clear with small validation sets, and within noise with large ones. Min-$V$ does not depend on the validation set at all, so it is unaffected by the size of the validation set, while holdout degrades as the validation set shrinks.

\paragraph{Results on one dataset (Table~\ref{tab:select}, Panel B; Figure~\ref{fig:overparam}).} We highlight four findings; brackets give cluster-bootstrap 95\% intervals for paired differences.
\begin{enumerate}[leftmargin=*]
\item \emph{Min-$V_2$ beats holdout and 2-fold CV at the same cost, and matches 5-fold CV at a quarter of the cost.} At $1.25\times$ the cost of one holdout fit, min-$V_2$ has $60\%$ less regret than holdout ($0.0055$ against $0.0136$; difference $-0.0081$ [$-0.0122$, $-0.0039$]) and $44\%$ less than 2-fold CV, which uses the same two fits ($-0.0043$ [$-0.0067$, $-0.0022$]). It is indistinguishable from 5-fold CV ($-0.0001$ [$-0.0019$, $0.0021$]), which costs $4.99\times$. The advantage over 2-fold CV holds under each noise model separately. Min-$V_2$ uses no held-out labels for the selection step: the two half-size models' disagreement on unlabelled inputs carries more information about the ranking than their errors on the 50\% of noisy labels they did not see.
\item \emph{Overlapping folds hurt $V$, as Remark~\ref{rem:overlap} predicts.} Min-$V_5$ uses more than twice as many models as min-$V_2$ and is worse ($0.0101$ against $0.0055$; $-0.0046$ [$-0.0133$, $0.0006$]). It is no better than 5-fold CV, which uses the same models ($+0.0044$ [$-0.0011$, $+0.0133$]). With overlapping folds, the shared noisy labels hide exactly the expression that $V$ is meant to measure.
\item \emph{Class-dependent noise is the hard case for $V$}, as Section~\ref{sec:sel:beyond} predicts: asymmetric noise creates a bias shared by all candidates, which $V$ cannot see. Here min-$V_2$ ($0.0121$) is behind 5-fold CV ($0.0100$), although still ahead of holdout and 2-fold CV.
\item \emph{Adding a consensus estimate of modal bias gives the best criterion overall.} The consensus criterion \eqref{eq:lf}, computed from two disjoint halves with a reference pool of 19 configurations, has the lowest mean regret ($0.0033$) and chooses the best candidate in 187 of 225 cases. It is significantly better than holdout ($-0.0103$ [$-0.0145$, $-0.0065$]), 2-fold CV ($-0.0065$ [$-0.0099$, $-0.0035$]), min-$V_2$ ($-0.0022$ [$-0.0051$, $-0.0001$]), 5-fold CV ($-0.0023$ [$-0.0047$, $-0.0006$]) and noise-corrected 5-fold CV ($-0.0024$ [$-0.0053$, $-0.0004$]), at under half the cost of 5-fold CV even when the reference pool is included ($2.39\times$ against $4.99\times$). Its gain is largest under class-dependent noise ($0.0057$ against $0.0100$ for 5-fold CV), where the consensus supplies the modal-bias information that $V$ lacks.
\end{enumerate}
Seed 0 alone gives the same ordering with wider intervals, and seeds 1--4 replicate it (mean regret over seed 0 and over seeds 1--4: min-$V_2$ $0.0069$ and $0.0051$, holdout $0.0125$ and $0.0139$, 5-fold CV $0.0035$ and $0.0062$). In summary: among overparameterized candidates, expressed information is a viable and cheap selection criterion. With independent samples it is the best criterion we tested. On a single dataset, estimated from two disjoint halves, it beats holdout and cross-validation of the same cost and matches 5-fold cross-validation; combined with a consensus estimate of modal bias, it beats 5-fold cross-validation at half the cost. The thresholds of Section~\ref{sec:sel:beyond} were set on earlier experiments and frozen before Protocol B.

\subsection{Comparison with selection by conditional mutual information}\label{sec:exp:cmi}

Section~\ref{sec:storage-expression} argued that CMI \citep{steinke2020cmi} measures the wrong thing for interpolators: it bounds the generalization gap, not excess risk. We now compare the two quantities directly as \emph{selection criteria}, on the same data, the same models and the same compute.

\paragraph{Setup.} We use the 225 cases of Protocol B (Section~\ref{sec:exp:select}). The two halves of the 2-fold split form a supersample of pairs $(a_i,b_i)$, one point from each half; the two half-sample models are trained on the members selected by $U$ and on the complement. Evaluated CMI (e-CMI; \citealp{steinke2020cmi}) is estimated per pair as the plug-in mutual information between the pair's loss pattern and the selection bit, and averaged over pairs; functional CMI \citep{harutyunyan2021fcmi} uses the prediction pattern instead and gives the same results here. Besides ranking by e-CMI itself, we rank by two generalization bounds built on it: training error plus $\sqrt{2\,\overline{\text{e-CMI}}}$ \citep{steinke2020cmi}, and training error plus the samplewise bound of \citet{hellstrom2022samplewise}. With one pair of fits, every criterion in the comparison uses exactly the two models of 2-fold CV and of min-$V_2$. We also use two further random splits of the same supersample (three pairs of fits), which lets the CMI estimate average over several draws of $U$, and compare with repeated 2-fold CV and with $V_2$ averaged over the three pairs. The refits reproduce the Protocol B models exactly; details are in Appendix~\ref{sec:add-twostage}.

There is one difference in what the criteria need. CMI conditions on the supersample, labels included, so its estimate uses the label of the held-out member of each pair, as cross-validation does. Min-$V_2$ uses only predictions on unlabelled inputs.

\begin{table}[t]\centering\small
\caption{Selection by CMI against selection by expressed information at matched cost (Protocol B, 225 cases; mean regret in clean test error, lower is better). Columns 3--6: the six overparameterized candidates of Table~\ref{tab:select}, Panel B; ``exact'': cases (of 225) in which the best candidate was chosen; cost: median CPU time relative to one holdout fit; last column: all 19 configurations. ``Labels'': whether the criterion uses held-out labels. Best per column in bold. Functional CMI gives the same regrets as e-CMI among the overparameterized candidates. \prop: proposed in this paper.}\label{tab:cmi}
\resizebox{\linewidth}{!}{%
\begin{tabular}{llccccccc}\toprule
& & \multicolumn{6}{c}{Six overparameterized candidates} & All 19\\
Criterion & Labels & Symmetric & Class-dep. & Instance-dep. & All & Exact & Cost & All\\\midrule
\multicolumn{8}{l}{\emph{One pair of half-sample fits}}\\
2-fold CV & yes & 0.0068 & 0.0173 & 0.0053 & 0.0098 & 164 & 1.24 & 0.0213\\
e-CMI & yes & 0.0056 & 0.0134 & 0.0051 & 0.0080 & 163 & 1.24 & 0.0891\\
Training error $+\sqrt{2\,\overline{\text{e-CMI}}}$ & yes & 0.0065 & 0.0153 & 0.0059 & 0.0092 & 164 & 1.24 & 0.0534\\
Training error $+$ samplewise e-CMI bound & yes & 0.0082 & 0.0150 & 0.0055 & 0.0096 & 163 & 1.24 & 0.0452\\
Min-$V_2$\prop{} & no & 0.0022 & 0.0121 & \best{0.0021} & 0.0055 & 181 & 1.25 & 0.0748\\
\multicolumn{8}{l}{\emph{Three pairs of half-sample fits}}\\
2-fold CV, 3 random splits & yes & 0.0055 & 0.0115 & 0.0048 & 0.0072 & 175 & 3.31 & \best{0.0191}\\
e-CMI & yes & 0.0124 & 0.0309 & 0.0065 & 0.0166 & 145 & 3.31 & 0.0855\\
Training error $+$ samplewise e-CMI bound & yes & 0.0058 & 0.0103 & 0.0044 & 0.0069 & 174 & 3.31 & 0.0493\\
Min-$V_2$\prop{}, averaged over the 3 pairs & no & \best{0.0016} & \best{0.0046} & 0.0030 & \best{0.0031} & \best{185} & 3.36 & 0.0650\\
\multicolumn{8}{l}{\emph{Reference}}\\
5-fold CV & yes & 0.0036 & 0.0100 & 0.0033 & 0.0056 & 177 & 4.99 & 0.0196\\
\bottomrule
\end{tabular}}
\end{table}

\paragraph{Results (Table~\ref{tab:cmi}).} Min-$V$ is the better criterion, and it needs no held-out labels.
\begin{enumerate}[leftmargin=*]
\item \emph{At the cost of two fits, min-$V_2$ beats every CMI criterion.} Its regret is $0.0055$, against $0.0080$ for e-CMI ($-0.0025$ [$-0.0052$, $-0.0001$]), $0.0092$ for the $\sqrt{2\,\overline{\text{e-CMI}}}$ bound ($-0.0037$ [$-0.0066$, $-0.0013$]) and $0.0096$ for the samplewise bound ($-0.0041$ [$-0.0077$, $-0.0012$]). It chooses the best candidate in 181 of 225 cases, against 163--164 for the CMI criteria. It is ahead under each noise model.
\item \emph{CMI behaves like cross-validation, because it is a function of held-out losses.} For an interpolator the loss on the training member of a pair is zero, so the e-CMI estimate from one pair of fits is $\ln2$ times the probability that at least one of the pair's two held-out predictions is wrong. Its regret is indeed close to that of 2-fold CV on the same fits ($0.0080$ against $0.0098$; $-0.0018$ [$-0.0051$, $+0.0014$]). So CMI brings no information beyond the held-out labels it consumes, whereas $V_2$ replaces them by unlabelled inputs and does better.
\item \emph{With more fits, the advantage of $V$ grows.} With three pairs of fits, $V_2$ averaged over the pairs has regret $0.0031$, the lowest in the table, against $0.0069$ for the best CMI criterion ($-0.0038$ [$-0.0071$, $-0.0013$]) and $0.0166$ for e-CMI itself, whose plug-in estimate is more biased with more draws. It also beats 5-fold CV ($0.0056$; $-0.0026$ [$-0.0058$, $-0.0004$]) at two thirds of its cost.
\item \emph{In a pool with rigid learners, neither criterion is enough on its own.} With all 19 configurations, e-CMI has regret $0.089$ and min-$V_2$ $0.075$. Both prefer rigid learners: CMI because their generalization gap is small whatever their error, $V$ because their predictions hardly depend on the sample whatever their modal bias. Adding the training error, as the CMI bounds do, helps ($0.045$--$0.053$) but stays far behind cross-validation, because the bounds are loose (for interpolators, $\sqrt{2\,\overline{\text{e-CMI}}}\approx0.8$). Section~\ref{sec:exp:beyond} discusses how to handle such pools.
\end{enumerate}
In summary, among overparameterized candidates and at matched cost, expressed information is both \emph{label-free} and \emph{more accurate} than CMI as a selection criterion. This is the empirical counterpart of Corollary~\ref{cor:cmi} and Proposition~\ref{prop:supersample}: CMI measures what the model reveals about \emph{which} examples it was trained on, while $V$ measures what reaches its predictions at fresh inputs, and only the latter is tied to excess risk.

\subsection{Beyond overparameterized candidates}\label{sec:exp:beyond}

Finally, we test what happens when the candidate pool also contains underparameterized learners, for which modal bias is not negligible (Section~\ref{sec:sel:beyond}). We use all 19 configurations under both protocols.

\begin{figure}[t]\centering
\includegraphics[width=\linewidth]{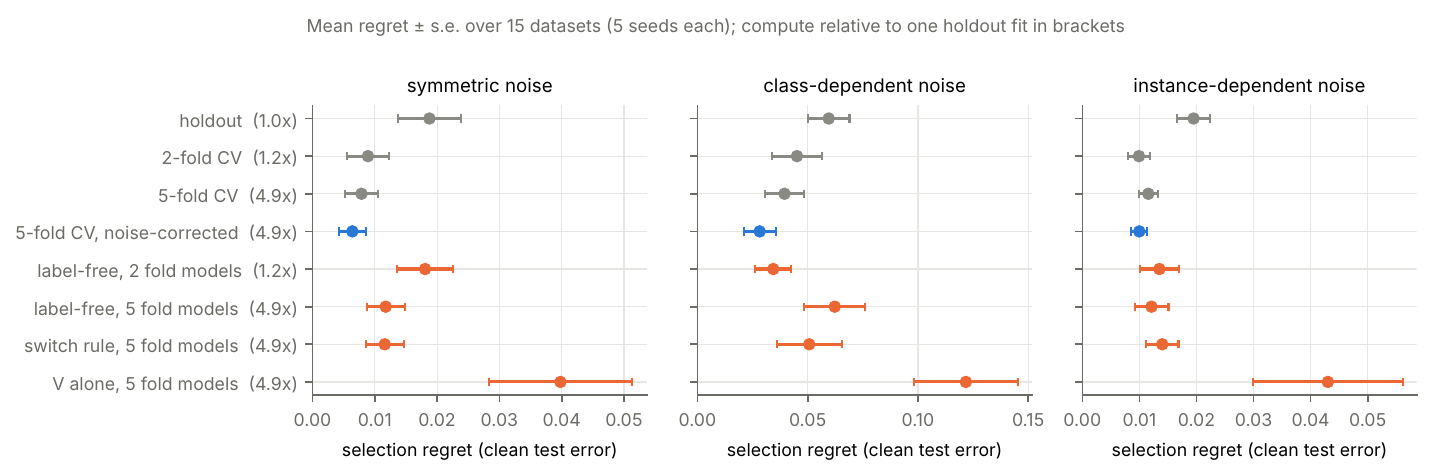}
\caption{Protocol B with all 19 configurations (15 datasets $\times$ 5 seeds per noise model; error bars: standard error over datasets). Grey: labelled criteria; dark blue: noise-corrected cross-validation; orange: criteria based on expressed information, computed from the cross-validation fold models. Bracketed: median compute relative to one holdout fit.}\label{fig:cost}
\end{figure}

\paragraph{Min-$V$ alone fails, as expected.} With all candidates, $V$ alone prefers rigid learners whose predictions hardly depend on the sample, even when their modal predictions are wrong. On one dataset (Protocol B; Figure~\ref{fig:cost}), min-$V_5$ has regret $0.068$ and min-$V_2$ $0.075$, against $0.020$ for 5-fold CV. This is the failure that Proposition~\ref{prop:rank} warns of, and the reason we recommend min-$V$ only among overparameterized candidates.

\paragraph{The extensions help, but on one dataset cross-validation remains best.} With independent samples (Protocol A; Appendix~\ref{sec:add-select}), the consensus criterion \eqref{eq:lf} has the lowest mean regret of all criteria, $0.0085$ against $0.0219$ for holdout, and it matches holdout with \emph{clean} validation labels under symmetric noise ($0.0037$ against $0.0040$). Under class-dependent noise, the switch rule of Section~\ref{sec:sel:beyond} detects the asymmetry and improves on both of its components (regret $0.0106$; overall $0.0060$). On one dataset (Protocol B, 225 cases), however, noise-corrected 5-fold CV has the lowest mean regret ($0.0149$), significantly below 5-fold CV ($0.0196$; $-0.0048$ [$-0.0073$, $-0.0025$]), with almost all of the gain under class-dependent noise. The consensus criterion from two halves ($0.0220$) and the switch rule built on it ($0.0217$) cost $1.21\times$ holdout, are significantly better than holdout ($0.0326$; $-0.0106$ [$-0.0187$, $-0.0028$]), and are on a par with 2-fold CV ($0.0213$), but they do not reach 5-fold or noise-corrected CV. The consensus criterion from five overlapping folds is worse ($0.0287$), for the reason given in Remark~\ref{rem:overlap}.

\paragraph{Screening by training fit first does not help.} A natural alternative is a two-stage rule: keep the candidates that fit the training data, for example the interpolators, and choose among them by min-$V_2$. This lowers the regret of min-$V_2$ in the 19-candidate pool from $0.075$ to $0.046$, but it stays behind holdout and cross-validation. Under label noise the best candidate is usually a regularized learner that does not interpolate, and the screen removes it in $85\%$ of the cases; min-$V$ then chooses well among interpolators, but the best interpolator is not the best candidate. Appendix~\ref{sec:add-twostage} gives the details, other screens (including the legacy LMC score), and the same finding with independent samples.

With a diverse pool and one dataset, our recommendation is therefore noise-corrected cross-validation, with expressed information computed from its fold models as a nearly free diagnostic (about $1\%$ of the cost). A stress test with a controlled share of redundant, equally biased candidates, and with growing asymmetric noise, confirms the two failure modes predicted by the theory (Appendix~\ref{sec:stress}).

%% file: sections/s8_related.tex
\section{Connections to existing work}\label{sec:related}

Throughout the paper we have pointed out connections as they arose. Let us now collect them, and say for each line of work what the present framework shares with it and what it adds.

\paragraph{Information bottleneck.} The IB method describes a good representation as one that keeps as little information about the input as possible while keeping a prescribed amount of information about the label \citep{tishby2000ib}. It has been proposed as a theory of deep learning \citep{tishby2015deep,shwartzziv2017opening} and as a training objective \citep{achille2018emergence,strouse2017dib}. Our framework relates to it in five ways.
\begin{enumerate}[label=(\roman*),leftmargin=*]
\item \emph{Where the ideal model sits.} Under log-loss, the indirect rate--distortion function \eqref{eq:irdf} is the IB curve, the minimal sufficient complexity $C^*_{\log}=H(\eta(X))$ is its full-relevance end point ($\beta\to\infty$ in the IB Lagrangian $I(T;X)-\beta I(T;Y)$), and the G-LMC is the minimal sufficient statistic (Section~\ref{sec:rd}). For other losses, \eqref{eq:irdf} is a loss-specific bottleneck whose end point is $H(a^*(X))$. Under 0-1 loss it ends at $H(b^*(X))$, which is smaller than $H(\eta(X))$ under label noise. So a classifier does not need to compress to the log-loss IB end point; this distinction disappears for noiseless labels, the only case the original LMC covered. Restricted to deterministic representations, our complexity $I(F;X)=H(F)$ is the objective of the deterministic IB \citep{strouse2017dib}.
\item \emph{Excess complexity as distance to the curve.} Excess complexity (Definition~\ref{def:complexities}) is the vertical distance of a model above the (loss-specific) IB curve at the risk it achieves, i.e.\ its redundancy. Proposition~\ref{prop:split} shows that, for a learner's output at fresh inputs, total output entropy splits into this complexity and expressed information, so an information-plane coordinate computed for a learner mixes the two.
\item \emph{Estimation and compression phases.} For deterministic maps of continuous inputs, $I(T;X)$ is degenerate or infinite, so information-plane analyses depend on binning or noise \citep{amjad2019learning,kolchinsky2019caveats,goldfeld2019estimating}, and compression phases depend on the estimator and the activation function \citep{saxe2018ib}. Our ideal-model quantities are defined for discrete variables, where they are finite, and $V$ is an entropy over training samples at a fixed input, which needs no binning. Our boosting experiment (Section~\ref{sec:exp:boost}) agrees with \citet{saxe2018ib} that compression in the information plane need not track generalization: under label noise, an apparent compression coincides with a $4.5$-fold increase in excess risk.
\item \emph{Information in the weights.} \citet{achille2018emergence} argued that limiting the information that the weights store about the training set, $I(W;S)$, favours invariant representations and good generalization. Theorem~\ref{thm:storage} shows that this cannot be a necessary condition for generalization: every interpolator, benign ones included, stores at least $nH(Y\mid X)$ bits. In our terms, what matters is not how much is stored in the weights but how much reaches predictions.
\item \emph{IB-based generalization bounds.} \citet{shamir2010learning} bound the finite-sample error of IB solutions, and \citet{kawaguchi2023ib} bound the generalization gap of a network through $I(X;Z\mid Y)$ for an intermediate representation $Z$, among other terms. Lemma~\ref{lem:noisecomp} shows that under label noise this quantity cannot vanish even for the ideal representation: a log-loss sufficient output must retain at least $H(\eta(X)\mid Y)>0$ bits about the input beyond the label, so such bounds have an irreducible term set by the noise.
\end{enumerate}

\paragraph{Decision-theoretic sufficiency.} $\ell$-sufficiency and $\ell$-deficiency (Section~\ref{sec:suff}) are the sufficiency and deficiency of \citet{blackwell1953equivalent} and \citet{lecam1964sufficiency}, specialized to a single decision problem. The characterization in Proposition~\ref{prop:suff}, that sufficiency means the Bayes act is recoverable, and the identification of the minimal sufficient complexity with $H(a^*(X))$ are elementary in that setting; what they give us is a loss-specific notion of ``minimal model'' that does not require the label to be a function of the input, unlike the original LMC (Proposition~\ref{prop:gk}).

\paragraph{Information-theoretic generalization bounds, CMI and memorization.} Bounds based on $I(W;S)$ \citep{russo2016controlling,xu2017information} control the expected generalization gap, but $I(W;S)$ can be infinite, for instance for deterministic learners with continuous parameters. \citet{steinke2020cmi} introduced the conditional mutual information $\mathrm{CMI}=I(W;U\mid\tilde Z)$ of a supersample, which is always at most $n\ln2$, recovers VC-type and compression-based bounds, and gives $|\E\,\mathrm{gap}|\le\sqrt{2\,\mathrm{CMI}/n}$. Variants evaluate the information in the losses or the predictions at the supersample points instead of the weights (functional CMI; \citealp{harutyunyan2021fcmi}). Several works show that such bounds must be large for some learners \citep{bassily2018learners,livni2023lower,haghifam2023limitations}, and that memorization of labels is sometimes necessary for good accuracy \citep{feldman2020memorization,brown2021memorization}. \citet{harutyunyan2020label} relate the information stored about the labels to the number of training errors; Theorem~\ref{thm:storage} sharpens their result into an equality for interpolators. We connect to this line of work in three ways.
\begin{enumerate}[label=(\roman*),leftmargin=*]
\item \emph{Gap versus excess risk.} All of these quantities bound the gap. For an interpolator the gap is its whole test error, at least $R^*$, for benign and tempered learners alike. So $I(W;S)\ge nH(Y\mid X)$ (Corollary~\ref{cor:xr}) and $\mathrm{CMI}\ge\frac n2(R^*)^2$ (Corollary~\ref{cor:cmi}): both grow linearly in $n$ for every interpolator, and neither can separate a benign interpolator from a tempered one. The separating quantity must concern excess risk, which requires looking at predictions at fresh inputs.
\item \emph{CMI at fresh inputs.} Proposition~\ref{prop:supersample} places $V$ in the supersample construction. $V(x)$ splits into the information the prediction at a fresh input carries about the pool $\tilde Z$, about the selection $U$ given the pool, and about the seed. The middle term is the fresh-input, prediction-level analogue of CMI. Functional CMI evaluates predictions at the $2n$ supersample points, half of which are training points, in order to bound the gap; $V$ evaluates them at new points and characterizes excess risk (Theorem~\ref{thm:decomp}).
\item \emph{Estimation.} The supersample construction is also how we estimate $V$ from one dataset: the two copies trained on disjoint halves are the models trained on $\tilde Z_U$ and $\tilde Z_{1-U}$ (Section~\ref{sec:sel:est}). The failure of overlapping folds (Remark~\ref{rem:overlap}) is the observation that, without this disjointness, the copies share noisy labels and the selection is no longer what distinguishes them.
\item \emph{Selection.} Evaluated and functional CMI can be estimated from the same pairs of half-sample fits as $V_2$, so the two can be compared as selection criteria at equal cost (Section~\ref{sec:exp:cmi}). CMI conditions on the labels of the held-out member of each pair, so it is not label-free, and for interpolators its estimate from one pair of fits ranks candidates like 2-fold cross-validation. Min-$V_2$ needs no held-out labels and, among overparameterized candidates, has lower regret than every CMI criterion computed from the same fits; in a pool with rigid learners both fail, since rigid learners have a small gap and a small $V$ whatever their error.
\end{enumerate}

\paragraph{Benign overfitting and double descent.} Benign overfitting has been established for minimum-norm least squares \citep{bartlett2020benign,hastie2022surprises,muthukumar2020harmless}, for local interpolating rules \citep{belkin2019interpolation}, and classified into benign, tempered and catastrophic regimes \citep{mallinar2022taxonomy}. Double descent was described by \citet{belkin2019reconciling} and analysed through bias--variance decompositions \citep{adlam2020understanding,yang2020rethinking,mei2022random}. Section~\ref{sec:benign} recovers several of these results through one quantity: the 1-NN limit of \citet{cover1967nearest} (Proposition~\ref{prop:knn}), the variance term of \citet{bartlett2020benign} (Proposition~\ref{prop:ridgeless}), the taxonomy of \citet{mallinar2022taxonomy} (Section~\ref{sec:taxonomy}), and double descent as a peak of expression. The ``interpolation as a combination of a smooth component and spiky memorization'' picture of \citet{wyner2017explaining} and \citet{belkin2019interpolation} is exactly what memorize-and-smooth makes explicit (Proposition~\ref{prop:memsmooth}).

\paragraph{Bias--variance decompositions for classification.} Section~\ref{sec:bvd} makes the relation exact, and Table~\ref{tab:dictionary} summarizes it in a full-page dictionary that maps each existing notion to its counterpart in this paper. The identity behind Theorem~\ref{thm:decomp} is the excess-risk form of the 0-1 decompositions of \citet{kohavi1996bias}, \citet{domingos2000unified} and \citet{james2003variance}: Domingos' main prediction is our modal prediction, his bias is our modal bias, his variance is the off-mode probability $\pi_D$, and James' systematic and variance effects add up to our identity \eqref{eq:exact} (Proposition~\ref{prop:dictionary}). Kohavi and Wolpert's variance is half the disagreement of two copies, and all of these variances are equivalent to $V$ up to the explicit bounds of Proposition~\ref{prop:dictionary}. What $V$ adds is its reading as a mutual information with the training sample, which connects variance to storage (Theorem~\ref{thm:storage}) and to output complexity (Proposition~\ref{prop:split}). It also gives matching upper and lower bounds for any number of classes, and its chain rule separates sample and seed contributions, as in the fine-grained decomposition of \citet{adlam2020understanding}. For squared loss, the classical variance is a limit of $V$ (Theorem~\ref{thm:sq}); for log-loss, the variance of \citet{heskes1998bias} is the analogous quantity (Appendix~\ref{sec:logloss}).

\paragraph{Epistemic uncertainty.} In Bayesian deep learning, the mutual information between a prediction and the model parameters under the posterior measures epistemic uncertainty \citep{depeweg2018decomposition}. $V$ is its frequentist counterpart for hard predictions: the randomness comes from the training sample rather than from a posterior. Ensembles of independently trained models, which estimate $V$, are also a common practical estimate of epistemic uncertainty.

\paragraph{Disagreement and agreement-based model assessment.} The disagreement of two independently trained networks tracks their test error \citep{jiang2022disagreement}; agreement between models predicts accuracy under distribution shift \citep{baek2022agreement}; and agreement rates among several classifiers can estimate their accuracies without labels \citep{platanios2016bayesian}. Proposition~\ref{prop:disagreement-main} and Table~\ref{tab:dis} explain what disagreement measures. It bounds \emph{excess} risk from above and below when modal bias is absent. It equals total risk only when the predictions are distributed like a fresh label: this is the case for tempered interpolators such as 1-NN, which show the distributional generalization of \citet{nakkiran2020distributional}, and not for benign interpolators under label noise, for which the GDE fails by exactly the Bayes risk. Seed-only disagreement, as used for the GDE, misses the sample part of $V$ altogether. Proposition~\ref{prop:rank} turns disagreement-type quantities into a ranking guarantee, and the consensus criterion (Section~\ref{sec:sel:beyond}) uses agreement for a different purpose, to estimate modal bias, with a guarantee that needs no independence assumption.

\paragraph{Learning with noisy labels and model selection.} Backward loss correction with a known or estimated transition matrix gives unbiased risk estimates under class-conditional noise \citep{natarajan2013noisy,patrini2017making}, and we use it in the switch rule of Section~\ref{sec:sel:beyond}. Our observation that correction cannot change the ranking under symmetric noise (Remark~\ref{rem:symcorr}) explains why it helps only when noise is asymmetric. Cross-validation is the standard tool for model selection \citep{arlot2010survey}; our experiments show that, among overparameterized candidates, expressed information estimated from 2-fold fold models adds a label-free criterion at almost no extra cost.

\paragraph{Margins and the LMC papers.} \citet{nikolaou2020margin} and \citet{nikolaou2021lossless} related margin maximization in boosting to lossless maximal compression; margin-based explanations of boosting go back to \citet{schapire1998boosting}, and the noise sensitivity of boosting is well known \citep{long2010random}. Section~\ref{sec:lmc} and Appendix~\ref{sec:erratum} explain what in that work does and does not survive. In particular, the result that log-loss sufficiency of a binary score is equivalent to its recoverability of $\eta$ (our Proposition~\ref{prop:suff}) survives, now stated for the population and for stochastic representations; the claims that do not survive are the existence of a lossless maximal compressor under label noise, the inheritance of margin guarantees, and the reading of information-plane ``compression'' as a sign that early stopping is unnecessary.

%% file: sections/s9_discussion.tex
\section{Discussion and conclusion}\label{sec:discussion}

Let us summarize what we have done, discuss uses of expressed information beyond the settings tested here, and then acknowledge the limitations of the framework and of the experiments, together with the steps we see for addressing them.

\paragraph{Summary.} We started from the ideal model (Section~\ref{sec:sufficiency}): for any loss, the least complex optimal representation is the Bayes act, its entropy is the minimal sufficient complexity, and anything above it is excess complexity. We then separated what a learner stores from what it expresses (Section~\ref{sec:storage-expression}). Storage is pinned at $n\,H(Y\mid X)$ for every interpolator, which is why storage-based bounds and training-set information measures cannot tell interpolators apart. Expression, measured by $V$, controls excess risk together with modal bias, and a consistent learner's outputs converge to the minimal sufficient representation even while it stores all the noise. Benign overfitting is therefore vanishing expression (Section~\ref{sec:benign}), a view that recovers classical and recent results for nearest neighbours, averaging and minimum-norm interpolation, and places double descent as a peak of expression. Finally (Sections~\ref{sec:selection} and~\ref{sec:experiments}), $V$ ranks overparameterized candidates, and it gives a viable label-free selection rule: with independent samples it beats holdout validation, and on a single dataset, estimated from disjoint halves, it beats holdout and 2-fold cross-validation at the same cost and matches 5-fold cross-validation. Section~\ref{sec:bvd} relates all of this to bias--variance decompositions and to disagreement.

\paragraph{Selection without ground truth.} Min-$V$ uses the training labels to fit the candidates, but no held-out labels: it needs only unlabelled inputs from the distribution on which the model will be used. This is an advantage wherever labels for validation are scarce, costly, delayed or untrustworthy while inputs are plentiful.
\begin{itemize}[leftmargin=*]
\item In \emph{semi-supervised} settings, a small labelled set is better spent on training than on validation, and the large unlabelled pool is exactly what $V$ needs.
\item In \emph{streaming} or online settings, labels often arrive late or not at all, while incoming inputs can be scored at once.
\item Under \emph{label noise}, validation labels are as noisy as the training labels (Section~\ref{sec:sel:why}). Panel A of Table~\ref{tab:select} shows the consequence: when the validation set shrinks from its full size to 40 points, the regret of holdout roughly triples, while min-$V$, which uses no validation labels, is unaffected.
\item Under \emph{covariate shift}, $V$ can be computed on unlabelled inputs from the target distribution itself.
\end{itemize}
The idea also has an unsupervised counterpart. Stability-based selection of the number of clusters fits a clustering method on two halves of the data and compares the two assignments on common points \citep{benhur2002stability,lange2004stability,tibshirani2005prediction,vonluxburg2010clustering}; min-$V_2$ is the same procedure applied to supervised learners. The known failure of clustering stability, that a stable but wrong solution such as too few clusters can be preferred \citep{bendavid2006sober}, is the unsupervised counterpart of modal bias, and Theorem~\ref{thm:decomp} makes this failure precise for classification. What the unsupervised case lacks is an analogue of the regime in which modal bias is small, and hence an analogue of Proposition~\ref{prop:rank}; we leave this open.

\paragraph{Limitations and directions for future work.} The framework and the experiments have limitations. We list them below, each with the steps we see for addressing it.
\begin{itemize}[leftmargin=*]
\item \emph{From a property of the learner to a property of a trained model.} $V$ is defined over independent training samples, so it describes the learner, not one trained model; a single model's excess risk over its learner's modal prediction is controlled by $V$ only in probability (Appendix~\ref{sec:single}). A natural route to a per-model quantity is the chain rule over the parts of the sample,
\[
V(x)\;=\;I\big(h_W(x);X^n\big)+I\big(h_W(x);Y^n\mid X^n\big)+I\big(h_W(x);\xi\mid S\big).
\]
The middle term, which we would call the \emph{label-expressed information}, fixes the training inputs and asks how much the prediction at $x$ depends on the particular labels the model was given; it is the closest formal counterpart of ``how much this model uses the training labels when it predicts at a new point''. For linear smoothers (nearest-neighbour rules, kernel ridge(less) regression, minimum-norm least squares, and forests viewed as adaptive nearest neighbours), the prediction is $\sum_iw_i(x)\,y_i$ with weights that are fixed once the model is trained, and the term can be computed from the fitted model itself: Proposition~\ref{prop:ridgeless} is exactly this computation for minimum-norm least squares, where it is a function of $\|(X^+)^\top x\|^2$. For general models such as deep networks, the weights $w_i(x)$ can be approximated by influence functions \citep{koh2017influence}, by linearization around the trained weights, or by datamodels \citep{ilyas2022datamodels}, or the term can be estimated by briefly fine-tuning the trained model on resampled labels; the per-example influence estimates of \citet{feldman2020neural} are a related per-model quantity. What remains to be done is to prove a conditional version of Theorem~\ref{thm:decomp} given the training inputs, and to test such estimators against the learner-level $V$.
\item \emph{Estimating modal bias.} $V$ does not see modal bias, and Proposition~\ref{prop:rank} assumes it away. We argued that this is reasonable among flexible overparameterized learners and measured it to be small in our synthetic experiments, but it cannot be checked directly on real data, and it fails for underfitting candidates and under asymmetric label noise. The consensus criterion and the switch rule of Section~\ref{sec:sel:beyond} address part of this. A next step is to estimate modal bias without a consensus, for instance from how $V$ changes with the sample size, or with a small labelled set used only to screen out biased candidates.
\item \emph{Continuous inputs and real-valued outputs.} The ideal-model quantities of Section~\ref{sec:sufficiency} assume discrete inputs: for continuous inputs, $H(X)$ and $I(F;X)$ for deterministic $F$ can be infinite. The learner-level quantities ($V$ and storage) do not need this assumption, and Theorem~\ref{thm:sq} handles continuous outputs by smoothing. For regression, the storage result carries over in a stronger form, but selection by variance works only when the candidates have similar bias (Appendix~\ref{sec:regression}), because interpolation does not keep bias small in regression. Extending the ideal-model quantities to continuous inputs through quantization or rate--distortion, and finding a reliable bias estimate for regression, are open problems.
\item \emph{Rates.} We give a rate of vanishing expression for $k_n$-NN only (Proposition~\ref{prop:knn-rate}). Deriving rates for kernel machines, forests and networks is a natural next step. For kernel ridge(less) regression and for networks in the lazy regime, the expressed information of minimum-norm interpolators can be computed exactly, as in Proposition~\ref{prop:ridgeless}, which makes these the most promising cases.
\item \emph{Scale of the experiments.} All experiments run on CPUs, on datasets of at most a few thousand points and with small networks, so whether min-$V$ selects well among large deep networks is not yet tested. The disagreement results of \citet{jiang2022disagreement} suggest that it may, since disagreement is a function of $V$ for binary labels. The next experiment is to choose among networks of different widths and training lengths on a benchmark with real label noise, such as CIFAR-10N, at a scale where the networks approach low modal bias.
\item \emph{Estimation from one dataset.} On one dataset, $V$ must be estimated from copies trained on disjoint parts of the data, which halves the training set of each copy and measures expression at half the sample size; with two copies the estimate is coarse. Averaging over several disjoint pairs already helps (Section~\ref{sec:exp:cmi}). Correcting for the change of sample size, and using several independent noisy labellings where they exist, are further steps.
\item \emph{Statistical dependence and tuning.} In Protocol B we account for the correlation between seeds of the same dataset with a cluster bootstrap, but the 15 datasets are a convenience sample of small tabular benchmarks, and in Protocol A cases from the same dataset are correlated, so those intervals are optimistic. The thresholds of the switch rule and of the clustering step were chosen on early experiments and then frozen. Validating them, and the whole comparison, on a larger and preregistered collection of datasets would strengthen the conclusions.
\item \emph{Training for low expression.} Since $V$ needs no labels, it can be estimated on unlabelled data during training. Using it as a regularizer for ensembles or as an early-stopping signal is a further direction that we have not explored.
\end{itemize}

\paragraph{Conclusion.} An interpolating learner memorizes all the noise in its training labels; whether it overfits benignly depends on whether that noise reaches its predictions. Expressed information measures exactly this, and it can be estimated without validation labels. It explains benign, tempered and catastrophic overfitting in one framework, and among overparameterized learners it is a practical criterion for choosing between them.

%% file: appendix.tex
\noindent These appendices form the supplementary material. Results stated only here are numbered S1, S2, \dots, as are the tables, figures and equations of the appendices. They are organized as follows.
\begin{itemize}[leftmargin=*,itemsep=0pt]
\item Section~\ref{sec:assumptions} states the regularity assumptions.
\item Sections~\ref{sec:proofs-suff}--\ref{sec:proofs-benign} give all proofs, in the order of the main text, together with results that we only mention there (held-out estimation of excess complexity, log-loss, single trained models, stability).
\item Section~\ref{sec:consensus} gives the theory of the consensus criterion and of noise correction (Section~\ref{sec:sel:beyond}).
\item Section~\ref{sec:erratum} is an erratum for the two earlier papers on lossless maximal compression.
\item Sections~\ref{sec:expdetails} and~\ref{sec:addresults} give experimental details and additional results.
\item Section~\ref{sec:regression} extends the storage result and the selection rule to regression, with one experiment.
\item Section~\ref{sec:code} describes the code package.
\end{itemize}

\section{Regularity assumptions}\label{sec:assumptions}

Recall the setting of Section~\ref{sec:setting}: $(X,Y)\sim P$, a loss $\ell:\cA\times\cY\to[0,\infty]$ on an action set $\cA$, $L(a,q)=\sum_kq_k\ell(a,k)$, and $L^*(q)=\inf_aL(a,q)$. In Section~\ref{sec:sufficiency} we use the following two assumptions.

\begin{assumption}[Regularity]\label{as:reg}
$\cA$ is a compact metric space, and $a\mapsto\ell(a,k)$ is lower semicontinuous for every $k$. Hence every $L(\cdot,q)$ attains its infimum.
\end{assumption}

\begin{assumption}[Unique Bayes act]\label{as:unique}
For $P_X$-almost every $x$, $L(\cdot,\eta(x))$ has a unique minimizer, denoted $a^*(x)$.
\end{assumption}

Under log-loss, $\cA=\Delta_K$ and $L(q,\eta)=H(\eta)+\KL(\eta\|q)$, so both assumptions hold (with the convention $-\log0=\infty$), with $a^*=\eta$ and $R_{\log}(T)=H(Y\mid T)$. Under 0-1 loss, $\cA=[K]$, $a^*=b^*$, and Assumption~\ref{as:unique} means that there are no ties, i.e.\ $\gamma(x)>0$ almost surely. A representation $T$ has countable range and satisfies the Markov chain $Y - X - T$.

\section{Proofs for main Section~\ref{sec:sufficiency}}\label{sec:proofs-suff}

\begin{lemma}[Recalibrated risk]\label{lem:recal}
Let $\eta_T(t)=\P(Y=\cdot\mid T=t)$. Then $R_\ell(T)=\sum_t\P(T=t)\,L^*(\eta_T(t))$, the infimum is attained by any $g^*(t)\in\argmin_aL(a,\eta_T(t))$, and $R_\ell(T)\ge R^*_\ell$.
\end{lemma}
\begin{proof}
By $Y - X - T$, $\E[\ell(g(T),Y)\mid X,T]=L(g(T),\eta(X))$, and $\E[\ell(g(T),Y)\mid T=t]=L(g(t),\eta_T(t))$. Minimizing pointwise in $t$ gives the formula; minimizers exist by Assumption~\ref{as:reg}. For the inequality, $\E\,\ell(g^*(T),Y)=\E\,L(g^*(T),\eta(X))\ge\E\,L^*(\eta(X))=R^*_\ell$.
\end{proof}

\begin{proof}[Proof of Proposition~\ref{prop:suff}]
($\Leftarrow$) If $a^*(X)=g(T)$, then $\E\,\ell(g(T),Y)=\E\,L(a^*(X),\eta(X))=R^*_\ell$.
($\Rightarrow$) Take $g^*$ from Lemma~\ref{lem:recal}. Then $0=R_\ell(T)-R^*_\ell=\E\big[L(g^*(T),\eta(X))-L^*(\eta(X))\big]$, with a non-negative integrand, so $g^*(T)$ minimizes $L(\cdot,\eta(X))$ almost surely. By Assumption~\ref{as:unique}, $g^*(T)=a^*(X)$ almost surely.
Under log-loss, $R_{\log}(T)=H(Y\mid T)\ge H(Y\mid X)=R^*_{\log}$, with equality iff $I(T;Y)=I(X;Y)$; apply the first part with $a^*=\eta$. For $K=2$ this is Lemma~3 of \citet{nikolaou2020margin}, here stated for the population and for stochastic $T$.
\end{proof}

\begin{proof}[Proof of Theorem~\ref{thm:minimal}]
Since $a^*(X)$ is a function of $X$, we have $T - X - a^*(X)$, and the data-processing inequality gives $I(T;X)\ge I(T;a^*(X))=H(a^*(X))-H(a^*(X)\mid T)=H(a^*(X))$, where the last step uses Proposition~\ref{prop:suff} ($a^*(X)$ is a function of $T$). Equality in data processing holds iff $I(T;X\mid a^*(X))=0$, i.e.\ iff $T - a^*(X) - X$. For deterministic $F$: $H(F)=H(F,a^*(X))=H(a^*(X))+H(F\mid a^*(X))$, which equals $H(a^*(X))$ iff $F$ is a function of $a^*(X)$; since $a^*(X)$ is also a function of $F$, the two are in bijection on their supports. Finally, $T=a^*(X)$ is sufficient with $I(T;X)=C^*_\ell$, so $\cR_\ell(R^*_\ell)=C^*_\ell$.
\end{proof}

\begin{proof}[Proof of Proposition~\ref{prop:excess}]
If $F$ is sufficient then $R_\ell(F)=R^*_\ell$ and $\cR_\ell(R^*_\ell)=C^*_\ell$, so $\cE_\ell(F)=H(F)-H(a^*(X))=H(F\mid a^*(X))$ by the proof of Theorem~\ref{thm:minimal}. Under log-loss $a^*=\eta$ and $\eta(X)=g(F)$, so $H(F\mid Y)=H(F,\eta(X)\mid Y)=H(\eta(X)\mid Y)+H(F\mid\eta(X),Y)$. Since $\P(Y\mid X)=\eta(X)=\P(Y\mid\eta(X))$, we have $Y\perp X\mid\eta(X)$, hence $Y\perp F\mid\eta(X)$, so $H(F\mid\eta(X),Y)=H(F\mid\eta(X))=\cE_{\log}(F)$.
\end{proof}

\subsection{Held-out estimation of excess complexity}\label{sec:heldout}

The log-loss identity of Proposition~\ref{prop:excess} suggests estimating excess complexity by $H(F\mid Y)$ on held-out data (minus the task constant $H(\eta(X)\mid Y)$). For non-sufficient outputs this is only a lower bound.

\begin{proposition}[The held-out estimator is a lower bound]\label{prop:g7lb}
For any representation $T$ with $Y - X - T$, under log-loss,
\[
H(T\mid Y)-H(\eta(X)\mid Y)\;=\;H(T\mid\eta(X))-H(\eta(X)\mid T,Y)\;\le\;H(T\mid\eta(X)),
\]
with equality iff $\eta(X)$ is a function of $(T,Y)$; in particular for every sufficient $T$.
\end{proposition}
\begin{proof}
$H(T\mid Y)=H(T,\eta(X)\mid Y)-H(\eta(X)\mid T,Y)=H(\eta(X)\mid Y)+H(T\mid\eta(X),Y)-H(\eta(X)\mid T,Y)$, and $H(T\mid\eta(X),Y)=H(T\mid\eta(X))$ because $Y\perp T\mid\eta(X)$.
\end{proof}

This is why, in Section~\ref{sec:exp:synth}, the held-out estimator ranks learners reasonably (Spearman $0.85$) but underestimates by $0.35$ bits on average: most trained outputs there are not log-loss sufficient. The task constant is strictly positive under label noise:

\begin{lemma}[Irreducible noise complexity]\label{lem:noisecomp}
For deterministic $F$, $I(F;X)=I(F;Y)+I(F;X\mid Y)$. If $F$ is log-loss sufficient then $I(F;X\mid Y)=H(F\mid Y)\ge H(\eta(X)\mid Y)$. If $K=2$, $I(X;Y)>0$ and $H(Y\mid X)>0$, then $H(\eta(X)\mid Y)>0$.
\end{lemma}
\begin{proof}
$I(F;X)=I(F;X,Y)$ because $I(F;Y\mid X)=0$; then apply the chain rule. The inequality follows from Proposition~\ref{prop:excess}. For the last claim, let $\eta_1=\P(Y=1\mid X)$. Since $I(X;Y)>0$, $\eta_1(X)$ takes at least two values $p\ne q$, and since $H(Y\mid X)>0$ we may take $p\in(0,1)$. If $q\in(0,1)$, both labels occur under both values. If $q=0$, label $0$ occurs under both; if $q=1$, label $1$ does. Either way, some $y$ has $\P(\eta_1(X)=p\mid Y=y)\in(0,1)$, so $H(\eta(X)\mid Y)>0$.
\end{proof}

So a log-loss sufficient model must retain at least $H(\eta(X)\mid Y)>0$ bits about its input beyond what the label explains. This is another way to see why the original LMC condition $I(F;X)=I(F;Y)$ cannot hold under noise.

\subsection{The original LMC}\label{sec:degeneracy}

\begin{proof}[Proof of Proposition~\ref{prop:gk}]
Let $C_{GK}(X;Y)=\max\{H(G):G=g_1(X)=g_2(Y)\text{ a.s.}\}$. Always $C_{GK}\le I(X;Y)$: if $G=g_1(X)=g_2(Y)$ then $I(X;Y)\ge I(G;Y)\ge I(G;G)=H(G)$.
($\Rightarrow$) $I(F;Y)=H(F)-H(F\mid Y)$ and $I(F;X)=H(F)$, so $I(F;X)=I(F;Y)$ forces $H(F\mid Y)=0$, i.e.\ $F=g_2(Y)$ a.s. Thus $F$ is common to $X$ and $Y$, and $I(X;Y)=H(F)\le C_{GK}$.
($\Leftarrow$) Take $G$ attaining $C_{GK}$ and $F=g_1(X)$; then $I(F;X)=H(G)=I(F;Y)=C_{GK}=I(X;Y)$.
For $K=2$, $g_2(Y)$ is either constant ($H(G)=0$) or a bijection of $Y$ ($H(G)=H(Y)$). So $C_{GK}=I(X;Y)$ forces $I(X;Y)=0$ or $I(X;Y)=H(Y)$, i.e.\ $H(Y\mid X)=0$.
\end{proof}

\begin{remark}[Empirical degeneracy]\label{rem:degeneracy}
If all $n$ training inputs are distinct, the empirical distribution has $H_S(Y\mid X)=0$ and $I_S(X;Y)=H_S(Y)$. Any interpolating hard classifier $h$ then has $h(X_i)=Y_i$ for all $i$, so $I_S(h(X);Y)=H_S(Y)=I_S(X;Y)$ and $I_S(h(X);X)=H_S(h(X))=H_S(Y)$: it is an empirical LMC, whatever its test behaviour (1-NN included). The legacy score $I_S(F;Y)/H_S(Y)$ of \citet{nikolaou2020margin} equals $1$ for every exact interpolator. In Table~\ref{tab:pilot} it is $0.99$ rather than $1$ because of duplicate training inputs. With $n=4096$ draws from $2^{19}$ inputs, about $\binom{4096}{2}2^{-19}\approx16$ pairs of training inputs coincide. Where the two labels of a pair conflict, no classifier can fit both, which leaves about $4.6$ unavoidable training errors (training error $0.0011$). The score is computed from hard predictions without binning. With training error $e$, it is close to $1-h(e)=1-h(0.0011)=0.987$, which matches the measured $0.988$--$0.989$.
\end{remark}

\section{Proofs for main Section~\ref{sec:storage-expression}}\label{sec:proofs-storage}

\begin{proof}[Proof of Theorem~\ref{thm:storage}]
$I(W;Y^n\mid X^n)=H(Y^n\mid X^n)-H(Y^n\mid X^n,W)$. Given $X^n$ the labels are independent with laws $\eta(X_i)$, so $H(Y^n\mid X^n)=nH(Y\mid X)$. By subadditivity and since conditioning reduces entropy,
\[
H(Y^n\mid X^n,W)\le\sum_iH(Y_i\mid X_i,W)\le\sum_iH\big(Y_i\mid h_W(X_i)\big)\le\sum_i\phi_K(p_i),
\]
with $p_i=\P(h_W(X_i)\ne Y_i)$, by Fano's inequality. $\phi_K$ is concave, so $\frac1n\sum_i\phi_K(p_i)\le\phi_K(\bar\varepsilon)$. If $\hat\varepsilon=0$ a.s., then $Y^n=h_W(X^n)$ is a function of $(X^n,W)$ and $H(Y^n\mid X^n,W)=0$. Finally, $I(W;S)=I(W;X^n)+I(W;Y^n\mid X^n)\ge I(W;Y^n\mid X^n)$.
\end{proof}

\begin{remark}[Stored task versus stored noise]\label{rem:tasknoise}
Place a prior on the task: a random $\theta$ indexes $P_\theta$, with $S\mid\theta\sim P_\theta^n$, so that $\theta - S - W$. The chain rule gives $I(W;Y^n\mid X^n)=I(W;\theta\mid X^n)+I(W;Y^n\mid X^n,\theta)$, because $I(W;\theta\mid S)=0$. The first term is stored information about the task, and the second is stored noise. For an interpolator the second term equals $nH(Y\mid X,\theta)$ exactly, by Theorem~\ref{thm:storage} applied conditionally on $\theta$.
\end{remark}

\begin{proof}[Proof of Corollary~\ref{cor:xr}]
By Theorem~\ref{thm:storage}, $I(W;S)\ge nH(Y\mid X)$ for every interpolator, so $\sqrt{2\sigma^2I(W;S)/n}\ge\sqrt{2\sigma^2H(Y\mid X)}$ for all $n$.
\end{proof}

\begin{proof}[Proof of Corollary~\ref{cor:cmi}]
For losses in $[0,1]$, Theorem~2 of \citet{steinke2020cmi} gives $|\E[R(h_W)-\hat R_S(h_W)]|\le\sqrt{2\,\mathrm{CMI}/n}$ (CMI in nats). An interpolator has $\hat R_S(h_W)=0$ almost surely, so $\E R(h_W)\le\sqrt{2\,\mathrm{CMI}/n}$, i.e.\ $\mathrm{CMI}\ge\frac n2(\E R(h_W))^2$, and $\E R(h_W)\ge R^*$. In the running example, $R^*=0.15$ gives $\mathrm{CMI}\ge0.0113\,n$ nats.
\end{proof}

\section{Proofs for main Section~\ref{sec:expression} and related results}\label{sec:proofs-expr}

\begin{lemma}[Entropy and mass off the mode; cf.\ \citealp{feder1994relations}]\label{lem:mode}
For any probability vector $q$ on $K$ points with largest atom $q_{\max}$,
\[
\phi_K^{-1}\big(H(q)\big)\ \le\ 1-q_{\max}\ \le\ \tfrac12H(q),
\]
where $\phi_K^{-1}$ is the inverse of $\phi_K(e)=h(e)+e\log_2(K-1)$ on $[0,1-1/K]$ and $H$ is in bits.
\end{lemma}
\begin{proof}
Lower bound: apply Fano's inequality to the estimator that always outputs the mode, whose error probability is $1-q_{\max}\le1-1/K$. Upper bound: let $p=q_{\max}$. If $p\ge\frac12$, the grouping property gives $H(q)=h(p)+(1-p)H(\tilde q)\ge h(p)$, and $h(p)\ge2(1-p)$ on $[\frac12,1]$ by concavity of $h$ (the chord from $(\frac12,1)$ to $(1,0)$). If $p<\frac12$, $H(q)\ge-\log_2p$ (min-entropy), and $g(p)=-\log_2p-2(1-p)$ satisfies $g(\frac12)=0$ and $g'(p)=2-1/(p\ln2)<0$ on $(0,\frac12]$, so $g\ge0$ there.
\end{proof}

\begin{proof}[Proof of Proposition~\ref{prop:supersample}]
$h_W(x)$ is a function of $(S,\xi)$, and $(S,\xi)$ is a function of $(\tilde Z,U,\xi)$, so $H(h_W(x)\mid S,\xi)=H(h_W(x)\mid\tilde Z,U,\xi)=0$ and $V(x)=H(h_W(x))=I(h_W(x);\tilde Z,U,\xi)$. The chain rule gives $I(h_W(x);\tilde Z,U,\xi)=I(h_W(x);\tilde Z)+I(h_W(x);U\mid\tilde Z)+I(h_W(x);\xi\mid\tilde Z,U)$. If the two halves $\tilde Z_U$ and $\tilde Z_{1-U}$ are used to train two copies, then, since the $2n$ examples are i.i.d., the two training samples are independent samples of size $n$; this is why the two-halves estimator of Section~\ref{sec:sel:est} estimates $\E V$ for training size $n/2$ when applied to a dataset of size $n$.
\end{proof}

\begin{proof}[Proof of Theorem~\ref{thm:decomp}]
For any classifier $c$, $R(c)-R^*=\E[\eta_{b^*}(X)-\eta_{c(X)}(X)]$. Averaging over $(S,\xi)$, which is independent of the test point, gives the identity
\begin{equation}\label{eq:exactK}
\E\,R(h_W)-R^*=\E\Big[\textstyle\sum_kq(k\mid X)\big(\eta_{b^*}(X)-\eta_k(X)\big)\Big].
\end{equation}
At fixed $x$, the inner sum lies in $[\gamma(x)(1-q_{b^*}),\,\Delta(x)(1-q_{b^*})]$. If $m=b^*$, then $1-q_{b^*}=1-q_{\max}$, and Lemma~\ref{lem:mode} gives both bounds. If $m\ne b^*$, then $q_{b^*}\le q_m$ and $q_{b^*}+q_m\le1$ give $\frac12\le1-q_{b^*}\le1$. For $K=2$, $\Delta=\gamma=|2\eta_1-1|$; the two-point law has $1-q_{\max}=\pi=h^{-1}(V)$, and $1-q_{b^*}$ equals $\pi$ or $1-\pi$ according to whether $m=b^*$.
\end{proof}

\begin{proof}[Proof of Corollary~\ref{cor:consistency}]
($\Leftarrow$) Use the upper bound of Theorem~\ref{thm:decomp} with $\Delta\le1$. ($\Rightarrow$) Let $Z_n=\phi_K^{-1}(V_n)\ind\{m_n=b^*\}+\frac12\ind\{m_n\ne b^*\}\in[0,1]$. The lower bound gives $\E[\gamma Z_n]\to0$. For $\delta,\tau>0$, $\P(Z_n>\delta)\le\P(\gamma<\tau)+\E[\gamma Z_n]/(\tau\delta)$; letting $n\to\infty$ and then $\tau\to0$ shows $Z_n\to0$ in probability. Hence $\P(m_n\ne b^*)\to0$, and $\phi_K^{-1}(V_n)\to0$ in probability; since $\phi_K$ is continuous with $\phi_K(0)=0$, so does $V_n$. As $V_n\le\log_2K$, bounded convergence gives $\E V_n\to0$.
\end{proof}

\begin{proof}[Proof of Corollary~\ref{cor:toglmc}]
Let $F_n=h_{W_n}(X')$ and $p_n=\P(F_n\ne b^*(X'))$. Since $R(c)-R^*=\E[\eta_{b^*}-\eta_c]\ge\E[\gamma\,\ind\{c\ne b^*\}]$, we have $\E[\gamma\,\ind\{F_n\ne b^*\}]\le\E R(h_{W_n})-R^*\to0$, and the same Markov argument as above gives $p_n\to0$. By Fano's inequality, $H(F_n\mid b^*(X'))\le\phi_K(p_n)$ and $H(b^*(X')\mid F_n)\le\phi_K(p_n)$, hence $|H(F_n)-H(b^*(X'))|\le\phi_K(p_n)\to0$. Corollary~\ref{cor:consistency} gives $\E V_n\to0$, and Proposition~\ref{prop:split} gives $I(F_n;X')=H(F_n)-\E V_n\to H(b^*(X'))=C^*_{01}$.
\end{proof}

\begin{proof}[Proof of Theorem~\ref{thm:sq}]
Since $f_W(x)$ is a function of $(S,\xi)$ and the added noise is independent, $V_\sigma(x)=I(f_W(x)+\sigma Z;f_W(x))=h(f_W(x)+\sigma Z)-h(\sigma Z)$ (differential entropies). (i) The Gaussian maximizes differential entropy at fixed variance, so $h(f_W(x)+\sigma Z)\le\frac12\log(2\pi e(v(x)+\sigma^2))$. (ii) Write $f_W(x)+\sigma Z=\sigma(\sqrt{\mathrm{snr}}\,f_W(x)+Z)$ with $\mathrm{snr}=\sigma^{-2}$. The I-MMSE relation \citep{guo2005immse} gives $\frac{d}{d\,\mathrm{snr}}I=\frac12\mathrm{mmse}(\mathrm{snr})$, with $\mathrm{mmse}$ continuous at $0$ and $\mathrm{mmse}(0)=v(x)$, so $I=\frac{\mathrm{snr}}2v(x)+o(\mathrm{snr})$. (iii) $\E(f_W(X)-Y)^2-\E(a^*(X)-Y)^2=\E(f_W(X)-a^*(X))^2=\E(\bar f-a^*)^2+\E v(X)$. By (i), $0\le2\sigma^2V_\sigma\le\sigma^2\log(1+v/\sigma^2)\le v$, so dominated convergence and (ii) give $\lim_\sigma2\sigma^2\E V_\sigma=\E v$.
\end{proof}

\subsection{Probabilistic outputs and log-loss}\label{sec:logloss}

\begin{proposition}[Log-loss; \citealp{heskes1998bias}]\label{prop:logloss}
Let the learner output class probabilities $q_W(x)\in\Delta_K$ with $\E|\log q_{W,k}(x)|<\infty$. Let $\bar q_k(x)=\exp(\E\log q_{W,k}(x))/Z(x)$ be the normalized geometric mean, where $Z(x)=\sum_k\exp(\E\log q_{W,k}(x))\le1$. Then, exactly,
\[
\E\,\KL\big(\eta(x)\,\|\,q_W(x)\big)=\underbrace{\KL\big(\eta(x)\,\|\,\bar q(x)\big)}_{\text{bias}}+\underbrace{\E\,\KL\big(\bar q(x)\,\|\,q_W(x)\big)}_{V_{\log}(x)\,=\,-\log Z(x)},
\]
and the left-hand side is the expected excess log-loss at $x$.
\end{proposition}
\begin{proof}
$\E\,\KL(\eta\|q_W)=\sum_k\eta_k\log\eta_k-\sum_k\eta_k\E\log q_{W,k}=\sum_k\eta_k\log\eta_k-\sum_k\eta_k(\log\bar q_k+\log Z)=\KL(\eta\|\bar q)-\log Z$; and $\E\,\KL(\bar q\|q_W)=\sum_k\bar q_k(\log\bar q_k-\E\log q_{W,k})=-\log Z$.
\end{proof}
$V_{\log}$ is the log-loss counterpart of $V$: it vanishes iff the learner's probabilities do not depend on the sample, and it is infinite for one-hot outputs that vary, which is how log-loss penalizes confident instability. With logistic regression on a misspecified task, the identity holds to $10^{-15}$ in simulation (\texttt{t\_theory2.py}).

\subsection{A single trained model}\label{sec:single}

\begin{proposition}[A single trained model]\label{prop:single}
For $K=2$ and any $\delta\in(0,1)$, with probability at least $1-\delta$ over $(S,\xi)$,
\[
R(h_W)-R^*\ \le\ \E\big[|2\eta_1-1|\,\ind\{m\ne b^*\}\big]+\frac{1}{2\delta}\,\E\big[|2\eta_1-1|\,V(X)\big].
\]
\end{proposition}
\begin{proof}
$R(h_W)-R^*=\E_X[|2\eta_1-1|\ind\{h_W\ne b^*\}]\le\E_X[|2\eta_1-1|\ind\{m\ne b^*\}]+Z$, where $Z=\E_X[|2\eta_1-1|\ind\{h_W\ne m\}]\ge0$ and $\E Z=\E[|2\eta_1-1|\pi]\le\frac12\E[|2\eta_1-1|V]$ by Lemma~\ref{lem:mode}. Apply Markov's inequality to $Z$.
\end{proof}
So all of a single model's excess over its learner's modal predictor is controlled by expressed information, in probability. Sharper concentration needs stability (Proposition~\ref{prop:stability}).

\subsection{Bias--variance decompositions and disagreement}\label{sec:proofs-bvd}

Table~\ref{tab:dictionary} collects the correspondence between existing variance, disagreement and information notions and the quantities of the main text (Sections~\ref{sec:bvd} and~\ref{sec:related}).

\begin{sidewaystable}\centering\small
\caption{Correspondence between existing information, variance and disagreement notions and the quantities of this paper, at a fixed input $x$ with prediction distribution $q$ and modal prediction $m$. All relations are exact; the last column gives where they are proved. This table accompanies Section~\ref{sec:related}.}\label{tab:dictionary}
\begin{tabular}{p{5.2cm}p{4.6cm}p{8.2cm}p{3.0cm}}\toprule
Existing notion & Definition & Relation to this paper & Where\\\midrule
Main prediction \citep{domingos2000unified} & most frequent prediction & modal prediction $m(x)$ & Section~\ref{sec:expression}\\
Bias \citep{domingos2000unified} & $\ind\{m\ne b^*\}$ & modal bias; second term of Thm.~\ref{thm:decomp} & Thm.~\ref{thm:decomp}\\
Variance \citep{domingos2000unified} & $\pi_D=1-q_m$ & $\phi_K^{-1}(V)\le\pi_D\le V/2$; $\pi_D=h^{-1}(V)$ for $K=2$ & Prop.~\ref{prop:dictionary}\\
Variance \citep{kohavi1996bias} & $\sigma^2_{KW}=\frac12(1-\sum_kq_k^2)$ & $\pi_D\le2\sigma^2_{KW}\le2\pi_D$ & Prop.~\ref{prop:dictionary}\\
Systematic and variance effects \citep{james2003variance} & $\mathrm{SE}$, $\mathrm{VE}$ & $\E[\mathrm{SE}+\mathrm{VE}]=\E R-R^*$; eq.~\eqref{eq:exact} & Prop.~\ref{prop:dictionary}\\
Disagreement of two copies \citep{jiang2022disagreement} & $\P(h_{W_1}(x)\ne h_{W_2}(x))$ & $=2\sigma^2_{KW}$; bounds excess risk from both sides & Prop.~\ref{prop:disagreement-main}\\
GDE \citep{jiang2022disagreement} & $\E R=\E\,\mathrm{Dis}$ & exact if $q=\eta$; holds for 1-NN, fails by $R^*$ for benign learners & Prop.~\ref{prop:disagreement-main}\\
Distributional generalization \citep{nakkiran2020distributional} & predictions mimic noisy labels & $q\to\eta$ for 1-NN; $V\to H(Y\mid X)$ & Props.~\ref{prop:knn}, \ref{prop:disagreement-main}\\
Fine-grained variance \citep{adlam2020understanding} & data / initialization parts & $V=I(h;S)+I(h;\xi\mid S)$ & Section~\ref{sec:bvd}\\
Variance, squared loss & $\Var f_W(x)$ & $\lim_{\sigma\to\infty}2\sigma^2V_\sigma(x)$ & Thm.~\ref{thm:sq}\\
Variance, log-loss \citep{heskes1998bias} & $\E\,\KL(\bar q\|q_W)$ & $V_{\log}=-\log Z$ & Section~\ref{sec:logloss}\\
IB curve end point \citep{tishby2000ib} & $\min I(T;X)$ s.t.\ $I(T;Y)=I(X;Y)$ & $C^*_{\log}=H(\eta(X))$; 0-1 analogue $C^*_{01}=H(b^*(X))$ & Thm.~\ref{thm:minimal}, Section~\ref{sec:rd}\\
Information in the weights \citep{achille2018emergence} & $I(W;S)$ & $\ge nH(Y\mid X)$ for every interpolator & Thm.~\ref{thm:storage}\\
CMI \citep{steinke2020cmi} & $I(W;U\mid\tilde Z)$ & $\ge\frac n2(R^*)^2$ for every interpolator; fresh-input analogue $I(h_W(x);U\mid\tilde Z)\le V(x)$ & Cor.~\ref{cor:cmi}, Prop.~\ref{prop:supersample}\\
Epistemic uncertainty \citep{depeweg2018decomposition} & $I(\text{prediction};\text{parameters})$ under a posterior & $V$ with the sampling distribution in place of the posterior & Section~\ref{sec:related}\\\bottomrule
\end{tabular}
\end{sidewaystable}

\begin{proof}[Proof of Proposition~\ref{prop:dictionary}]
Fix $x$ and write $q_k=q(k\mid x)$, $q_m=\max_kq_k$. The first and last inequalities are Lemma~\ref{lem:mode}, since $\pi_D=1-q_m$. Next, $\sum_kq_k^2\le q_m\sum_kq_k=q_m$, so $\mathrm{Dis}=1-\sum_kq_k^2\ge1-q_m=\pi_D$. Also $\sum_kq_k^2\ge q_m^2$, so $\mathrm{Dis}\le1-q_m^2=(1-q_m)(1+q_m)\le2(1-q_m)=2\pi_D$. Two independent copies disagree with probability $1-\sum_k\P(h_{W_1}(x)=k)\P(h_{W_2}(x)=k)=1-\sum_kq_k^2$. For $K=2$, the two-point law has $\pi_D=\min(q_1,q_2)=h^{-1}(V)$ and $\mathrm{Dis}=2q_1q_2=2\pi_D(1-\pi_D)$; both maps are increasing on $[0,\frac12]$.
For the decomposition, $\P(c(x)\ne Y)=1-\eta_{c(x)}(x)$ for any fixed classifier $c$, so $\mathrm{SE}(x)=\eta_{b^*}(x)-\eta_m(x)$ and $\mathrm{VE}(x)=\sum_kq_k(\eta_m(x)-\eta_k(x))$. Their sum is $\sum_kq_k(\eta_{b^*}(x)-\eta_k(x))$, whose expectation is $\E R(h_W)-R^*$ by \eqref{eq:exactK}. For $K=2$: if $B=0$, $\mathrm{SE}=0$ and $\mathrm{VE}=|2\eta_1-1|\pi_D$; if $B=1$, $\mathrm{SE}=|2\eta_1-1|$ and $\mathrm{VE}=q_{b^*}(\eta_m-\eta_{b^*})=-|2\eta_1-1|\pi_D$, since $q_{b^*}=1-q_m=\pi_D$.
\end{proof}

\begin{proof}[Proof of Proposition~\ref{prop:disagreement-main}]
(a) At $x$, $\E\,\ind\{h_W(x)\ne Y\}=\sum_kq_k(1-\eta_k)$, which equals $1-\sum_k\eta_k^2=\mathrm{Dis}(x)$ when $q=\eta$.
(b) By \eqref{eq:exact} with $m=b^*$, $\E R-R^*=\E[|2\eta_1-1|\pi_D]\le\E\pi_D$. The remaining inequalities are Proposition~\ref{prop:dictionary} averaged over $X$. If $|2\eta_1-1|\ge\gamma_0$, then $\E R-R^*\ge\gamma_0\E\pi_D\ge\frac{\gamma_0}{2}\mathrm{Dis}$.
(c) By Proposition~\ref{prop:knn} with $k=1$, $\pi_D(x)\to\mu(x)=\min(\eta_1(x),1-\eta_1(x))$ for almost every $x$, so $\mathrm{Dis}(x)=2\pi_D(1-\pi_D)\to2\mu(1-\mu)=2\eta_1(1-\eta_1)$; bounded convergence gives the limit of $\mathrm{Dis}$, which is the Cover--Hart limit of $\E R$. For a consistent sequence, Corollary~\ref{cor:consistency} gives $\E V_n\to0$, hence $\mathrm{Dis}\le\E V_n\to0$, while $\E R\to R^*$.
\end{proof}

\section{Proofs for main Section~\ref{sec:benign}}\label{sec:proofs-benign}

Throughout this section $K=2$, $p_n(x)=\P(h_W(x)=1)$ over $(S,\xi)$, and $\pi_n(x)=\min(p_n(x),1-p_n(x))$, so that $V_n(x)=h(\pi_n(x))$.

\begin{proof}[Proof of Proposition~\ref{prop:memsmooth}]
(a) For two independent inputs, $\P(X_i=X_j)=\sum_z\P(Z=z)^2\,2^{-d}\le2^{-d}$, so a union bound over pairs gives a collision probability of at most $\binom n22^{-d}\le n^22^{-d-1}\to0$. Without collisions the learner interpolates.
(b) The training error satisfies $\hat\varepsilon\le\ind\{\text{collision}\}$, so $\bar\varepsilon\le n^22^{-d-1}\to0$, and Theorem~\ref{thm:storage} gives $I(W;Y^n\mid X^n)\ge n[H(Y\mid X)-\phi_2(\bar\varepsilon)]=n(H(Y\mid X)-o(1))$.
(c) For a fresh $x'=(z,u)$, $\P(x'\in\{X_1,\dots,X_n\})\le n2^{-d}\to0$. Off the sample, the prediction is the majority of $N_z\sim\mathrm{Bin}(n,\P(Z=z))$ labels with mean $\eta_1(z)\ne\frac12$, which is wrong with probability at most $\E\exp(-2N_z(\eta_1(z)-\frac12)^2)\to0$ by Hoeffding's inequality. So $\sup_{x'}\P(h_W(x')\ne b^*(x'))\to0$ (finitely many $z$). Hence $m_n=b^*$ eventually, $\pi_n\to0$ uniformly, $V_n\to0$, and $\E R\to R^*$ by Theorem~\ref{thm:decomp}.
\end{proof}

\begin{proof}[Proof of Proposition~\ref{prop:knn}]
Break distance and vote ties uniformly at random. For $x$ in the support of $P_X$, the distance from $x$ to its $k$-th nearest neighbour tends to $0$ a.s.\ as $n\to\infty$ with $k$ fixed \citep[Lemma~5.1]{devroye1996probabilistic}. Given $X^n$, the $k$ neighbour labels are independent Bernoulli$(\eta_1(X_{(i)}(x)))$, and by continuity of $\eta_1$ at $x$ their conditional law converges a.s.\ to that of $k$ i.i.d.\ Bernoulli$(\eta_1(x))$ labels. Bounded convergence gives $p_n(x)\to\beta_k(\eta_1(x))$, and continuity of $t\mapsto h(\min(t,1-t))$ gives the limit of $V_n(x)$; bounded convergence again gives the limit of $\E V_n$. The bound is Hoeffding's inequality, $\min(\beta_k,1-\beta_k)\le\exp(-2k(\eta_1-\frac12)^2)$, together with monotonicity of $h$ on $[0,\frac12]$. For odd $k$, $\beta_k(t)>\frac12$ iff $t>\frac12$, so the modal prediction converges to $b^*$ wherever $\eta_1\ne\frac12$. The excess-risk limit follows from the exact identity \eqref{eq:exact} by bounded convergence: $\E R-R^*\to\E[|2\eta_1-1|\min\{\beta_k(\eta_1),1-\beta_k(\eta_1)\}]$. For $k=1$, $\beta_1(t)=t$ and $h(\min(\eta_1,1-\eta_1))=h(\eta_1)$, whose expectation is $H(Y\mid X)$. With $\mu=\min(\eta_1,1-\eta_1)$, $|2\eta_1-1|\mu=\mu-2\mu^2=2\eta_1(1-\eta_1)-\mu$, and $\E\mu=R^*$.
\end{proof}

\begin{proof}[Proof of Proposition~\ref{prop:knn-rate}]
Since $h(\pi)\le2\sqrt\pi$ on $[0,\frac12]$, it suffices to bound $\E\sqrt{\pi_n(X)}$, and $\pi_n(x)\le\P(\text{vote}\ne b^*(x))$. Fix $x$, let $\gamma=|\eta_1(x)-\frac12|$, and let $r_k(x)$ be the $k$-th nearest-neighbour distance. On $\{Lr_k\le\gamma/2\}$, the neighbours' labels are independent with means on the correct side of $\frac12$ by at least $\gamma/2$, so Hoeffding gives $\P(\text{vote wrong}\mid X^n)\le e^{-k\gamma^2/2}$. Hence $\pi_n(x)\le e^{-k\gamma^2/2}+\P(r_k(x)>\gamma/(2L))$, and $\sqrt{\pi_n(x)}\le e^{-k\gamma^2/4}+\P(r_k(x)>\gamma/(2L))^{1/2}$. The margin condition gives $\E\,e^{-k\gamma^2/4}\le\sum_{j\ge0}\P(\gamma\le(j+1)/\sqrt k)e^{-j^2/4}\le c_1k^{-\alpha/2}$. Under the density assumption, a Chernoff bound gives $\P(r_k(x)>r)\le e^{-cnr^d}$ whenever $nr^d\ge c'k$ \citep{devroye1996probabilistic}. Split on $\gamma\le t_n$ with $t_n=2L(c'k/n)^{1/d}$: the event $\{\gamma\le t_n\}$ has probability at most $C_0t_n^\alpha=c_2(k/n)^{\alpha/d}$, and on its complement $\P(r_k>\gamma/(2L))\le e^{-cc'k}$. Altogether $\E V_n\le c_1k^{-\alpha/2}+c_2(k/n)^{\alpha/d}+c_3e^{-c_4k}$, and $k\asymp n^{2/(d+2)}$ balances the first two terms at $n^{-\alpha/(d+2)}$.
\end{proof}

\begin{proposition}[Averaging, full statement]\label{prop:ensemble-full}
Let $g_B$ be the majority vote of $B$ independent copies of a binary learner ($B$ odd; independent samples and seeds), and write $\pi=\pi_n$, $m=m_n$. Then for every $x$:
(a) the modal prediction of $g_B$ is $m(x)$;
(b) $\pi_{g_B}(x)=\P(\mathrm{Bin}(B,\pi(x))>B/2)\le\exp(-2B(\frac12-\pi(x))^2)$;
(c) as $B\to\infty$,
\[
\E R(g_B)-\E R(h_W)\ \to\ \E\Big[|2\eta_1-1|\,\pi\,\big(\ind\{m\ne b^*\}-\ind\{m=b^*\}\big)\,\ind\{\pi<\tfrac12\}\Big].
\]
\end{proposition}
\begin{proof}
(a)--(b) Given $x$, the $B$ votes are i.i.d.\ Bernoulli$(p(x))$. For odd $B$, $\P(\mathrm{Bin}(B,p)>B/2)>\frac12$ iff $p>\frac12$, and the probability that the vote disagrees with $m(x)$ is $\P(\mathrm{Bin}(B,\pi)>B/2)$, which Hoeffding's inequality bounds. (c) By \eqref{eq:exact}, $\E R(g_B)-R^*=\E[|2\eta_1-1|(\pi_{g_B}\ind\{m=b^*\}+(1-\pi_{g_B})\ind\{m\ne b^*\})]$. As $B\to\infty$, $\pi_{g_B}\to0$ where $\pi<\frac12$ and stays $\frac12$ where $\pi=\frac12$. Subtract the same identity for $h_W$ and use bounded convergence.
\end{proof}
Part (c) says that averaging reduces risk exactly when the expression it removes where the mode is right outweighs the expression it removes where the mode is wrong (where averaging locks the error in). Practical bagging uses bootstrap resamples of one sample, which are dependent, so its gain is smaller than (b) suggests; random forests partly restore independence by feature subsampling \citep{breiman1996bagging}.

\begin{proposition}[Stable learners express little]\label{prop:stability}
Let the learner be deterministic, let $S^{(i)}$ be $S$ with its $i$-th example replaced by an independent copy, and let $\beta_n(x)=\max_i\P(h_S(x)\ne h_{S^{(i)}}(x))$. Then $\pi_n(x)\le n\beta_n(x)$ and $V_n(x)\le h(\min\{\frac12,n\beta_n(x)\})$.
\end{proposition}
\begin{proof}
Apply the Efron--Stein inequality to $f(S)=\ind\{h_S(x)=1\}$: $p(1-p)=\Var f(S)\le\frac12\sum_i\E(f(S)-f(S^{(i)}))^2\le\frac n2\beta_n(x)$. Since $\pi\le\frac12$, $\pi(1-\pi)\ge\pi/2$, so $\pi\le n\beta_n(x)$. Then use monotonicity of $h$ on $[0,\frac12]$.
\end{proof}
Uniform stability at rate $o(1/n)$ \citep{bousquet2002stability} thus forces $V_n\to0$. Interpolators of noisy labels cannot be that stable, since replacing a training point near $x$ flips the prediction there, so this result is mainly informative for regularized learners.

\begin{proof}[Proof of Proposition~\ref{prop:ridgeless}]
The prediction $x'^\top X^+(X\beta+\varepsilon)$ is Gaussian with variance $\sigma_\varepsilon^2\|(X^+)^\top x'\|^2$ given the design and $x'$. Adding independent $N(0,\sigma^2)$ noise gives a Gaussian channel with Gaussian input, whose mutual information is the stated formula. For $p\ge n$ (full row rank), $X^+=X^\top(XX^\top)^{-1}$, and $\E_{x'}\|(XX^\top)^{-1}Xx'\|^2=\operatorname{tr}\big((XX^\top)^{-1}X\Sigma X^\top(XX^\top)^{-1}\big)=\operatorname{tr}(C)$ (for centred $x'$; otherwise $\Sigma$ is the second-moment matrix). The limit follows from $\log(1+a)/a\to1$ and dominated convergence, as in Theorem~\ref{thm:sq}.
\end{proof}
A Monte Carlo check with $n=50$, $p=200$ matches the predictive variance $\sigma_\varepsilon^2\|(X^+)^\top x'\|^2$ to within $2\%$ (\texttt{t\_theory2.py}).

\section{Theory of the consensus criterion and of noise correction}\label{sec:consensus}

In Section~\ref{sec:sel:beyond} we introduced the label-free consensus criterion \eqref{eq:lf} and a switch to noise-corrected validation. This section gives the theory behind both: what disagreement with a consensus measures, when the consensus is close to the Bayes classifier, and the two ways it can fail.

Let $m_1,\dots,m_J$ be the candidates' modal predictors and $c$ their plurality vote (fixed tie-break). Write $\beta_j=\P(m_j(X)\ne b^*(X))$ for candidate $j$'s modal-bias mass, $\bar\beta$ for its average over candidates, $D_j=\P(m_j(X)\ne c(X))$ for its disagreement with the consensus, and $\delta=\P(c(X)\ne b^*(X))$ for the consensus error. Let $W(x)=\frac1J\sum_j\ind\{m_j(x)\ne b^*(x)\}$ be the fraction of candidates that are wrong at $x$.

\begin{proposition}[Consensus error is shared-bias mass]\label{prop:consensus-error}
$\delta\le\P(W(X)\ge\frac12)\le2\bar\beta$. The same holds when the voters are clusters of candidates, each voting with its own plurality, with $W$ and $\bar\beta$ computed over clusters.
\end{proposition}
\begin{proof}
If more than half of the candidates predict $b^*(x)$, then $b^*(x)$ is the plurality, so $\{c\ne b^*\}\subseteq\{W\ge\frac12\}$. Markov's inequality gives $\P(W\ge\frac12)\le2\E W=2\bar\beta$. The cluster version is the same argument with clusters as voters.
\end{proof}

In words, the consensus is wrong only where at least half of the voters share an error. Many candidates with \emph{different} biases do no harm; a majority with the \emph{same} bias does.

\begin{proposition}[What disagreement measures]\label{prop:disagreement}
Let $\mathcal S=\{c\ne b^*\}$.
(a) For $K=2$, $D_j=\beta_j+\delta-2s_j$, where $s_j=\P(\mathcal S\cap\{m_j\ne b^*\})\le\min(\beta_j,\delta)$ is the part of candidate $j$'s bias that it shares with the consensus.
(b) For any $K$, $|D_j-\beta_j|\le\delta$.
Consequently $|(D_i-D_j)-(\beta_i-\beta_j)|\le2\delta$ for all $i,j$, and the candidate with the smallest disagreement has modal-bias mass within $2\delta$ of the smallest.
\end{proposition}
\begin{proof}
(a) For binary labels, $m_j(x)\ne c(x)$ iff exactly one of $m_j(x)\ne b^*(x)$ and $c(x)\ne b^*(x)$ holds, so $D_j=\beta_j+\delta-2s_j$. (b) $\P(u\ne w)$ is a pseudometric on classifiers, and the triangle inequality gives $|\P(m_j\ne c)-\P(m_j\ne b^*)|\le\P(c\ne b^*)$. The consequence follows by subtracting the bounds for $i$ and $j$.
\end{proof}

Part (a) states the failure mode exactly: disagreement counts a candidate's private errors in full, but \emph{subtracts} the errors it shares with the consensus, so conforming to a biased majority is rewarded. Two mechanisms produce such a majority.

\emph{Redundant candidates.} When many candidates come from one family with a common misspecification (for example several regularization levels of a linear model on a curved boundary), they form a majority with a common bias. Giving one vote per cluster of near-identical candidates removes the redundancy: by Proposition~\ref{prop:consensus-error}, only clusters then count. Our \emph{adaptive clustering} uses average-linkage clusters at disagreement $0.05$, one vote per cluster, and only when some cluster holds at least a third of the candidates.

\emph{Asymmetric label noise.} The second mechanism affects every candidate at once.

\begin{proposition}[Noise-induced shared bias]\label{prop:noise-shift}
Let $K=2$ and let the training labels pass through class-conditional noise with $\rho_0=\P(\tilde Y=1\mid Y=0)$ and $\rho_1=\P(\tilde Y=0\mid Y=1)$, where $\rho_0+\rho_1<1$. Then $\tilde\eta_1=\rho_0+(1-\rho_0-\rho_1)\eta_1$, and the Bayes classifier of the noisy problem is $\tilde b^*(x)=\ind\{\eta_1(x)>\theta\}$ with $\theta=(\frac12-\rho_0)/(1-\rho_0-\rho_1)$. If every candidate is consistent for the noisy problem, then $\delta\to\delta_T:=\P(\tilde b^*(X)\ne b^*(X))$, the mass of $\eta_1$ between $\theta$ and $\frac12$. In particular $\delta_T=0$ when $\rho_0=\rho_1$: symmetric noise creates no shared bias.
\end{proposition}
\begin{proof}
$\P(\tilde Y=1\mid X)=\rho_0(1-\eta_1)+(1-\rho_1)\eta_1$, which exceeds $\frac12$ iff $\eta_1>\theta$. Consistency for the noisy problem drives each candidate's prediction, and hence its mode, to $\tilde b^*(x)$ wherever $\eta_1(x)\ne\theta$, so the plurality is $\tilde b^*$ and $\delta\to\P(\tilde b^*\ne b^*)$. If $\rho_0=\rho_1<\frac12$, then $\theta=\frac12$.
\end{proof}

The shift cannot be seen in the candidates' predictions, because every candidate shares it. Holdout on noisy labels does not see it either, since it estimates the noisy risk, which $\tilde b^*$ minimizes. Removing it needs knowledge of the transition matrix $T$, through backward correction \citep{natarajan2013noisy,patrini2017making}: for $T_{jk}=\P(\tilde Y=k\mid Y=j)$ invertible, $\E[1-(T^{-1})_{\tilde Y\hat y}\mid Y=y]=1-\sum_kT_{yk}(T^{-1})_{k\hat y}=\ind\{\hat y\ne y\}$, an unbiased estimate of the clean 0-1 loss. The next remark shows when correction can change a selection at all.

\begin{remark}[Noise correction cannot reorder models under symmetric noise]\label{rem:symcorr}
If $T=(1-\rho)I+\frac{\rho}{K-1}(J-I)$ with $\rho<\frac{K-1}{K}$ ($J$ the all-ones matrix), then $T^{-1}=aI+bJ$ with $a=\big(1-\rho K/(K-1)\big)^{-1}>0$. The corrected loss $1-(T^{-1})_{\tilde y\hat y}$ therefore equals $1-b-a+a\,\ind\{\hat y\ne\tilde y\}$, an increasing affine function of the noisy loss. Averaged over any validation set, it ranks models exactly as uncorrected holdout does. Correction can change a selection only when the noise is asymmetric.
\end{remark}

\paragraph{Estimating $T$ and the switch rule.} We estimate $T$ without extra labels: $\hat T_{jk}$ is the fraction of training points whose consensus out-of-fold prediction is $j$ and whose noisy label is $k$. For numerical stability we shrink, $\hat T\leftarrow0.95\hat T+0.05I$, and fall back to no correction when the condition number exceeds $50$. The \emph{asymmetry} of $\hat T$ is its largest entrywise deviation from the symmetric noise matrix with the same mean flip rate $\hat\rho=1-\operatorname{tr}(\hat T)/K$. The switch rule uses noise-corrected validation when the asymmetry exceeds $0.1$, and the label-free criterion otherwise. On the 90 Protocol-A cases, the asymmetry is $0.023$ on average (at most $0.087$) under symmetric noise and $0.144$ on average (at least $0.067$) under class-dependent noise, so the threshold separates the two noise types; thresholds between $0.05$ and $0.15$ give the same regret to within $0.001$.

\section{Erratum for the earlier work on lossless maximal compression}\label{sec:erratum}

This section corrects two earlier non-archival papers: the preprint \citet{nikolaou2020margin}, ``Margin Maximization as Lossless Maximal Compression'' (arXiv:2001.10318), and the short paper \citet{nikolaou2021lossless}, ``Lossless Compression \& Generalization in Overparameterized Models: The Case of Boosting'' (ICLR 2021). We call them ``the preprint'' and ``the short paper''. All numbered items below refer to the preprint's arXiv version, unless stated otherwise. For each problem we state what was claimed, what is wrong, and how the main text corrects it. We end with what survives.

\paragraph{What was claimed.} The preprint defines, on a training set $S$, noiselessness ($H_S(Y\mid X)=0$; Def.~1), losslessness ($I_S(F;Y)=I_S(X;Y)$; Def.~2), maximal compression ($I_S(F;X)=I_S(F;Y)$; Def.~3), and a \emph{lossless maximal compressor} (LMC) as a model that is both (Def.~4). Prop.~1 states that $f$ is an LMC iff $I_S(F;X)=I_S(F;Y)=I_S(X;Y)$, and Theorem~1 states that, for a noiseless finitely supported $S$, $f$ is an LMC iff some invertible transform $g\circ f$ is a margin maximizer (Theorem~2 and Prop.~2 in its supplement extend these to a general finitely supported distribution). Table~1 and Figure~1 enumerate the relations between the information quantities in eight scenarios: noiseless or noisy data, lossy or lossless, under- or maximally compressed. Section~4 reports that boosting ``traces a trajectory on the information plane that leads to the LMC point'' and never leaves it. Section~4.2 is titled ``Early stopping does not improve generalization in gradient boosting''. The short paper restates the definitions and Propositions (I)--(III), argues that ``adequate overparameterization guarantees losslessness'' and that inductive biases then drive training towards maximal compression, and shows the same trajectories on the \texttt{musk} and \texttt{credit} datasets. Both papers estimate all quantities on the training data, after discretizing features and outputs into $b=100$ equal-width bins and encoding the joint input as $X=\sum_iX_ib^{i-1}$, with 100 random 50/50 train/test splits.

\paragraph{Error 1: the noisy-case LMC does not exist.} For a deterministic output, $I(F;X)=I(F;Y)$ forces $H(F\mid Y)=0$, so $F$ would have to be a function of the label. By Proposition~\ref{prop:gk}, a deterministic LMC exists iff the G\'acs--K\"orner common information equals $I(X;Y)$; for binary labels, this means $H(Y\mid X)=0$ or $I(X;Y)=0$. Prop.~1 (and its supplementary version, Prop.~2) is correct as a characterization, but for noisy data with informative features the set it characterizes is empty. In particular:
\begin{itemize}[leftmargin=*,itemsep=0pt]
\item the ``noisy, lossless, maximally compressed'' row of Table~1, $I(F;Y)=I(X;Y)=I(F;X)<H(Y)$, and the corresponding bottom-right panel of Figure~1 (noisy column, lossless maximally compressed row) describe an impossible configuration;
\item so does the statement in Table~2 that, on noisy data, ``some training error minimizers are LMCs'';
\item and so does the suggestion in Section~3.3 that models capturing the label distribution are ``ideally'' LMCs.
\end{itemize}
The preprint's own observation that ``in the noisy case, a margin maximizer cannot even be a lossless model'' is correct. \emph{Correction:} Section~\ref{sec:minimal} replaces LMC by the minimal $\ell$-sufficient representation (G-LMC), which always exists. It is the Bayes act $a^*(X)$: $\eta(X)$ under log-loss and $b^*(X)$ under 0-1 loss. Lemma~\ref{lem:noisecomp} quantifies the gap: a log-loss sufficient output must retain at least $H(\eta(X)\mid Y)>0$ bits beyond $I(F;Y)$.

\paragraph{Error 2: empirical information cannot tell interpolators apart.} With distinct training inputs, the empirical distribution is noiseless, so every interpolating hard classifier is an empirical LMC (Remark~\ref{rem:degeneracy}). This includes the 1-NN rule, whose asymptotic risk can approach twice the Bayes risk. The binning used in both papers makes this worse: with $b=100$ bins per feature and several features, almost every training input occupies its own cell. The preprint notes that discretization did not make any dataset noisy, which is precisely this degeneracy, and it reports that ``all results shown are on noiseless datasets'', including a synthetic dataset with $1\%$ label flips. Table~\ref{tab:pilot} confirms the consequence. The legacy score $I_S(F;Y)/H_S(Y)$ is $0.99$ for five interpolators whose excess risks range over two orders of magnitude. In Section~\ref{sec:exp:real} it is $0.91$--$1.00$ for every interpolator on real data with injected noise, against test errors of $0.04$--$0.50$. \emph{Correction:} every quantity in the main text is defined at the population level, or through the learner's behaviour over independent training samples.

\paragraph{Error 3: estimation artifacts.} Binned estimates of mutual information for deterministic maps depend strongly on the bin width, and they can show spurious ``compression'' \citep{saxe2018ib,kolchinsky2019caveats,goldfeld2019estimating}. On the training set, the normalized coordinate $I_S(F;X)/H_S(X)$ of a hard or binned output is $H_S(F)/\log_2n$. It falls whenever the binned training scores concentrate into fewer cells, for example as boosting pushes scores towards the extremes. That is a property of the score distribution on the training set, not evidence of better generalization (Error~5). \emph{Correction:} the main text's ideal-model quantities use discrete inputs and outputs, where they are well defined. $V$ is an entropy of hard predictions over training samples at a fixed input, so it needs no binning.

\paragraph{Error 4: margin guarantees are not inherited by LMC models in general.} The preprint argues that ``the existence of margin-based bounds on the generalization error implies that margin maximization is beneficial'', and therefore so is lossless maximal compression. This inference holds only within Theorem~1: noiseless data, and only for the transformed score $g\circ f$ with some invertible $g$. On the training set, Theorem~1 applies to every interpolating score, because the empirical distribution is noiseless (Error~2). Margin bounds depend on the margin distribution of the actual score, which an arbitrary invertible $g$ can change at will. So an LMC on the training set carries no margin guarantee; the 1-NN rule is an example. Under noise, LMC models do not exist (Error~1). \emph{Correction:} the main text makes no margin claims. The link between learners and risk goes through Theorem~\ref{thm:decomp}.

\paragraph{Error 5: compression and early stopping under label noise.} Both papers read the late-stage decrease of $I_S(F;X)$ as compression that improves generalization, and concluded that early stopping is unnecessary in boosting. Their experiments used data that were noiseless, at least empirically (Error~2). We replicated the experiment on the running example with $20\%$ label noise (Section~\ref{sec:exp:boost}; $n=1024$, depth-6 trees, exponential loss, 200 rounds).
\begin{itemize}[leftmargin=*,itemsep=0pt]
\item \emph{With the original setting} (learning rate 1, no shrinkage), the old coordinate falls from $0.52$ at round 5 to $0.10$ by round 25 and stays there. Over the same rounds, excess risk stays at $0.058$--$0.065$, and $V$ at $0.39$--$0.43$ bits.
\item \emph{With shrinkage $0.1$} (Figure~\ref{fig:boost}), the coordinate falls from $0.58$ to $0.12$, while excess risk rises from $0.011$ at round 15 to $0.050$ at round 200, and $V$ from $0.09$ to $0.35$ bits.
\end{itemize}
So ``compression'' does not improve generalization under noise, and early stopping with shrinkage gives a five times lower excess risk than the original unstopped setting. On four real datasets with $20\%$ injected noise (depth-3 trees, shrinkage $0.1$) the same reversal occurs (Figure~\ref{fig:boostreal}). Without injected noise, the old reading is reproduced on the nearly noiseless \texttt{wdbc} and \texttt{biopsy}, and reversed on \texttt{OJ} and \texttt{Caravan}, whose labels are intrinsically noisy. \emph{Correction:} the claim holds only on (nearly) noiseless data. Under label noise, early stopping helps, and expressed information tracks test error along the trajectory (Spearman $0.82$--$1.00$ on real data).

\paragraph{What survives.}
\begin{enumerate}[label=(\roman*),leftmargin=*,itemsep=0pt]
\item The noiseless-case identification of LMC with the Bayes decision, up to relabelling, is the special case $H(Y\mid X)=0$ of Theorem~\ref{thm:minimal}. The noiseless part of Theorem~1 is unaffected as a statement about distributions.
\item Lemma~3 of the preprint, that $I(F;Y)=I(X;Y)$ iff $x\mapsto\P(Y=1\mid X=x)$ is constant on the level sets of $f$, is the binary log-loss case of Proposition~\ref{prop:suff}. It is now stated for general losses, $K$ classes and stochastic representations.
\item The observation that margin maximizers cannot be lossless under noise is correct. In the terms of the main text, a margin maximizer can be 0-1 sufficient but not log-loss sufficient, because $C^*_{01}=H(b^*(X))<H(\eta(X))=C^*_{\log}$ in general.
\item The motivation, that a model's output complexity should be compared with a minimal sufficient level, survives as the excess complexity of Section~\ref{sec:minimal}. It is complemented by expressed information, which explains generalization where the output-level picture cannot.
\item On noiseless data, the boosting trajectories reported earlier are reproduced.
\end{enumerate}

\section{Experimental details}\label{sec:expdetails}

All experiments use Python with \texttt{numpy}, \texttt{scipy} and \texttt{scikit-learn} \citep{pedregosa2011scikit}, and run on CPUs. Section~\ref{sec:code} lists the script for each experiment.

\subsection{Synthetic tasks}
\paragraph{Running example (Section~\ref{sec:exp:synth}).} As in Example~\ref{ex:running}. Excess risks are computed exactly from $\eta$ on a fixed Monte Carlo set of test inputs ($20{,}000$ inputs). The learners are: memorize-and-smooth; 1-NN and $k$-NN with $k$ the odd integer nearest $\sqrt n$ (Hamming distance); a full-depth tree; a random forest (100 trees); gradient boosting with exponential loss, learning rate 1, depth 6 and 100 rounds (the setting of the earlier papers); logistic regression with $C=10^4$; and a 512-unit MLP trained without regularization until it interpolates. $R=30$ independent training samples per configuration. Soft outputs, where needed for log-loss quantities, are quantized to multiples of $1/20$. Table~\ref{tab:pilot} is a separate pilot run of the same task; its numbers differ from those of Section~\ref{sec:exp:synth} in the third decimal.

\paragraph{Double descent (Section~\ref{sec:exp:dd}).} Random ReLU features $\phi(x)=\max(Gx+b,0)/\sqrt p$ with Gaussian $G$, $b$, on the one-hot encoding of the running example; minimum-norm least squares on $\pm1$ labels (or one-hot targets for \texttt{digits}); prediction by sign (argmax). Widths $p\in\{16,\dots,4096\}$, $n=512$, 40 independent training sets per width; ridge $\lambda=10^{-2}$ for the regularized branch. The feature draw is part of the learner's internal randomness $\xi$.

\paragraph{Boosting (Section~\ref{sec:exp:boost}).} Gradient boosting with exponential loss. Running example: $n=1024$, $R=30$, depth-6 trees, 200 rounds, learning rate $1$ (the setting of the earlier papers) and $0.1$. Real data: 300 rounds, (learning rate $1$, depth 6) and (learning rate $0.1$, depth 3); the figures show the latter. The old coordinate is the normalized empirical $I_S(F;X)/H_S(X)=H_S(F)/\log_2n$ of the earlier papers, computed on the training data with the real-valued score binned into 100 bins, as in those papers.

\paragraph{Stress tests (Section~\ref{sec:stress}).} $X\sim U[0,1]^6$, $\eta_1(x)=\frac12+0.4\tanh\big(8(x_2-\frac12-0.3\sin3\pi x_1)\big)$. Eighteen candidates: nine in a shared-bias family (logistic regression with five values of $C$, two linear SVMs, LDA, a ridge classifier), and nine diverse learners ($k$-NN with $k=1,15$, random forest, extra trees, gradient boosting, RBF SVM, MLP, two trees). Each candidate is trained 10 times on independent samples of size 400. Validation: 300 points, or random subsets of 40. 12 replicates per condition; 20 random pools of 12 candidates per replicate.

\subsection{Real data}
\paragraph{Datasets.} \texttt{wdbc} and \texttt{digits} from scikit-learn; \texttt{OJ} and \texttt{Caravan} (class-balanced by subsampling the majority class) from the ISLP package \citep{james2023islp}; \texttt{biopsy} (original Wisconsin breast cancer). For Protocol B we add ten benchmarks from PMLB \citep{olson2017pmlb} not used before: \texttt{twonorm}, \texttt{ring}, \texttt{magic}, \texttt{phoneme}, \texttt{spambase}, \texttt{churn}, \texttt{waveform-40}, \texttt{satimage}, \texttt{optdigits}, \texttt{kr-vs-kp}, each subsampled to at most 2{,}000 points. Features are standardized, and categorical features are one-hot encoded.

\paragraph{The 19 learner configurations.} $k$-NN with $k\in\{1,3,9,27\}$; decision trees of depth $2,4,8$ and unlimited; random forests (200 trees) with minimum leaf size 1 and 5; gradient boosting with (100 rounds, depth 3, learning rate $0.1$) and (300 rounds, depth 6, learning rate 1); logistic regression with $C\in\{0.01,1,10^4\}$; RBF SVMs with $C\in\{1,10^4\}$; 512-unit MLPs with $\alpha\in\{10^{-5},1\}$. The six \emph{overparameterized} configurations of Section~\ref{sec:exp:select} are 1-NN, the unlimited tree, the forest with leaf size 1, the (300, 6, 1) boosting, the SVM with $C=10^4$ and the MLP with $\alpha=10^{-5}$. They were fixed before running the selection analysis, as the configurations designed to interpolate.

\paragraph{Noise models.} \emph{Symmetric}: each label is replaced with probability $\rho$ by a uniformly random different class. \emph{Class-dependent}: for binary tasks, $0\to1$ with probability $0.4$ and $1\to0$ with probability $0.05$; for multiclass tasks, $k\to k+1\pmod K$ with probability $0.35$. \emph{Instance-dependent} (Protocol B only): a label is flipped with probability $\min(0.5,0.7\,u(x))$ to the most confusable other class, where $u(x)=1-(p_{(1)}(x)-p_{(2)}(x))$ is one minus the class-probability margin of a reference random forest fit to clean data; the mean flip rate is $0.16$.

\paragraph{Protocol A (independent samples).} Clean test set 30\%; validation set 20\% (113--359 points), with labels corrupted by the same noise process; training pool 50\%. Each configuration is trained $R=12$ times on random halves of the pool, each with a fresh noise draw and seed; all configurations share the same halves and draws. Six random splits per dataset and noise setting (symmetric $0.2$ and $0.3$; class-dependent): 60 symmetric and 30 class-dependent cases. Small validation budgets: 40 random validation points, 100 draws per case. $V$ is estimated on the (unlabelled) test inputs.

\paragraph{Protocol B (one dataset).} Five seeds per dataset and noise model (225 cases); the seed sets the subsample (datasets above 2{,}000 points), the train/test split, the noise draw, the holdout split and the folds (\texttt{r6\_cost.py} for seed 0, \texttt{r6\_seeds.py} for seeds 1--4). Clean test set 30\%; the remaining 70\% receive one noise draw. Each configuration is fit (i) on a random 80\% of the training set (holdout on the other 20\%), (ii) in 2-fold and 5-fold cross-validation, and (iii) on the full training set. Fold models also predict on the unlabelled test inputs, from which $V_2$, $V_5$ and the consensus criteria are computed. The noise-corrected CV estimates $\hat T$ from the plurality of the out-of-fold predictions of all 19 configurations. CPU time is recorded for every fit and every prediction; costs are reported relative to the holdout fit of the same candidate set. Regret uses the full refit of the chosen configuration. Paired differences between criteria are summarized by a cluster bootstrap (4{,}000 resamples) over the 45 dataset--noise combinations (\texttt{r6\_multiseed.py}).

\paragraph{Compute.} All experiments run on a 2-core CPU cloud machine; no GPU is used. Table~\ref{tab:compute} estimates the cost of one full reproduction. Where running times were recorded with the results (the running example, its 120-configuration grid, synthetic double descent, and Protocol B, whose per-case wall-clock times were logged and whose every fit and prediction was timed), we use them directly. For the other experiments, we timed a reduced run on the same machine (fewer repetitions) and scaled it linearly to the number of repetitions used in the paper. The total is about $9.8$ hours of wall-clock time, or at most about $20$ CPU-hours, since two cores were used; Protocol B with five seeds and its extension in Section~\ref{sec:add-twostage} account for about $60\%$ of it. It excludes exploratory runs and analysis scripts; the analysis and all figures take about one minute from the stored predictions.

\begin{table}[h]\centering\small
\caption{Estimated compute for one full reproduction on a 2-core CPU machine. ``Recorded'': timings stored with the results. ``Scaled'': a reduced run, timed and scaled linearly.}\label{tab:compute}
\resizebox{\linewidth}{!}{%
\begin{tabular}{llrl}\toprule
Experiment & Script & Wall-clock & Source\\\midrule
Running example (Table~\ref{tab:pilot}) & \texttt{pilot.py} & 7 min & recorded\\
Synthetic grid, 120 configurations (\S\ref{sec:exp:synth}) & \texttt{e1.py} & 39 min & recorded\\
Synthetic double descent (\S\ref{sec:exp:dd}) & \texttt{e2\_dd.py} & 4 min & recorded\\
Estimator bias (\S\ref{sec:estimator}) & \texttt{e3\_estimator.py} & 1 min & scaled ($\times20$)\\
Boosting, running example (\S\ref{sec:exp:boost}) & \texttt{e5\_boost.py} & 5 min & scaled ($\times30$)\\
Real-data ranking (\S\ref{sec:exp:real}) & \texttt{r1\_real.py} & 12 min & scaled ($\times12$)\\
Real boosting and double descent & \texttt{r23\_real.py} & 2 min & scaled ($\times6$)\\
Protocol A, 90 cases (\S\ref{sec:exp:select}) & \texttt{r4/r5\_select.py} & 88 min & scaled ($\times12$)\\
Stress tests (\S\ref{sec:stress}) & \texttt{s\_consensus.py} & 10 min & scaled ($\times60$)\\
Protocol B, 225 cases (\S\ref{sec:exp:select}) & \texttt{r6\_cost.py}, \texttt{r6\_seeds.py} & 200 min & recorded (5.4 CPU-h)\\
Two-stage and CMI (\S\ref{sec:add-twostage}) & \texttt{r8\_screen\_cmi.py} & 130 min & recorded (2.0 CPU-h)\\
Regression (\S\ref{sec:regression}) & \texttt{r9\_*.py} & 80 min & recorded\\
Disagreement table (\S\ref{sec:exp:synth}) & \texttt{t\_disagreement.py} & $<10$ min & single run\\
Mechanism checks (\S\ref{sec:mech-checks}) & \texttt{t\_*.py} & 1 min & timed\\\midrule
Total & & $\approx9.8$ h & ($\le20$ CPU-h)\\\bottomrule
\end{tabular}}
\end{table}

\section{Additional results}\label{sec:addresults}

\subsection{Estimating $V$ from few training runs}\label{sec:estimator}

\begin{figure}[h]\centering
\includegraphics[width=0.55\linewidth]{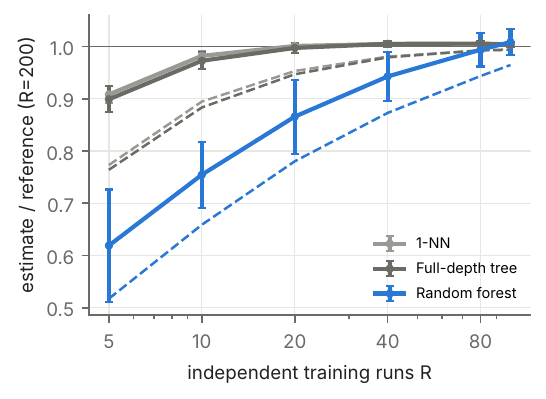}
\caption{Estimated expressed information as a fraction of the $R_{\rm ref}=200$ reference, against the number of independent training runs $R$. Solid: Miller--Madow corrected; dashed: plug-in. Error bars: one standard deviation over 20 random subsets.}\label{fig:est}
\end{figure}

We trained each of four learners on $R_{\rm ref}=200$ independent training sets of the running example ($\varepsilon=0.2$, $n=1024$, $5{,}000$ test inputs), and compared estimates from random subsets of $R$ runs against the plug-in value from all 200 (Figure~\ref{fig:est}). At $R=5$, the plug-in estimate recovers $77\%$ of the reference for 1-NN, $76\%$ for the full tree and $52\%$ for the random forest; with the Miller--Madow correction these become $91\%$, $90\%$ and $62\%$. From $R=20$, the corrected estimate is within $2\%$ of the reference for 1-NN and the tree, and within $14\%$ for the forest. The ranking of the four learners is the same at every $R\ge5$.

\subsection{Checks of the mechanism results}\label{sec:mech-checks}
\begin{itemize}[leftmargin=*]
\item \emph{Fixed-$k$ nearest neighbours} (Proposition~\ref{prop:knn}): on a one-dimensional task with $\eta_1(x)=\frac12+0.4\sin2\pi x$ and $n=20{,}000$ (300 training sets), the measured $\E V_n$ for $k=1,3,9$ matches the limit $\E\,h(\min\{\beta_k(\eta_1),1-\beta_k(\eta_1)\})$ to within $0.003$ bits (\texttt{t\_knn\_check.py}).
\item \emph{Rates} (Proposition~\ref{prop:knn-rate}): with $d=\alpha=1$, $k_n=n^{2/3}$ and $n$ from $250$ to $16{,}000$, the fitted slopes of $\log\E V_n$ and of $\log$ excess risk against $\log n$ are $-0.32$ and $-0.68$, against the predicted $-\frac13$ and $-\frac23$ (\texttt{t\_theory2.py}).
\item \emph{Averaging} (Proposition~\ref{prop:ensemble-full}): for 1-NN, a depth-2 tree and a misspecified logistic regression on the same one-dimensional task, the predicted excess risk of majority votes of $B\in\{1,5,25\}$ independent copies matches simulation to within $0.0005$ (\texttt{t\_ens\_check.py}).
\item \emph{Minimum-norm least squares} (Proposition~\ref{prop:ridgeless}) and \emph{log-loss} (Proposition~\ref{prop:logloss}): see the remarks after their proofs.
\end{itemize}

\subsection{Double descent on real data}\label{sec:add-dd}

\begin{figure}[h]\centering
\includegraphics[width=0.8\linewidth]{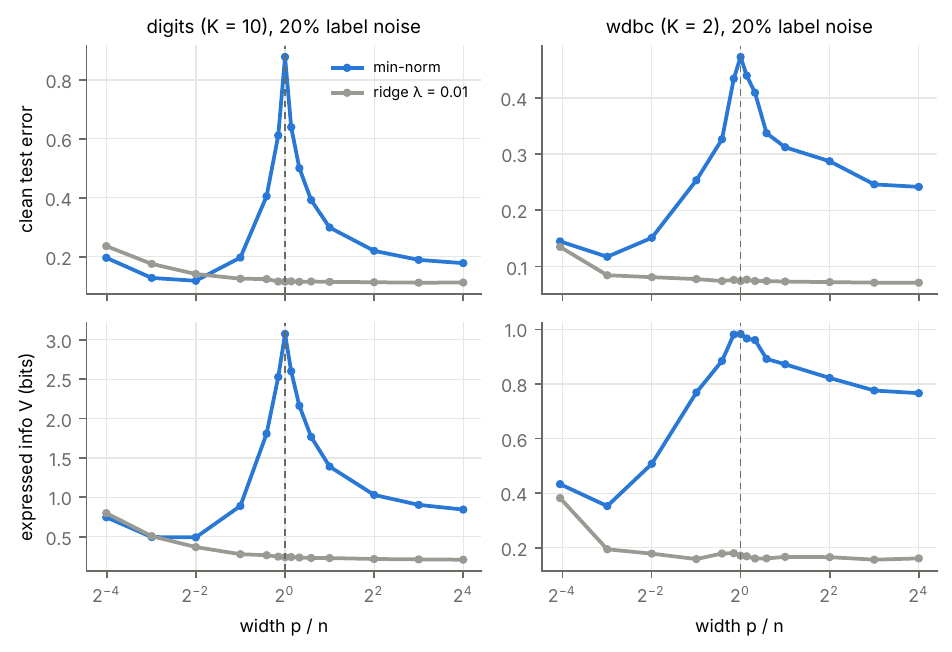}
\caption{Random-features double descent on real data with 20\% label noise: clean test error (top) and expressed information (bottom) against $p/n$.}\label{fig:ddreal}
\end{figure}

Random-features least squares on \texttt{digits} ($n=629$) and \texttt{wdbc} ($n=199$) with $20\%$ injected noise (Figure~\ref{fig:ddreal}) reproduces Section~\ref{sec:exp:dd}. Test error and $V$ peak together at $p=n$: on \texttt{digits}, error $0.88$ and $V=3.07$ bits against a ceiling of $\log_210=3.32$ bits; on \texttt{wdbc}, $0.47$ and $0.98$ bits. A small ridge removes both peaks. The Spearman correlation of $V$ with test error over widths is $0.97$--$0.99$ for minimum-norm fits. Table~\ref{tab:width} gives the width-selection results of Section~\ref{sec:exp:dd}.

\begin{table}[h]\centering\small
\caption{Choosing the width among interpolating (minimum-norm, $p\ge n$) random-features models: Spearman correlation between $V$ and test error over the eight widths, regret of min-$V$\prop, and expected regret of a random width. Best (lowest) regret per row in bold.}\label{tab:width}
\begin{tabular}{llccc}\toprule
Data & Noise & Spearman$(V,\text{error})$ & Min-$V$\prop{} regret & Random regret\\\midrule
Synthetic & $0$ & 1.00 & \best{0.0000} & 0.0018\\
Synthetic & $0.2$ & 1.00 & \best{0.0000} & 0.1242\\
\texttt{digits} & $0$ & 0.99 & \best{0.0000} & 0.1296\\
\texttt{digits} & $0.2$ & 1.00 & \best{0.0000} & 0.2339\\
\texttt{wdbc} & $0$ & 0.98 & \best{0.0029} & 0.0980\\
\texttt{wdbc} & $0.2$ & 1.00 & \best{0.0000} & 0.1017\\\bottomrule
\end{tabular}
\end{table}

\subsection{Boosting on real data}\label{sec:add-boost}

\begin{figure}[h]\centering
\includegraphics[width=0.9\linewidth]{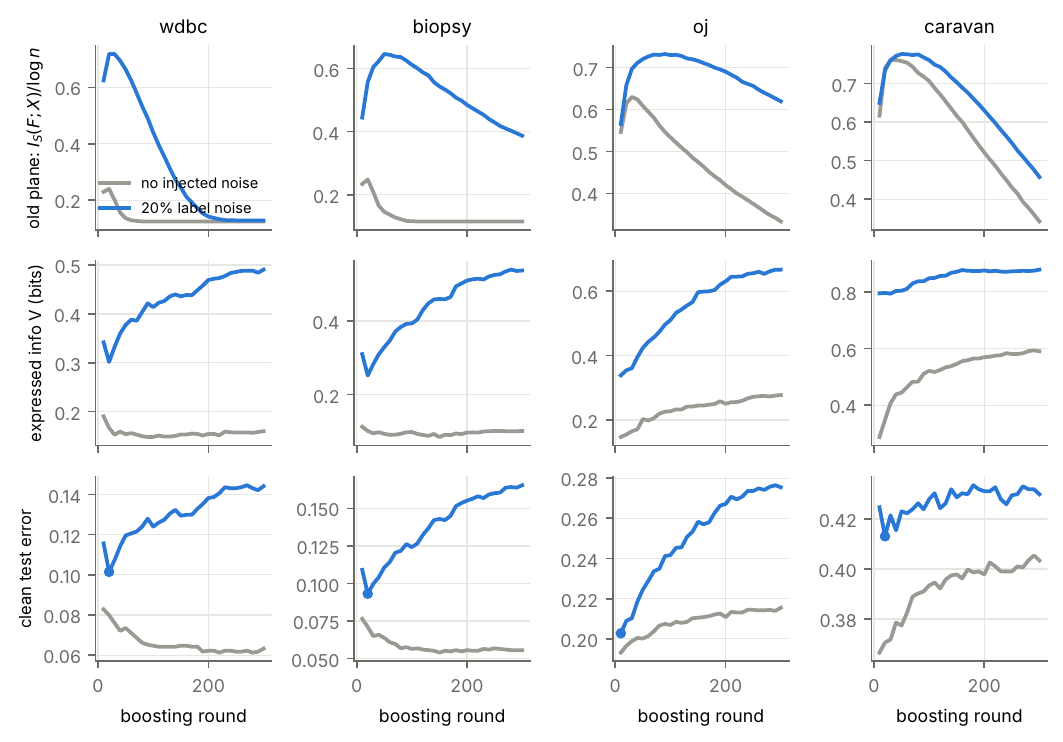}
\caption{Gradient boosting (exponential loss, shrinkage $0.1$) on four real datasets. Rows: the old information-plane coordinate, expressed information, clean test error (dot: best round under noise). Grey: original labels; blue: 20\% injected label noise.}\label{fig:boostreal}
\end{figure}

With $20\%$ injected noise (Figure~\ref{fig:boostreal}), clean test error rises from its minimum at round 10--20 to its final value on every dataset: $0.101\to0.144$ (\texttt{wdbc}), $0.093\to0.165$ (\texttt{biopsy}), $0.203\to0.275$ (\texttt{OJ}) and $0.413\to0.430$ (\texttt{Caravan}), while the old coordinate rises and then falls. $V$ rises with test error; their Spearman correlation along the trajectory is $0.82$--$1.00$.

\subsection{Model selection with all 19 candidates}\label{sec:add-select}

Table~\ref{tab:selectall} gives the Protocol-A comparison with all 19 configurations as candidates. Besides holdout, min-$V$ and the consensus criterion, it includes: \emph{corrected holdout} (backward correction with $\hat T$ estimated from the training labels); \emph{combined} (the modal predictor's noisy validation error $\hat e_m$ plus $\bar\pi(1-\hat e_mK/(K-1))$); and \emph{gated} (the one-standard-error rule on corrected holdout, then the least $V$). As a reference that is unavailable in practice, we also report holdout with clean validation labels.

\begin{table}[h]\centering\small
\caption{Protocol A with all 19 configurations: selection regret in clean test error (mean over cases; lower is better). Best per column in bold (excluding the clean-label reference). \prop: proposed in this paper. A random choice has regret $0.10$.}\label{tab:selectall}
\resizebox{\linewidth}{!}{%
\begin{tabular}{lccccc}\toprule
& \multicolumn{2}{c}{Symmetric (60 cases)} & \multicolumn{2}{c}{Class-dependent (30 cases)} & \\
Criterion & full validation & 40 points & full validation & 40 points & mean\\\midrule
Holdout (noisy labels) & 0.0085 & 0.0183 & 0.0168 & 0.0439 & 0.0219\\
Corrected holdout ($\hat T$ from training labels) & 0.0089 & 0.0185 & \best{0.0112} & 0.0271 & 0.0164\\
Min-$V$\prop{} (no validation labels) & 0.0063 & 0.0063 & 0.0537 & 0.0537 & 0.0300\\
Combined\prop & 0.0088 & 0.0180 & 0.0274 & 0.0351 & 0.0223\\
Gated\prop{} (corrected holdout, then $V$) & 0.0065 & 0.0068 & 0.0149 & 0.0270 & 0.0138\\
Consensus criterion\prop{} (no validation labels) & \best{0.0037} & \best{0.0037} & 0.0133 & \best{0.0133} & \best{0.0085}\\\midrule
\emph{Holdout, clean labels (reference)} & \emph{0.0040} & \emph{0.0087} & \emph{0.0053} & \emph{0.0085} & \emph{0.0066}\\\bottomrule
\end{tabular}}
\end{table}

Five findings, with 95\% bootstrap intervals over cases for differences from holdout:
\begin{enumerate}[leftmargin=*]
\item \emph{Min-$V$ alone} matches holdout with ample symmetric-noise validation data ($-0.002$ [$-0.006$, $0.002$]) and beats it with 40 points ($-0.012$ [$-0.016$, $-0.008$]). Under class-dependent noise it is worse ($+0.037$ [$0.017$, $0.059$]), its predicted blind spot (Proposition~\ref{prop:noise-shift}).
\item \emph{Noise correction helps only under asymmetric noise}, as Remark~\ref{rem:symcorr} predicts: under class-dependent noise, $-0.006$ [$-0.011$, $-0.001$] (full validation) and $-0.017$ [$-0.023$, $-0.012$] (40 points); with the oracle $T$, $-0.010$ and $-0.024$.
\item \emph{The consensus criterion has the lowest regret overall} ($0.0085$ against $0.0219$), and under symmetric noise it matches holdout with \emph{clean} labels.
\item \emph{The consensus need not be large or include the best models}: consensus from 5 or 10 random candidates gives $0.0059$ and $0.0050$ (symmetric) and $0.0153$, $0.0123$ (class-dependent); dropping the three best candidates gives $0.0057$ ($0.0136$).
\item \emph{The switch rule} (Section~\ref{sec:consensus}) has regret $0.0060$ over the 90 full-validation cases, against $0.0069$ for the consensus criterion, $0.0097$ for corrected holdout and $0.0113$ for holdout; it keeps $0.0037$ under symmetric noise and improves on both components under class-dependent noise ($0.0106$).
\end{enumerate}

\paragraph{Protocol B with all 19 candidates (Figure~\ref{fig:cost}).} Table~\ref{tab:protB-all} gives mean regrets over the 225 cases. Paired differences (cluster bootstrap over the 45 dataset--noise combinations): noise-corrected minus plain 5-fold CV $-0.0048$ [$-0.0073$, $-0.0025$], almost all of it under class-dependent noise ($-0.0113$ [$-0.0168$, $-0.0061$]); consensus (2 halves) minus 2-fold CV $+0.0007$ [$-0.0060$, $0.0072$]; consensus (2 halves) minus holdout $-0.0106$ [$-0.0187$, $-0.0028$]; consensus (5 folds) minus 5-fold CV $+0.0091$ [$+0.0008$, $+0.0197$]; min-$V_5$ minus 5-fold CV $+0.0486$ [$+0.0293$, $+0.0711$]. The switch rule's asymmetry test triggers in 7\% of symmetric-noise cases, 49\% of class-dependent cases and 12\% of instance-dependent cases. Computing $V$ or the consensus criterion from the CV fold models adds about $1\%$ of the CV cost.

\begin{table}[h]\centering\small
\caption{Protocol B with all 19 configurations: mean regret over 15 datasets $\times$ 5 seeds per noise model (lower is better). Best per column in bold. Cost: median CPU time relative to one holdout fit. \prop: proposed in this paper.}\label{tab:protB-all}
\begin{tabular}{lccccc}\toprule
Criterion & Symmetric & Class-dep. & Instance-dep. & All (225) & Cost\\\midrule
Holdout (80/20) & 0.0188 & 0.0595 & 0.0195 & 0.0326 & \best{1.00}\\
2-fold CV & 0.0089 & 0.0451 & \best{0.0099} & 0.0213 & 1.19\\
5-fold CV & 0.0079 & 0.0395 & 0.0116 & 0.0196 & 4.89\\
5-fold CV, noise-corrected & \best{0.0064} & \best{0.0282} & 0.0100 & \best{0.0149} & 4.89\\
Min-$V_2$\prop & 0.0580 & 0.1073 & 0.0593 & 0.0748 & 1.21\\
Min-$V_5$\prop & 0.0398 & 0.1217 & 0.0430 & 0.0682 & 4.94\\
Consensus criterion\prop, 2 halves & 0.0181 & 0.0344 & 0.0135 & 0.0220 & 1.21\\
Consensus criterion\prop, 5 folds & 0.0117 & 0.0623 & 0.0121 & 0.0287 & 4.94\\
Consensus, adaptive clustering\prop, 5 folds & 0.0129 & 0.0623 & 0.0128 & 0.0293 & 4.94\\
Switch rule\prop, 2 halves & 0.0178 & 0.0342 & 0.0132 & 0.0217 & 1.21\\
Switch rule\prop, 5 folds & 0.0116 & 0.0507 & 0.0140 & 0.0254 & 4.94\\\midrule
\emph{Random choice} & \emph{0.0990} & \emph{0.1301} & \emph{0.0668} & \emph{0.0986} & \\\bottomrule
\end{tabular}
\end{table}

\subsection{Two-stage selection, and details of the CMI comparison}\label{sec:add-twostage}

This subsection reports a further comparison on Protocol B (Section~\ref{sec:exp:select}), \emph{two-stage} selection, which first screens the candidates by how well they fit the training data and then applies min-$V_2$ among those kept, and it gives the estimation details of the comparison with CMI in Section~\ref{sec:exp:cmi}. Both use the same 225 cases (15 datasets $\times$ 3 noise models $\times$ 5 seeds) and the same regret as Table~\ref{tab:select}. The new fits were made with \texttt{r8\_screen\_cmi.py}; they reproduce the 2-fold models of Protocol B exactly (same data, noise, folds and seeds), so all criteria are compared on identical models.

\paragraph{Setup.} The training set of each case is split into pairs $(a_i,b_i)$, with $a_i$ from one fold and $b_i$ from the other fold of the Protocol B 2-fold split. This is a supersample in the sense of \citet{steinke2020cmi}. A draw of the selection vector $U\in\{0,1\}^{n/2}$ gives two models, one trained on the members selected by $U$ and one on the complement; the Protocol B 2-fold split is one such draw, and we add two draws with independent uniform $U$. Each model predicts on all training points and on the unlabelled test inputs.
\begin{itemize}[leftmargin=*]
\item \emph{Training-fit screens} use only the training labels: the training error of the two 2-fold models on their own halves; the legacy LMC score $I_S(F;Y)/H_S(Y)$ on the same halves; and a label-free estimate $\hat\rho$ of the noise rate, the disagreement of the plurality of all 19 configurations' out-of-fold predictions with the noisy labels (mean absolute error $0.061$).
\item \emph{e-CMI} (evaluated CMI; \citealp{steinke2020cmi}) is estimated per pair as the plug-in mutual information between the pair's loss pattern $(\ell(a_i),\ell(b_i))\in\{0,1\}^2$ and the selection bit, over the models of all draws, and averaged over pairs (in nats). \emph{f-CMI} \citep{harutyunyan2021fcmi} uses the prediction pattern instead; since the labels of the pair are fixed, it equals e-CMI for binary labels, and it gives the same regrets as e-CMI among the overparameterized candidates. The two bounds used for ranking are training error plus $\sqrt{2\,\overline{\text{e-CMI}}}$ \citep{steinke2020cmi} and training error plus the samplewise bound $\frac1{n/2}\sum_i\sqrt{2\,I_i}$ \citep{hellstrom2022samplewise}, with $I_i$ the estimate for pair $i$. With one draw (two fits, identical to those of 2-fold CV and of $V_2$), the per-pair estimate is $\ln2$ if the two models' loss patterns differ and $0$ otherwise; with three draws it is a plug-in estimate from six observations per pair, which is biased upwards, more so for candidates whose held-out losses vary between draws. This bias is why e-CMI from three pairs does worse than from one (Table~\ref{tab:cmi}).
\item The CMI bounds are loose: for interpolators, $\sqrt{2\,\overline{\text{e-CMI}}}$ is about $0.8$ (10th--90th percentile $0.63$--$1.01$, on seed 0), larger than the differences in error between candidates. With all 19 configurations, f-CMI from three pairs has regret $0.082$ and the $\sqrt{2\,\overline{\text{f-CMI}}}$ bound $0.055$, close to the e-CMI values in Table~\ref{tab:cmi}.
\end{itemize}

\begin{table}[h]\centering\small
\caption{Two-stage selection on Protocol B (225 cases; mean regret in clean test error, lower is better). Columns 2--5: all 19 configurations, by noise model; column 6: the six overparameterized configurations of Table~\ref{tab:select}, Panel B. Best per column in bold among the criteria shown; with all 19 candidates, noise-corrected 5-fold CV remains best overall ($0.0149$; Table~\ref{tab:protB-all}). Cost: median CPU time relative to one holdout fit, for the 19-candidate pool. $^\ast$The consensus criterion uses the consensus of all 19 configurations (with the six-candidate pool its cost is $2.39$). \prop: proposed in this paper.}\label{tab:twostage}
\resizebox{\linewidth}{!}{%
\begin{tabular}{lcccccc}\toprule
& \multicolumn{4}{c}{All 19 configurations} & Overparam. & \\
Criterion & Symmetric & Class-dep. & Instance-dep. & All & All & Cost\\\midrule
\multicolumn{7}{l}{\emph{Held-out labels}}\\
Holdout (80/20) & 0.0188 & 0.0595 & 0.0195 & 0.0326 & 0.0136 & 1.00\\
2-fold CV & 0.0089 & 0.0451 & \best{0.0099} & 0.0213 & 0.0098 & 1.19\\
5-fold CV & \best{0.0079} & 0.0395 & 0.0116 & \best{0.0196} & 0.0056 & 4.89\\
\multicolumn{7}{l}{\emph{No held-out labels, one stage}}\\
Min-$V_2$\prop{} & 0.0580 & 0.1073 & 0.0593 & 0.0748 & 0.0055 & 1.21\\
Consensus criterion\prop{}$^\ast$, 2 halves & 0.0181 & \best{0.0344} & 0.0135 & 0.0220 & \best{0.0033} & 1.21\\
\multicolumn{7}{l}{\emph{Two-stage (training labels only): screen by training fit, then min-$V_2$}}\\
Keep interpolators (training error $\le1\%$) & 0.0314 & 0.0885 & 0.0171 & 0.0457 & 0.0074 & 1.21\\
Keep legacy LMC score $\ge0.9$ & 0.0314 & 0.0902 & 0.0168 & 0.0461 & 0.0079 & 1.21\\
Keep training error below the median & 0.0182 & 0.0698 & 0.0140 & 0.0340 & 0.0080 & 1.21\\
Keep training error $\le\hat\rho+0.02$ & 0.0207 & 0.0951 & 0.0196 & 0.0451 & 0.0055 & 1.21\\
Keep training error $\le\hat\rho+0.02$, then consensus\prop{}$^\ast$ & 0.0151 & \best{0.0344} & 0.0133 & 0.0209 & \best{0.0033} & 1.21\\
\midrule
\emph{Random choice} & \emph{0.0990} & \emph{0.1301} & \emph{0.0668} & \emph{0.0986} & \emph{0.0850} & \\\bottomrule
\end{tabular}}
\end{table}

\paragraph{Findings on two-stage selection.}
\begin{enumerate}[leftmargin=*]
\item \emph{Screening by training fit does not rescue min-$V$ in a mixed pool.} Keeping only the interpolators before applying min-$V_2$ lowers its regret from $0.0748$ to $0.0457$ ($-0.029$ [$-0.058$, $-0.003$]), but it remains well behind holdout ($0.0326$), 2-fold CV ($0.0213$) and 5-fold CV ($0.0196$; difference $+0.026$ [$+0.015$, $+0.039$]). Screening by the legacy LMC score, which for hard predictions is a function of the training fit, behaves in the same way ($0.0461$).
\item \emph{The reason is that, under label noise, the best candidate usually does not fit the training data well.} With all 19 configurations, the best one (most often an RBF SVM with $C=1$, 63 of 225 cases, or a random forest with leaf size 5, 39 cases; the interpolating forest with leaf size 1 is best in 29 cases) is removed by the interpolation screen in $85\%$ of the cases. A training-fit screen removes the regularized candidates together with the underfitting ones, so the second stage chooses the best \emph{interpolator}, which is not the best candidate. A looser screen (training error below the median) keeps the best candidate in only $34\%$ of the cases and has regret $0.0340$.
\item \emph{The same happens with independent samples.} In Protocol A (Table~\ref{tab:select}, Panel A), restricting to interpolators and then taking the least $V$ has regret $0.065$ measured against the best of all 19 configurations, against $0.022$ for min-$V$ over all 19 and $0.011$ for holdout with the full validation set (\texttt{r8\_protA.py}). The small regret reported in Panel A of Table~\ref{tab:select} ($0.0010$ under symmetric noise) is measured against the best \emph{interpolator}: min-$V$ chooses well among interpolators, and the training-fit screen is what loses.
\item \emph{A screen anchored at the estimated noise rate is harmless but adds little.} Keeping candidates with training error at most $\hat\rho+0.02$ removes only clear underfitters (17.6 of 19 kept on average) and never removes the best candidate. Followed by the consensus criterion it has regret $0.0209$, on a par with 2-fold and 5-fold CV ($+0.0013$ [$-0.0039$, $+0.0062$] against 5-fold CV) and slightly below the consensus criterion alone ($0.0220$; $-0.0011$ [$-0.0031$, $0$]). Followed by min-$V_2$ it does not help enough ($0.0451$), because the remaining rigid candidates still have low $V$ and large modal bias.
\item \emph{Among overparameterized candidates the screens are unnecessary}: they can only remove candidates, and the interpolation screen removes the best one in $12\%$ of cases (regret $0.0074$ against $0.0055$ for min-$V_2$).
\end{enumerate}
In short, the first stage of a two-stage rule must remove candidates with \emph{modal bias}, not candidates with training error. Training error is a poor proxy for modal bias under label noise, because good candidates do not fit noisy labels. The consensus criterion estimates modal bias directly, and this is why it, rather than a training-fit screen followed by min-$V$, is our recommendation for mixed pools.

\subsection{Stress tests of the consensus criterion}\label{sec:stress}

\begin{figure}[h]\centering
\includegraphics[width=\linewidth]{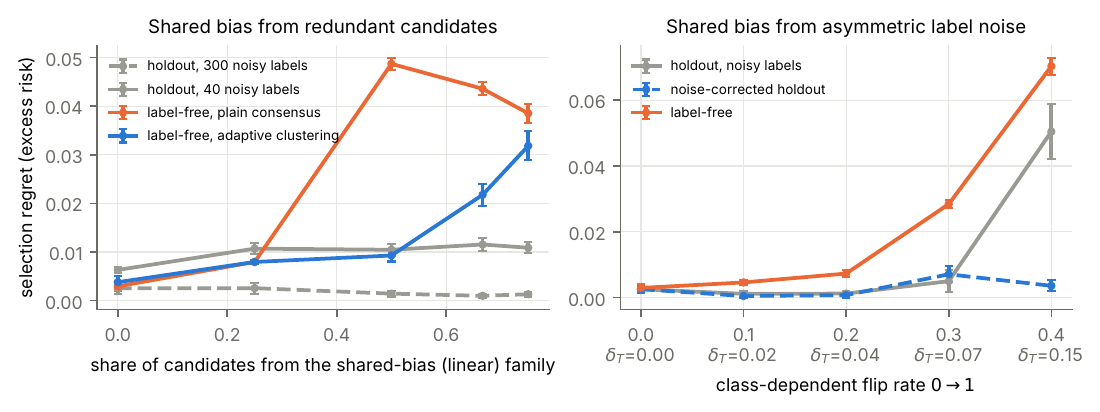}
\caption{Stress tests of the consensus criterion (synthetic task with known Bayes classifier; means and standard errors over 12 replicates). \emph{Left:} a growing share of candidates with a common misspecification. \emph{Right:} growing asymmetric label noise; $\delta_T$ is the shared-bias mass of Proposition~\ref{prop:noise-shift}.}\label{fig:stress}
\end{figure}

\paragraph{Redundant candidates (Figure~\ref{fig:stress}, left).} With at most a quarter of the pool from the shared-bias family, the plain consensus criterion is competitive (regret $0.0029$, against $0.0025$ for holdout with 300 labels and $0.0063$ with 40). Once half the pool shares the bias, its regret jumps to $0.049$ and the consensus error rises from $\delta=0.080$ to $0.148$. Adaptive clustering lowers the regret to $0.0093$, below holdout with 40 labels ($0.0104$). With two thirds or more of the pool redundant, it helps but is not enough ($0.022$--$0.032$). In all 1{,}248 pools, the candidate with the smallest disagreement had modal-bias mass within $2\delta$ of the best, as Proposition~\ref{prop:disagreement} guarantees.

\paragraph{Asymmetric noise (Figure~\ref{fig:stress}, right).} Flipping labels $0\to1$ with probability $f$ grows the shared-bias mass from $\delta_T=0$ to $0.15$. The consensus criterion degrades from $0.0029$ to $0.070$; holdout on noisy labels is robust up to $f=0.3$ but fails at $f=0.4$ ($0.051$); backward-corrected holdout with the true $T$ stays below $0.008$.

\section{Extension to regression}\label{sec:regression}

The main paper treats classification, with a brief squared-loss statement in Section~\ref{sec:sq}. This section works out what carries over to regression: a storage statement (Proposition~\ref{prop:reg-storage}), a label-free selection rule with a guarantee (Proposition~\ref{prop:reg-select}), a regression consensus criterion (Proposition~\ref{prop:reg-cons}), and one experiment. In short, storage carries over in a stronger form, and selection carries over only when the candidates have similar bias.

\paragraph{Setting.} Let $Y\in\mathbb R$, with squared loss and $f^*(x)=\E[Y\mid X=x]$, so the excess risk of a predictor $f$ is $\E(f(X)-f^*(X))^2$. A learner outputs $f_W$, random through the sample and the seed. Write $\bar f(x)=\E f_W(x)$ for its mean prediction, $v(x)=\Var f_W(x)$ for its variance and $B(x)=(\bar f(x)-f^*(x))^2$ for its squared bias. The mean prediction plays the role of the modal prediction, and $B$ the role of modal bias. Theorem~\ref{thm:sq} gives $v(x)=\lim_{\sigma\to\infty}2\sigma^2V_\sigma(x)$ and the exact decomposition
\begin{equation}\label{eq:reg-bv}
\E R(f_W)-R^*\;=\;\E B(X)+\E v(X).
\end{equation}
So in regression the variance plays the role of $V$: it is expressed information measured at low resolution.

\subsection{Storage}

\begin{proposition}[Storage of label noise, regression]\label{prop:reg-storage}
Assume $Y\mid X$ has a density with finite conditional differential entropy $h(Y\mid X)$, and let $D=\frac1n\sum_{i=1}^n\E(f_W(X_i)-Y_i)^2$ be the learner's expected training MSE.
\begin{enumerate}[label=(\alph*),leftmargin=*]
\item For every learner, $I(W;Y^n\mid X^n)\ge n\big[h(Y\mid X)-\frac12\log(2\pi e\,D)\big]_+$. With Gaussian noise of variance $\sigma^2$, the bound is $\frac n2\log(\sigma^2/D)$ when $D\le\sigma^2$. In particular, every exact interpolator ($D=0$) stores infinitely much information about its training labels.
\item Fix $\Delta>0$ and let $Y^\Delta=\Delta\lfloor Y/\Delta+\frac12\rfloor$ be the labels rounded to the grid $\Delta\mathbb Z$. If the learner interpolates at resolution $\Delta$, i.e.\ $f_W(X_i)$ rounds to $Y_i^\Delta$ for every $i$ (for example, if it interpolates the $Y_i$ exactly), then
\begin{align*}
I(W;Y^n\mid X^n)\;\ge\;I\big(W;(Y^\Delta)^n\mid X^n\big)\;&=\;n\,H(Y^\Delta\mid X)\\
&=\;n\big[h(Y\mid X)+\log\tfrac1\Delta+o(1)\big]\qquad(\Delta\to0).
\end{align*}
\end{enumerate}
\end{proposition}
\begin{proof}
(a) The fitted values $\hat Y_i=f_W(X_i)$ are functions of $(W,X^n)$, so $I(W;Y^n\mid X^n)\ge I(\hat Y^n;Y^n\mid X^n)$. Given $X^n$ the labels are independent, so $I(\hat Y^n;Y^n\mid X^n)\ge\sum_iI(\hat Y^n;Y_i\mid X^n)\ge\sum_iI(\hat Y_i;Y_i\mid X^n)$. For each $i$,
\[
I(\hat Y_i;Y_i\mid X^n)=h(Y_i\mid X^n)-h(Y_i\mid\hat Y_i,X^n)\ \ge\ h(Y\mid X)-h(Y_i-\hat Y_i)\ \ge\ h(Y\mid X)-\tfrac12\log(2\pi e\,D_i),
\]
with $D_i=\E(Y_i-\hat Y_i)^2$, because conditioning reduces entropy and the Gaussian maximizes entropy for a given second moment. Summing over $i$ and using the concavity of $\log$ gives the bound; mutual information is non-negative, hence the positive part. With Gaussian noise $h(Y\mid X)=\frac12\log(2\pi e\sigma^2)$.
(b) $Y^\Delta$ is a function of $Y$, which gives the inequality. Interpolation at resolution $\Delta$ makes $(Y^\Delta)^n$ a function of $(W,X^n)$, so $I(W;(Y^\Delta)^n\mid X^n)=H((Y^\Delta)^n\mid X^n)=nH(Y^\Delta\mid X)$; this is Theorem~\ref{thm:storage} applied to the discrete labels $Y^\Delta$. The last equality is the standard relation between the entropy of a fine quantization and differential entropy, valid for instance when $Y\mid X$ has a bounded density with finite second moment.
\end{proof}

In plain terms, storage is again decided by the training fit and not by generalization: every learner with the same training MSE faces the same lower bound. In regression the stored amount is not even finite for an interpolator; it grows by $\log(1/\Delta)$ bits per example for each refinement of the resolution. This makes the conclusion of Corollary~\ref{cor:xr}, that storage-based quantities cannot separate interpolators, even starker. In the experiment of Table~\ref{tab:reg-exact} at noise level $0.5$, the weakly regularized MLP must store at least $0.28$ bits per example and the random forest with leaf size 1 at least $1.05$, while extra-trees, kernel ridgeless regression and distance-weighted $k$-NN interpolate exactly and store infinitely much.

\subsection{Selection}

\paragraph{The rule.} For each candidate $j$, train two copies on disjoint halves of the data, $f_{j,1}$ and $f_{j,2}$, and compute on unlabelled inputs $x_1,\dots,x_M$
\[
\widehat{\Var}_{2,j}\;=\;\frac1M\sum_{m=1}^M\tfrac12\big(f_{j,1}(x_m)-f_{j,2}(x_m)\big)^2 .
\]
\emph{Min-variance} chooses the candidate with the smallest value. The two copies are independent, so $\E\,\widehat{\Var}_{2,j}$ is exactly the mean variance of the learner at half the sample size. The rule uses no held-out labels, and by Theorem~\ref{thm:sq} it is min-$V$ at low resolution. With $R$ disjoint pairs, the $R$ estimates are averaged.

\begin{proposition}[Min-variance selection]\label{prop:reg-select}
Let candidates $j=1,\dots,J$ have squared biases $B_j=\E B_j(X)$, variances $v_j=\E v_j(X)$ and excess risks $\mathrm{Exc}_j=B_j+v_j$.
\begin{enumerate}[label=(\alph*),leftmargin=*]
\item If all $B_j$ are equal (in particular, if all are zero), ranking by $v_j$ is exactly ranking by excess risk.
\item In general, if $\hat\jmath$ minimizes estimates $\hat v_j$ with $|\hat v_j-v_j|\le\varepsilon$ for all $j$, then $\mathrm{Exc}_{\hat\jmath}\le\min_j\mathrm{Exc}_j+\max_jB_j-\min_jB_j+2\varepsilon$.
\item If the prediction distribution at $x$ is Gaussian, $V_\sigma(x)=\frac12\log(1+v(x)/\sigma^2)$ is increasing in $v(x)$ at every resolution $\sigma$.
\end{enumerate}
\end{proposition}
\begin{proof}
(a) is \eqref{eq:reg-bv}. (b) Let $j^*$ minimize $\mathrm{Exc}_j$. Then $v_{\hat\jmath}\le\hat v_{\hat\jmath}+\varepsilon\le\hat v_{j^*}+\varepsilon\le v_{j^*}+2\varepsilon$, so $\mathrm{Exc}_{\hat\jmath}\le B_{\hat\jmath}+v_{j^*}+2\varepsilon=\mathrm{Exc}_{j^*}+(B_{\hat\jmath}-B_{j^*})+2\varepsilon$. (c) is the capacity formula of the Gaussian channel.
\end{proof}

\begin{proposition}[Regression consensus criterion]\label{prop:reg-cons}
Let $\hat m_j=\frac12(f_{j,1}+f_{j,2})$, $\hat v_j(x)=\frac12(f_{j,1}(x)-f_{j,2}(x))^2$, and let $c(x)$ be a reference function (we use the pointwise median of all candidates' $\hat m_j$). Score candidate $j$ by $\mathrm{LF}_j=\frac1M\sum_m\big[(\hat m_j(x_m)-c(x_m))^2+\frac12\hat v_j(x_m)\big]$. If $c=f^*$, then $\E\,\mathrm{LF}_j$ is the excess risk of candidate $j$ at half the sample size. For any $c$, with $\|g\|^2=\E g(X)^2$, $\big|\,\|\hat m_j-c\|-\|\hat m_j-f^*\|\,\big|\le\|c-f^*\|$, so the bias part of the score is off by at most the consensus error, with no independence assumption.
\end{proposition}
\begin{proof}
$\E(\hat m_j-f^*)^2=B_j+v_j/2$ pointwise (the mean of two independent copies) and $\E\hat v_j=v_j$, so $\E[(\hat m_j-f^*)^2+\hat v_j/2]=B_j+v_j$ at sample size $n/2$. The second claim is the triangle inequality in $L^2(P_X)$.
\end{proof}

\begin{remark}[Why bias matters more in regression]\label{rem:reg-bias}
In classification a learner's bias costs nothing as long as its modal prediction is on the correct side of $\frac12$, which is why modal bias was negligible among flexible interpolators (Section~\ref{sec:sel:rank}). In regression every unit of squared bias is paid in full. Interpolation does not limit bias either: the memorize-and-smooth construction (Proposition~\ref{prop:memsmooth}) shows that an interpolator can be built on top of \emph{any} smoother, and it inherits the smoother's bias. Distance-weighted 25-NN is such an interpolator. It returns the training label at a training input but averages 25 neighbours elsewhere, so in our experiment it has the smallest variance of all interpolators and the largest bias (Table~\ref{tab:reg-exact}). This is the regression counterpart of the rigid-learner failure of min-$V$, and the failure that part (b) of Proposition~\ref{prop:reg-select} allows.
\end{remark}

\subsection{Experiment}

\begin{figure}[h]\centering\includegraphics[width=0.9\linewidth]{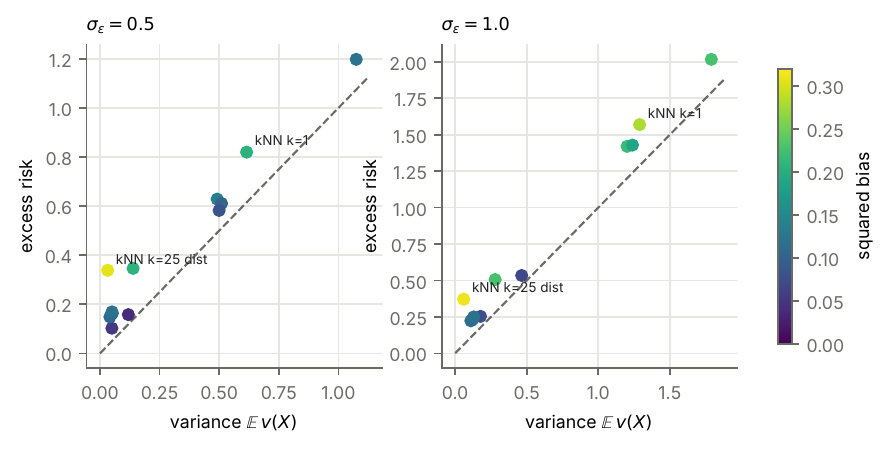}
\caption{Excess risk against variance for the interpolating regressors of Table~\ref{tab:reg-exact} (RBF ridgeless regression omitted at noise level $1$, where its excess risk is $4.9$). Dashed: excess $=$ variance; the vertical distance to it is the squared bias (colour).}\label{fig:reg-exact}
\end{figure}

\paragraph{Design.} This is a Protocol B analogue for regression: one noisy dataset and matched, measured compute (\texttt{r9\_regression.py}).
\begin{itemize}[leftmargin=*]
\item \emph{Data.} Eleven datasets: Friedman \#1 (synthetic, with known $f^*$, so its regret is exact excess risk), \texttt{diabetes}, and nine PMLB regression benchmarks (\texttt{echoMonths}, \texttt{2dplanes}, \texttt{puma8NH}, \texttt{cpu\_act}, \texttt{wind}, \texttt{pollen}, \texttt{houses}, \texttt{fried}, \texttt{mv}). Targets are standardized and at most 1{,}000 training points are used. A 30\% test set keeps the original targets. The training targets receive Gaussian noise with standard deviation $0.5$ or $1$, in units of the target's standard deviation. With five seeds this gives 110 cases.
\item \emph{Candidates.} 22 regressors. Twelve are designed to interpolate: 1-NN, distance-weighted 5-NN and 25-NN, a full tree, extra-trees, a random forest without bootstrap, a random forest with leaf size 1, gradient boosting (300 depth-6 trees, learning rate 1), kernel ridgeless regression with RBF and with Laplace kernel, min-norm random ReLU features ($p=4n$), and a weakly regularized 512-unit MLP. The other ten are regularized: $k$-NN, shallower trees, a random forest with leaf size 5, shrunk boosting, kernel ridge, linear ridge, and an MLP with weight decay.
\item \emph{Pools.} Three candidate pools: the first seven interpolators, which mirror the six overparameterized candidates of Table~\ref{tab:select}, Panel B; all twelve interpolators; and all 22 regressors.
\item \emph{Regret.} Clean test MSE of the chosen candidate's full refit minus the best candidate's. CPU time of every fit and prediction is recorded.
\end{itemize}
The analysis (\texttt{r9\_analyse.py}) was fixed before the results were computed. The exact decomposition on Friedman \#1 (Table~\ref{tab:reg-exact}, Figure~\ref{fig:reg-exact}) uses 10 independent training samples of size 500 per learner (\texttt{r9\_exact.py}).

\begin{table}[h]\centering\small
\caption{Selection among regressors (110 cases; mean regret in clean test MSE, in units of the target variance; lower is better). ``Labels'': uses held-out labels. ``Exact'' and ``cost'' are for the 7-interpolator pool (cases of 110 in which the best candidate was chosen; median CPU time relative to one holdout fit). Best per column in bold. The median-candidate row gives the regret of a typical candidate; the mean is dominated by catastrophic interpolators (RBF ridgeless regression and min-norm random features have test MSE above 10 in 40 and 22 cases). \prop: proposed in this paper.}\label{tab:reg-main}
\resizebox{\linewidth}{!}{%
\begin{tabular}{llccccccc}\toprule
& & \multicolumn{3}{c}{7 interpolators} & 12 interpolators & All 22 & & \\
Criterion & Labels & noise $0.5$ & noise $1$ & All & All & All & Exact & Cost\\\midrule
Holdout (80/20) & yes & 0.0065 & 0.0111 & 0.0088 & 0.0201 & 0.0219 & 99 & 1.00\\
2-fold CV & yes & 0.0019 & \best{0.0017} & \best{0.0018} & 0.0120 & 0.0155 & \best{106} & 1.10\\
5-fold CV & yes & \best{0.0015} & 0.0038 & 0.0026 & \best{0.0076} & 0.0109 & 102 & 5.08\\
2-fold CV, 3 random splits & yes & 0.0017 & 0.0035 & 0.0026 & 0.0102 & \best{0.0095} & 104 & 3.28\\
Min-variance\prop{} (2 disjoint halves) & no & 0.0059 & \best{0.0017} & 0.0038 & 0.0947 & 0.1076 & 97 & 1.12\\
Min-variance\prop{}, averaged over 3 pairs & no & 0.0071 & \best{0.0017} & 0.0044 & 0.0938 & 0.1032 & 96 & 3.33\\
Consensus criterion\prop{}, 2 halves & no & 0.0157 & 0.0116 & 0.0137 & 0.0261 & 0.0354 & 83 & 2.34\\
Consensus criterion\prop{}, 3 pairs & no & 0.0157 & 0.0144 & 0.0151 & 0.0266 & 0.0370 & 82 & 6.95\\
\midrule
\emph{Median candidate} & & \emph{0.481} & \emph{1.231} & \emph{0.856} & \emph{0.276} & \emph{0.185} & & \\\bottomrule
\end{tabular}}
\end{table}

\begin{table}[h]\centering\small
\caption{Exact decomposition on Friedman \#1 for the interpolating regressors (10 independent training samples of size $500$; standardized units), sorted by excess risk at noise level $0.5$. ``Storage $\ge$'': the lower bound of Proposition~\ref{prop:reg-storage}(a), in bits per example, at noise level $0.5$. $\widehat{\Var}_2$: the two-halves estimate from a single dataset.}\label{tab:reg-exact}
\resizebox{\linewidth}{!}{%
\begin{tabular}{lccccccccc}\toprule
& \multicolumn{6}{c}{Noise level $0.5$} & \multicolumn{3}{c}{Noise level $1$}\\
Learner & train MSE & storage $\ge$ & excess & bias$^2$ & variance & $\widehat{\Var}_2$ & excess & bias$^2$ & variance\\\midrule
MLP512 alpha=1e-05 & 0.170 & 0.28 & 0.102 & 0.053 & 0.050 & 0.089 & 0.254 & 0.077 & 0.177\\
ExtraTrees & 0.000 & $\infty$ & 0.148 & 0.107 & 0.041 & 0.049 & 0.223 & 0.113 & 0.110\\
KRR ridgeless Laplace & 0.000 & $\infty$ & 0.158 & 0.039 & 0.119 & 0.134 & 0.534 & 0.069 & 0.465\\
RF leaf=1 & 0.059 & 1.05 & 0.163 & 0.111 & 0.051 & 0.055 & 0.233 & 0.111 & 0.122\\
RF no-bootstrap mf=1/3 & 0.000 & $\infty$ & 0.169 & 0.119 & 0.050 & 0.070 & 0.250 & 0.120 & 0.129\\
kNN k=25 dist & 0.000 & $\infty$ & 0.338 & 0.307 & 0.032 & 0.036 & 0.372 & 0.312 & 0.060\\
kNN k=5 dist & 0.000 & $\infty$ & 0.346 & 0.207 & 0.139 & 0.155 & 0.507 & 0.227 & 0.279\\
RF min-norm p=4n & 0.000 & $\infty$ & 0.582 & 0.083 & 0.499 & 0.576 & 2.017 & 0.227 & 1.789\\
GB 300x d6 lr1 & 0.000 & $\infty$ & 0.611 & 0.101 & 0.510 & 0.602 & 1.428 & 0.190 & 1.239\\
Tree depth=None & 0.000 & $\infty$ & 0.628 & 0.137 & 0.491 & 0.593 & 1.420 & 0.219 & 1.201\\
kNN k=1 & 0.000 & $\infty$ & 0.820 & 0.205 & 0.615 & 0.738 & 1.569 & 0.280 & 1.289\\
KRR ridgeless & 0.000 & $\infty$ & 1.198 & 0.123 & 1.075 & 1.301 & 4.871 & 0.475 & 4.396\\\bottomrule
\end{tabular}}
\end{table}

\paragraph{Findings.} Brackets give 95\% cluster-bootstrap intervals (resampling the 22 dataset--noise combinations) for paired differences.
\begin{enumerate}[leftmargin=*]
\item \emph{Among the seven standard interpolators, min-variance is competitive, and best at high noise.} Its regret is $0.0038$, against $0.0088$ for holdout ($-0.0050$ [$-0.0115$, $+0.0005$]), $0.0018$ for 2-fold CV ($+0.0020$ [$0$, $+0.0050$]) and $0.0026$ for 5-fold CV ($+0.0012$ [$-0.0022$, $+0.0047$]), at the cost of 2-fold CV. At noise level $1$ it ties for best with 2-fold CV ($0.0017$) and beats 5-fold CV ($0.0038$) and holdout ($0.0111$); at noise level $0.5$ it is behind cross-validation. This pool is easy, though: the random forest with leaf size 1 or the MLP is best in every case, and all criteria except the consensus criterion choose the best candidate in at least 96 of the 110 cases.
\item \emph{When the pool contains smoothed interpolators, min-variance fails.} With all twelve interpolators its regret is $0.095$, against $0.008$--$0.020$ for the labelled criteria. It picks distance-weighted 25-NN in 107 of the 110 cases, the interpolator with the least variance and the most bias (Remark~\ref{rem:reg-bias}). On Friedman \#1 the squared bias of the interpolators ranges from $0.04$ to $0.31$, which is large compared with the differences in variance among the good ones.
\item \emph{The consensus criterion repairs most of this failure, but does not reach cross-validation.} It lowers the regret to $0.026$ with twelve interpolators and to $0.035$ with all 22 (min-variance: $0.108$), but 2-fold CV ($0.012$ and $0.016$) remains better. Robust variants of the consensus (the median over interpolators only, a trimmed mean, the median over the more stable half; weight $\frac12$ or $1$ on the variance term) give $0.020$--$0.030$ with twelve interpolators. A plain mean fails ($0.09$), because catastrophic interpolators pull it away.
\item \emph{Averaging over three pairs does not help} ($0.0044$ and $0.094$), unlike in classification (Section~\ref{sec:exp:cmi}): the bottleneck is bias, not the noise of the variance estimate.
\item \emph{The exact decomposition behaves as expected} (Table~\ref{tab:reg-exact}, Figure~\ref{fig:reg-exact}). 1-NN, the full tree, boosting with learning rate 1, min-norm random features and RBF ridgeless regression are tempered or catastrophic through their variance. Among the good interpolators, bias and variance are comparable. The two-halves estimate tracks the full-size variance with a modest upward drift, since it measures the variance at half the sample size.
\end{enumerate}
In summary, the storage half of the framework carries over to regression unchanged, and in a stronger form. The selection half carries over only partly: min-variance works when the interpolators being compared have similar bias, especially at high noise. Unlike in classification, interpolation does not keep bias small in regression, so in general a bias estimate is needed, and even with the consensus criterion 2-fold cross-validation of the same cost is better. Part of the reason is that additive zero-mean label noise only shifts held-out MSE by the constant $\sigma^2$, so it leaves the ranking by cross-validation unbiased, and validation on noisy labels is less handicapped than it is in classification.

\section{Code package}\label{sec:code}

The code package, available at \url{https://github.com/nnikolaou/GLMC}, contains every script used for the paper, the result files, and the figure scripts. Table~\ref{tab:code} maps each result to its script. The \texttt{README.md} in the package gives installation and run instructions; in brief:
{\small
\begin{verbatim}
pip install -r requirements.txt
python data/fetch_pmlb.py          # ten PMLB datasets for Protocol B
bash run_all.sh figures            # figures from the stored results (minutes)
bash run_all.sh all                # all experiments from scratch (~10 h on 2 cores)
\end{verbatim}
}
All scripts are seeded. The stored JSON results reproduce every number in the paper; the large intermediate prediction files (\texttt{*.npz}) are regenerated by the experiment scripts.

\begin{table}[h]\centering\small
\caption{Map from results to scripts (all in \texttt{code/}).}\label{tab:code}
\begin{tabular}{p{0.42\linewidth}p{0.50\linewidth}}\toprule
Result & Script(s) $\to$ output\\\midrule
Table~\ref{tab:pilot} & \texttt{pilot.py} $\to$ \texttt{results.json}\\
\S\ref{sec:exp:synth}, Figure~\ref{fig:synth} & \texttt{e1.py} $\to$ \texttt{e1\_results.json}; \texttt{make\_fig\_e1\_band.py} (left), \texttt{make\_figs.py} (right)\\
\S\ref{sec:exp:dd}, Figure~\ref{fig:dd}, Table~\ref{tab:width} & \texttt{e2\_dd.py}, \texttt{r23\_real.py}; \texttt{width\_select.py}\\
\S\ref{sec:exp:boost}, Figures~\ref{fig:boost}, \ref{fig:boostreal} & \texttt{e5\_boost.py}, \texttt{r23\_real.py}; \texttt{make\_figs*.py}\\
\S\ref{sec:exp:real}, Figure~\ref{fig:real} & \texttt{r1\_real.py}; \texttt{make\_figs\_real.py}\\
Table~\ref{tab:select}, Panel A; Table~\ref{tab:selectall} & \texttt{r4\_select.py}, \texttt{r5\_select.py}; \texttt{r4\_analyse.py}, \texttt{r5\_analyse.py}, \texttt{r5\_consensus\_check.py}, \texttt{r5\_remedy.py}\\
Table~\ref{tab:select}, Panel B; Figure~\ref{fig:overparam} & \texttt{r6\_cost.py}, \texttt{r6\_seeds.py}; \texttt{r6\_overparam.py}, \texttt{r6\_multiseed.py}; \texttt{make\_fig\_overparam.py}\\
Figure~\ref{fig:cost}; Table~\ref{tab:protB-all} & \texttt{r6\_cost.py}, \texttt{r6\_seeds.py}; \texttt{r6\_analyse.py}, \texttt{r6\_multiseed.py}; \texttt{make\_fig\_cost.py}\\
Table~\ref{tab:cmi} (CMI), Table~\ref{tab:twostage} (two-stage) & \texttt{r8\_screen\_cmi.py} $\to$ \texttt{r8\_*.npz}; \texttt{r8\_analyse.py} $\to$ \texttt{r8\_analysis.json}; \texttt{r8\_protA.py} (Protocol A)\\
Tables~\ref{tab:reg-main}, \ref{tab:reg-exact}, Figure~\ref{fig:reg-exact} (regression) & \texttt{r9\_regression.py} (and \texttt{--batch-b}) $\to$ \texttt{r9*\_*.npz}; \texttt{r9\_exact.py}; \texttt{r9\_analyse.py}, \texttt{r9\_tables.py}\\
Table~\ref{tab:dis} & \texttt{t\_disagreement.py}\\
Figure~\ref{fig:est} & \texttt{e3\_estimator.py}; \texttt{make\_figs.py}\\
Figure~\ref{fig:stress} & \texttt{s\_consensus.py}; \texttt{s\_analyse.py}, \texttt{s\_adaptive.py}; \texttt{make\_fig\_consensus.py}\\
\S\ref{sec:mech-checks} & \texttt{t\_knn\_check.py}, \texttt{t\_ens\_check.py}, \texttt{t\_theory2.py}\\
Shared helpers & \texttt{pilot.py} (running example), \texttt{realdata.py} (datasets)\\\bottomrule
\end{tabular}
\end{table}

%% file: refs.bib
@article{nikolaou2020margin, title={Margin Maximization as Lossless Maximal Compression}, author={Nikolaou, Nikolaos and Reeve, Henry and Brown, Gavin}, journal={arXiv preprint arXiv:2001.10318}, year={2020}}

@inproceedings{nikolaou2021lossless, title={Lossless Compression \& Generalization in Overparameterized Models: The Case of Boosting}, author={Nikolaou, Nikolaos}, booktitle={ICLR (short paper)}, year={2021}}

@article{tishby2000ib, title={The Information Bottleneck Method}, author={Tishby, Naftali and Pereira, Fernando C. and Bialek, William}, journal={arXiv preprint physics/0004057}, year={2000}}

@inproceedings{tishby2015deep, title={Deep Learning and the Information Bottleneck Principle}, author={Tishby, Naftali and Zaslavsky, Noga}, booktitle={IEEE Information Theory Workshop}, year={2015}}

@article{shwartzziv2017opening, title={Opening the Black Box of Deep Neural Networks via Information}, author={Shwartz-Ziv, Ravid and Tishby, Naftali}, journal={arXiv preprint arXiv:1703.00810}, year={2017}}

@inproceedings{saxe2018ib, title={On the Information Bottleneck Theory of Deep Learning}, author={Saxe, Andrew M. and others}, booktitle={ICLR}, year={2018}}

@inproceedings{kolchinsky2019caveats, title={Caveats for Information Bottleneck in Deterministic Scenarios}, author={Kolchinsky, Artemy and Tracey, Brendan D. and Van Kuyk, Steven}, booktitle={ICLR}, year={2019}}

@article{amjad2019learning, title={Learning Representations for Neural Network-Based Classification Using the Information Bottleneck Principle}, author={Amjad, Rana Ali and Geiger, Bernhard C.}, journal={IEEE TPAMI}, year={2019}}

@inproceedings{goldfeld2019estimating, title={Estimating Information Flow in Deep Neural Networks}, author={Goldfeld, Ziv and others}, booktitle={ICML}, year={2019}}

@article{strouse2017dib, title={The Deterministic Information Bottleneck}, author={Strouse, DJ and Schwab, David J.}, journal={Neural Computation}, year={2017}}

@article{achille2018emergence, title={Emergence of Invariance and Disentanglement in Deep Representations}, author={Achille, Alessandro and Soatto, Stefano}, journal={JMLR}, year={2018}}

@article{shamir2010learning, title={Learning and Generalization with the Information Bottleneck}, author={Shamir, Ohad and Sabato, Sivan and Tishby, Naftali}, journal={Theoretical Computer Science}, year={2010}}

@inproceedings{kawaguchi2023ib, title={How Does Information Bottleneck Help Deep Learning?}, author={Kawaguchi, Kenji and Deng, Zhun and Ji, Xu and Huang, Jiaoyang}, booktitle={ICML}, year={2023}}

@article{russo2016controlling, title={How Much Does Your Data Exploration Overfit? Controlling Bias via Information Usage}, author={Russo, Daniel and Zou, James}, journal={IEEE Transactions on Information Theory}, volume={66}, number={1}, pages={302--323}, year={2020}, note={Conference version: AISTATS 2016}}

@inproceedings{xu2017information, title={Information-Theoretic Analysis of Generalization Capability of Learning Algorithms}, author={Xu, Aolin and Raginsky, Maxim}, booktitle={NeurIPS}, year={2017}}

@inproceedings{steinke2020cmi, title={Reasoning About Generalization via Conditional Mutual Information}, author={Steinke, Thomas and Zakynthinou, Lydia}, booktitle={COLT}, year={2020}}

@inproceedings{harutyunyan2020label, title={Improving Generalization by Controlling Label-Noise Information in Neural Network Weights}, author={Harutyunyan, Hrayr and Reing, Kyle and Ver Steeg, Greg and Galstyan, Aram}, booktitle={ICML}, year={2020}}

@inproceedings{harutyunyan2021fcmi, title={Information-Theoretic Generalization Bounds for Black-Box Learning Algorithms}, author={Harutyunyan, Hrayr and Raginsky, Maxim and Ver Steeg, Greg and Galstyan, Aram}, booktitle={NeurIPS}, year={2021}}

@inproceedings{bassily2018learners, title={Learners that Use Little Information}, author={Bassily, Raef and Moran, Shay and Nachum, Ido and Shafer, Jonathan and Yehudayoff, Amir}, booktitle={ALT}, year={2018}}

@inproceedings{livni2023lower, title={Information Theoretic Lower Bounds for Information Theoretic Upper Bounds}, author={Livni, Roi}, booktitle={NeurIPS}, year={2023}}

@inproceedings{haghifam2023limitations, title={Limitations of Information-Theoretic Generalization Bounds for Gradient Descent Methods in Stochastic Convex Optimization}, author={Haghifam, Mahdi and Rodr{\'\i}guez-G{\'a}lvez, Borja and Thobaben, Ragnar and Skoglund, Mikael and Roy, Daniel M. and Dziugaite, Gintare Karolina}, booktitle={ALT}, year={2023}}

@inproceedings{feldman2020memorization, title={Does Learning Require Memorization? A Short Tale about a Long Tail}, author={Feldman, Vitaly}, booktitle={STOC}, year={2020}}

@inproceedings{brown2021memorization, title={When is Memorization of Irrelevant Training Data Necessary for High-Accuracy Learning?}, author={Brown, Gavin and Bun, Mark and Feldman, Vitaly and Smith, Adam and Talwar, Kunal}, booktitle={STOC}, year={2021}}

@article{bartlett2020benign, title={Benign Overfitting in Linear Regression}, author={Bartlett, Peter L. and Long, Philip M. and Lugosi, G{\'a}bor and Tsigler, Alexander}, journal={PNAS}, volume={117}, number={48}, pages={30063--30070}, year={2020}}

@inproceedings{belkin2019interpolation, title={Does Data Interpolation Contradict Statistical Optimality?}, author={Belkin, Mikhail and Rakhlin, Alexander and Tsybakov, Alexandre B.}, booktitle={AISTATS}, year={2019}}

@article{belkin2019reconciling, title={Reconciling Modern Machine-Learning Practice and the Classical Bias--Variance Trade-off}, author={Belkin, Mikhail and Hsu, Daniel and Ma, Siyuan and Mandal, Soumik}, journal={PNAS}, year={2019}}

@article{muthukumar2020harmless, title={Harmless Interpolation of Noisy Data in Regression}, author={Muthukumar, Vidya and Vodrahalli, Kailas and Subramanian, Vignesh and Sahai, Anant}, journal={IEEE JSAIT}, year={2020}}

@inproceedings{mallinar2022taxonomy, title={Benign, Tempered, or Catastrophic: A Taxonomy of Overfitting}, author={Mallinar, Neil and Simon, James B. and Abedsoltan, Amirhesam and Pandit, Parthe and Belkin, Mikhail and Nakkiran, Preetum}, booktitle={NeurIPS}, year={2022}}

@article{wyner2017explaining, title={Explaining the Success of AdaBoost and Random Forests as Interpolating Classifiers}, author={Wyner, Abraham J. and Olson, Matthew and Bleich, Justin and Mease, David}, journal={JMLR}, year={2017}}

@inproceedings{adlam2020understanding, title={Understanding Double Descent Requires a Fine-Grained Bias-Variance Decomposition}, author={Adlam, Ben and Pennington, Jeffrey}, booktitle={NeurIPS}, year={2020}}

@inproceedings{yang2020rethinking, title={Rethinking Bias-Variance Trade-off for Generalization of Neural Networks}, author={Yang, Zitong and Yu, Yaodong and You, Chong and Steinhardt, Jacob and Ma, Yi}, booktitle={ICML}, year={2020}}

@inproceedings{domingos2000unified, title={A Unified Bias-Variance Decomposition for Zero-One and Squared Loss}, author={Domingos, Pedro}, booktitle={AAAI}, year={2000}}

@inproceedings{jiang2022disagreement, title={Assessing Generalization of {SGD} via Disagreement}, author={Jiang, Yiding and Nagarajan, Vaishnavh and Baek, Christina and Kolter, J. Zico}, booktitle={ICLR}, year={2022}}

@article{nakkiran2020distributional, title={Distributional Generalization: A New Kind of Generalization}, author={Nakkiran, Preetum and Bansal, Yamini}, journal={arXiv preprint arXiv:2009.08092}, year={2020}}

@article{long2010random, title={Random Classification Noise Defeats All Convex Potential Boosters}, author={Long, Philip M. and Servedio, Rocco A.}, journal={Machine Learning}, year={2010}}

@article{schapire1998boosting, title={Boosting the Margin: A New Explanation for the Effectiveness of Voting Methods}, author={Schapire, Robert E. and Freund, Yoav and Bartlett, Peter and Lee, Wee Sun}, journal={Annals of Statistics}, year={1998}}

@article{gacs1973common, title={Common Information is Far Less than Mutual Information}, author={G{\'a}cs, Peter and K{\"o}rner, J{\'a}nos}, journal={Problems of Control and Information Theory}, year={1973}}

@article{feder1994relations, title={Relations Between Entropy and Error Probability}, author={Feder, Meir and Merhav, Neri}, journal={IEEE Trans. Inf. Theory}, year={1994}}

@article{guo2005immse, title={Mutual Information and Minimum Mean-Square Error in {G}aussian Channels}, author={Guo, Dongning and Shamai, Shlomo and Verd{\'u}, Sergio}, journal={IEEE Transactions on Information Theory}, volume={51}, number={4}, pages={1261--1282}, year={2005}}

@inproceedings{zhang2016understanding, title={Understanding Deep Learning Requires Rethinking Generalization}, author={Zhang, Chiyuan and Bengio, Samy and Hardt, Moritz and Recht, Benjamin and Vinyals, Oriol}, booktitle={ICLR}, year={2017}}

@article{mei2022random, title={The Generalization Error of Random Features Regression: Precise Asymptotics and the Double Descent Curve}, author={Mei, Song and Montanari, Andrea}, journal={Communications on Pure and Applied Mathematics}, volume={75}, number={4}, pages={667--766}, year={2022}}

@article{blackwell1953equivalent, title={Equivalent Comparisons of Experiments}, author={Blackwell, David}, journal={Annals of Mathematical Statistics}, volume={24}, pages={265--272}, year={1953}}

@article{lecam1964sufficiency, title={Sufficiency and Approximate Sufficiency}, author={Le Cam, Lucien}, journal={Annals of Mathematical Statistics}, volume={35}, pages={1419--1455}, year={1964}}

@article{dobrushin1962information, title={Information Transmission with Additional Noise}, author={Dobrushin, Roland L. and Tsybakov, Boris S.}, journal={IEEE Transactions on Information Theory}, volume={8}, number={5}, pages={293--304}, year={1962}}

@article{witsenhausen1980indirect, title={Indirect Rate Distortion Problems}, author={Witsenhausen, Hans S.}, journal={IEEE Transactions on Information Theory}, volume={26}, number={5}, pages={518--521}, year={1980}}

@article{crutchfield1989inferring, title={Inferring Statistical Complexity}, author={Crutchfield, James P. and Young, Karl}, journal={Physical Review Letters}, volume={63}, number={2}, pages={105--108}, year={1989}}

@article{cover1967nearest, title={Nearest Neighbor Pattern Classification}, author={Cover, Thomas M. and Hart, Peter E.}, journal={IEEE Transactions on Information Theory}, year={1967}}

@book{devroye1996probabilistic, title={A Probabilistic Theory of Pattern Recognition}, author={Devroye, Luc and Gy{\"o}rfi, L{\'a}szl{\'o} and Lugosi, G{\'a}bor}, publisher={Springer}, year={1996}}

@article{breiman1996bagging, title={Bagging Predictors}, author={Breiman, Leo}, journal={Machine Learning}, year={1996}}

@article{bousquet2002stability, title={Stability and Generalization}, author={Bousquet, Olivier and Elisseeff, Andr{\'e}}, journal={JMLR}, year={2002}}

@inproceedings{natarajan2013noisy, title={Learning with Noisy Labels}, author={Natarajan, Nagarajan and Dhillon, Inderjit S. and Ravikumar, Pradeep K. and Tewari, Ambuj}, booktitle={NeurIPS}, year={2013}}

@inproceedings{patrini2017making, title={Making Deep Neural Networks Robust to Label Noise: A Loss Correction Approach}, author={Patrini, Giorgio and Rozza, Alessandro and Menon, Aditya Krishna and Nock, Richard and Qu, Lizhen}, booktitle={CVPR}, year={2017}}

@inproceedings{kohavi1996bias, title={Bias Plus Variance Decomposition for Zero-One Loss Functions}, author={Kohavi, Ron and Wolpert, David H.}, booktitle={ICML}, year={1996}}

@article{heskes1998bias, title={Bias/Variance Decompositions for Likelihood-Based Estimators}, author={Heskes, Tom}, journal={Neural Computation}, volume={10}, number={6}, pages={1425--1433}, year={1998}}

@inproceedings{platanios2016bayesian, title={Estimating Accuracy from Unlabeled Data: A {B}ayesian Approach}, author={Platanios, Emmanouil Antonios and Dubey, Avinava and Mitchell, Tom M.}, booktitle={ICML}, year={2016}}

@inproceedings{baek2022agreement, title={Agreement-on-the-Line: Predicting the Performance of Neural Networks under Distribution Shift}, author={Baek, Christina and Jiang, Yiding and Raghunathan, Aditi and Kolter, J. Zico}, booktitle={NeurIPS}, year={2022}}

@inproceedings{depeweg2018decomposition, title={Decomposition of Uncertainty in {B}ayesian Deep Learning for Efficient and Risk-Sensitive Learning}, author={Depeweg, Stefan and Hern{\'a}ndez-Lobato, Jos{\'e} Miguel and Doshi-Velez, Finale and Udluft, Steffen}, booktitle={ICML}, year={2018}}

@article{arlot2010survey,
  author = {Arlot, Sylvain and Celisse, Alain},
  title = {A survey of cross-validation procedures for model selection},
  journal = {Statistics Surveys},
  volume = {4},
  pages = {40--79},
  year = {2010}
}

@book{cover2006elements,
  author = {Cover, Thomas M. and Thomas, Joy A.},
  title = {Elements of Information Theory},
  edition = {2nd},
  publisher = {Wiley},
  year = {2006}
}

@incollection{miller1955note,
  author = {Miller, George A.},
  title = {Note on the bias of information estimates},
  booktitle = {Information Theory in Psychology: Problems and Methods},
  editor = {Quastler, Henry},
  pages = {95--100},
  publisher = {Free Press},
  year = {1955}
}

@article{hastie2022surprises,
  author = {Hastie, Trevor and Montanari, Andrea and Rosset, Saharon and Tibshirani, Ryan J.},
  title = {Surprises in high-dimensional ridgeless least squares interpolation},
  journal = {The Annals of Statistics},
  volume = {50},
  number = {2},
  pages = {949--986},
  year = {2022}
}

@article{audibert2007fast,
  author = {Audibert, Jean-Yves and Tsybakov, Alexandre B.},
  title = {Fast learning rates for plug-in classifiers},
  journal = {The Annals of Statistics},
  volume = {35},
  number = {2},
  pages = {608--633},
  year = {2007}
}

@article{olson2017pmlb,
  author = {Olson, Randal S. and La Cava, William and Orzechowski, Patryk and Urbanowicz, Ryan J. and Moore, Jason H.},
  title = {{PMLB}: a large benchmark suite for machine learning evaluation and comparison},
  journal = {BioData Mining},
  volume = {10},
  pages = {36},
  year = {2017}
}

@book{james2023islp,
  author = {James, Gareth and Witten, Daniela and Hastie, Trevor and Tibshirani, Robert and Taylor, Jonathan},
  title = {An Introduction to Statistical Learning: with Applications in {Python}},
  publisher = {Springer},
  year = {2023}
}

@article{pedregosa2011scikit,
  author = {Pedregosa, Fabian and others},
  title = {Scikit-learn: Machine Learning in {Python}},
  journal = {Journal of Machine Learning Research},
  volume = {12},
  pages = {2825--2830},
  year = {2011}
}

@article{james2003variance,
  author = {James, Gareth M.},
  title = {Variance and bias for general loss functions},
  journal = {Machine Learning},
  volume = {51},
  number = {2},
  pages = {115--135},
  year = {2003}
}

@inproceedings{benhur2002stability,
  author = {Ben-Hur, Asa and Elisseeff, Andr{\'e} and Guyon, Isabelle},
  title = {A stability based method for discovering structure in clustered data},
  booktitle = {Pacific Symposium on Biocomputing},
  pages = {6--17},
  year = {2002}
}

@article{lange2004stability,
  author = {Lange, Tilman and Roth, Volker and Braun, Mikio L. and Buhmann, Joachim M.},
  title = {Stability-based validation of clustering solutions},
  journal = {Neural Computation},
  volume = {16},
  number = {6},
  pages = {1299--1323},
  year = {2004}
}

@article{tibshirani2005prediction,
  author = {Tibshirani, Robert and Walther, Guenther},
  title = {Cluster validation by prediction strength},
  journal = {Journal of Computational and Graphical Statistics},
  volume = {14},
  number = {3},
  pages = {511--528},
  year = {2005}
}

@article{vonluxburg2010clustering,
  author = {von Luxburg, Ulrike},
  title = {Clustering stability: an overview},
  journal = {Foundations and Trends in Machine Learning},
  volume = {2},
  number = {3},
  pages = {235--274},
  year = {2010}
}

@inproceedings{bendavid2006sober,
  author = {Ben-David, Shai and von Luxburg, Ulrike and P{\'a}l, D{\'a}vid},
  title = {A sober look at clustering stability},
  booktitle = {Conference on Learning Theory (COLT)},
  pages = {5--19},
  year = {2006}
}

@inproceedings{hellstrom2022samplewise,
  author = {Hellstr{\"o}m, Fredrik and Durisi, Giuseppe},
  title = {A new family of generalization bounds using samplewise evaluated {CMI}},
  booktitle = {Advances in Neural Information Processing Systems},
  year = {2022}
}

@inproceedings{koh2017influence,
  author = {Koh, Pang Wei and Liang, Percy},
  title = {Understanding black-box predictions via influence functions},
  booktitle = {International Conference on Machine Learning (ICML)},
  year = {2017}
}

@inproceedings{ilyas2022datamodels,
  author = {Ilyas, Andrew and Park, Sung Min and Engstrom, Logan and Leclerc, Guillaume and Madry, Aleksander},
  title = {Datamodels: Predicting predictions from training data},
  booktitle = {International Conference on Machine Learning (ICML)},
  year = {2022}
}

@inproceedings{feldman2020neural,
  author = {Feldman, Vitaly and Zhang, Chiyuan},
  title = {What neural networks memorize and why: Discovering the long tail via influence estimation},
  booktitle = {Advances in Neural Information Processing Systems},
  year = {2020}
}
